%% file: arxiv_main.tex
\pdfoutput=1
\documentclass[10pt]{article}
\usepackage[preprint]{tmlr}
\input{math_commands}

\usepackage{xcolor}
\usepackage[hidelinks]{hyperref}
\usepackage{url}
\usepackage{booktabs}
\usepackage{amsmath}
\usepackage{amssymb}
\usepackage{graphicx}
\usepackage{algorithm}
\usepackage{algpseudocode}

\title{When Residualization Helps an Audit:\\Format Effects, Slice Gains, and Their Limits}

\author{%
\centering
\begin{tabular}{@{}cc@{}}
{\normalsize\bf Daein Weon} & {\normalsize\bf Dong Ho Kang} \\[1pt]
{\small\it Kookmin University} & {\small\it UStechlab} \\[1pt]
{\small\tt wdi1024@kookmin.ac.kr} & {\small\tt donghokang@ustechlab.com} \\[5pt]
\multicolumn{2}{c}{\small Correspondence to: Dong Ho Kang \textless\texttt{donghokang@ustechlab.com}\textgreater.}
\end{tabular}}

\newcommand{\srep}{s_{\mathrm{adj}}}
\definecolor{passgreen}{HTML}{2E8540}
\definecolor{failred}{HTML}{B3392E}
\newcommand{\pass}{\textcolor{passgreen}{\checkmark}}
\newcommand{\fail}{\textcolor{failred}{$\times$}}
\newcommand{\vrep}{\textcolor{passgreen}{\textbf{slice AUC improved}}}
\newcommand{\vscr}[1]{\textcolor{failred}{no usable slice} {\footnotesize(#1)}}
\newcommand{\vnv}[1]{\textcolor{failred}{refused at the screen} {\footnotesize(#1)}}

\begin{document}
\maketitle

\begin{abstract}
Evaluation scores used around LLM systems---including reward models, rerankers, and LLM judges---can track
surface form instead of the quality they claim to measure. When presented with a terse correct
solution and a commented buggy solution for the same MBPP problem, a public preference reward
model selects the correct one no better than a coin flip ($0.507$). Subtracting the predictable
surface component from such scores is increasingly common, but removal alone does not yield a more
valid measurement: the removed component may carry construct-relevant signal, and residualization cannot tell which is
which. Under designed
interventions---unit-test labels with comment-only edits---residualization attenuates the reward
model's format effects by about $0.12$ on both correct and buggy code, while the
correct-versus-buggy margins move by less than $0.01$; a scalar length-only correction can amplify
the same dependence. In observational NLI and
QA settings, we freeze a held-out replication before scoring and re-evaluate it using labels from
disjoint annotators; this supports only a narrower conclusion: better agreement with the construct
labels on a pre-declared slice where a surface-only predictor errs, not a repaired score. Full-population agreement falls in every observational setting with a reported positive slice
gain, and within-question ranking falls in every such QA setting. When construct and surface features are entangled,
residualization can decorrelate a score while degrading construct alignment, and, in a controlled
model, configurations just as damaging to construct alignment pass every pre-adjustment check, so no
committed gate is a guarantee. Decorrelation alone therefore does not establish
validity restoration. We assemble these distinctions into a reporting protocol whose outcomes,
refusal included, state what an adjusted score may be claimed to show: an audit-time diagnostic
reported beside the construct-alignment cost it incurs, never a replacement for the raw score.
\end{abstract}

\section{Introduction}
\label{sec:intro}

The scores that steer LLM systems---reward models, rerankers, and LLM judges among them---are
often predictable from surface features such as length, position, and lexical overlap;
LLM judges in particular exhibit position, verbosity, and self-enhancement biases
\citep{zheng2023judging}. Benchmark labels have likewise long been recoverable from
annotation artifacts
\citep{gururangan2018annotation,poliak2018hypothesis,geirhos2020shortcut,singhal2023long,measuring2025matters}.
The dependence can be substantial: when offered two solutions to an MBPP problem, one
terse and correct and the other commented and buggy, a public preference reward model
selects the correct solution at a rate near chance ($0.507$), compared with $0.615$ when the verbosity--correctness
pairing is reversed. Subtracting a predictable surface component before reporting
has accordingly become routine, as in length-controlled win rates
\citep{dubois2024length}, but such adjustments are rarely accompanied by evidence that
the resulting score is a more valid measurement. This distinction matters: subtraction
is always computationally possible, but when the construct itself correlates with the
artifact, the residual can discard construct-relevant signal and become a worse
measurement than the original score.

We ask one question: \emph{when an adjusted evaluation score improves, what exactly has been shown to
improve?} Our goal is not a better deployment score but a determination of what audit evidence a
post-hoc adjustment legitimately provides: throughout, residualization is used as an audit
perturbation, not as a repaired score. The answer depends on the evidential regime: direct
effect attenuation under a designed intervention, a slice-restricted alignment change under
observational conditioning, and no validity conclusion under construct--artifact entanglement.

Where a designed intervention is available, the answer is direct. In a code audit where unit-test execution supplies
correctness labels and only comments and docstrings manipulate the format channel, residualization attenuates the format effects to estimates whose intervals include zero while leaving the correctness margins essentially
unchanged, reducing the $0.108$ gap between the mirrored preferences to $0.037$
(Section~\ref{sec:rmcode}). None of the length-only corrections we evaluated closes that gap.

Observational settings cannot support the same claim, and we do not extend it to them. Because the
slice on which improvement is measured is itself defined by the errors of a surface-only predictor,
the held-out and disjoint-annotator tests support only a slice-restricted alignment gain, a
deliberately smaller claim than score repair, and every outcome measured outside that slice declines
(Section~\ref{sec:positives}). When construct and artifact are entangled, neither decorrelation nor
a slice gain identifies validity: the adjustment can damage an otherwise effective judge while the
measured diagnostics improve (Section~\ref{sec:synthetic}). In the four current-generation judge$\times$setting probes we ran, every probe falls below the
committed loading gate (Section~\ref{sec:judgeprobe}).
Section~\ref{sec:protocol} assembles these distinctions into a reporting protocol with an
explicit refusal outcome, a claim withheld for insufficient evidence.

The closest prior work asks a related question in the pairwise setting.
\citet{xu2026debias} characterize when pairwise judge debiasing is identifiable, and supply
gate-based analyses and paired rendering designs. We ask a different question for pointwise scores:
once a post-hoc adjustment has been applied, which observed improvements warrant a measurement claim
and which do not --- for a residualized pointwise score, surface-effect removal, conditioned slice
alignment, construct validity, and deployment utility turn out to be distinct claims that can move
in opposite directions on the same data. The measurements below supply the evidence for that
distinction; none of them is evaluated in the pairwise setting.

\textbf{Contributions.} The paper tests one thesis in three evidence regimes --- that surface-effect
removal, slice-level alignment, construct validity, and deployment utility are empirically distinct
outcomes of one adjustment:
\textbf{(1) Designed evidence.} Where paired label-preserving interventions supply the measurement target, the
attenuation of a named format effect is directly measurable; the code audit measures it, and also
bounds it: across three held-out edit templates the same operator overshoots one, attenuates
another, and amplifies a third (Section~\ref{sec:rmcode}).
\textbf{(2) Observational evidence.} Held-out and disjoint-annotator replication establishes a
slice-restricted alignment gain and nothing wider: every outcome measured outside the declared
slice---full-population alignment, the observable flagged subgroup, within-question
ranking---declines under adjustment (Sections~\ref{sec:positives} and~\ref{sec:costs}).
\textbf{(3) The validity boundary.} Under construct--artifact entanglement, neither decorrelation
nor a slice gain identifies validity restoration: the measured loading can rise while the
adjustment damages an otherwise effective judge, and in a controlled model every pre-adjustment
gate passes in exactly that harmful regime (Section~\ref{sec:synthetic}).
These findings motivate the conservative reporting procedure of Section~\ref{sec:protocol}, which
we present as an implication of the results rather than as a separate method.

\section{Measurement Setup and Analysis Criteria}
\label{sec:method}

An audit provides a score, surface features, and a provenance-separated construct measurement. This section fixes the three different things ``the adjusted score is better'' can mean (Section~\ref{sec:targets}), the objects and the operator that measure them (Section~\ref{sec:setup}), and the diagnostics and thresholds (Sections~\ref{sec:condition} and~\ref{sec:checks}). These are fixed analysis diagnostics that organize the empirical analyses of
Sections~\ref{sec:rmcode}--\ref{sec:synthetic}; we introduce no new residualization estimator, and
the methodological contribution is the evidence-and-reporting framework these diagnostics compose.
Their development chronology and subsequent freezes are reported in Appendix~\ref{app:accounting}. None of them certifies a safe
adjustment; they determine only which audit statement is admissible under the measurements
available. Section~\ref{sec:protocol} assembles them into the reporting procedure, which determines
what an observed adjustment may support.

\subsection{Three evaluation targets}
\label{sec:targets}

The statement ``the adjusted score is better'' can refer to three distinct claims, each measured on a different object. \emph{Removal of a named surface effect} is a claim about paired intervention contrasts: the same item, the surface channel manipulated, everything else fixed by construction; it is verifiable exactly when such pairs exist with labels fixed under the edit (Section~\ref{sec:rmcode}). \emph{Construct alignment} is a claim about agreement with the construct measurement on a stated population, and it is population-relative: the full evaluation set, a label-defined slice, and an observable subgroup can move in opposite directions under one adjustment, and in this paper they systematically do (Sections~\ref{sec:positives} and~\ref{sec:discussion}). \emph{Deployment performance} is a claim about a consumer of the score, within-question ranking or an observable decision rule; a slice-restricted alignment gain does not imply it. Every improvement claim below names its target and its population; keeping the three apart is most of what the reporting protocol of Section~\ref{sec:protocol} enforces.

\begin{center}\small
\begin{tabular}{@{}lll@{}}
\toprule
Regime & What can be established & What cannot \\
\midrule
Designed intervention & attenuation of a named surface effect & transfer to unseen edit styles \\
Observational slice & a slice-restricted alignment change & repair, validity, or deployment gain \\
Entangled construct--artifact & \multicolumn{2}{l}{no validity claim is identified by decorrelation or a slice gain alone} \\
\bottomrule
\end{tabular}
\end{center} 

\subsection{Ingredients and operator}
\label{sec:setup}

\textbf{The procedure in one setting.} In WikiQA the audited score is a
cross-encoder's relevance score for a candidate answer sentence. We first fit a predictor that sees
only surface cues---how much the candidate overlaps the question, how long it is. Its errors define
the candidate slice, and we use that slice only when the predictor is informative enough to make it
non-arbitrary; otherwise the setting has no usable slice. That eligibility bar does not make the
cues construct-external --- a cue can predict the label without belonging to the construct the
score is meant to measure, a distinction Section~\ref{sec:condition} takes up. We cross-fit the part of the audited score that the surface cues predict in order to measure the
score's loading on them; the setting passes the screen only if that loading is present and the
scorer is markedly worse on the slice than on the full set, and only then do we form the adjusted
score by subtracting the fitted part. The single quantity we call a \emph{gain} is the change in label alignment
on that same slice, before versus after adjustment; what happens off the slice is reported beside it as a
cost. The rest of this section names these steps.

That walkthrough fixes the symbols: in WikiQA, $s$ is the cross-encoder's relevance score, $\phi$ collects the answer-only surface features plus question--answer overlap just described, and $y$ is the crowdsourced relevance label. Table~\ref{tab:phispec} (Appendix~\ref{app:settings}) specifies $\phi$ for every setting. In general an audit supplies three objects on an evaluation set: the score $s(x)\in\mathbb{R}$; prespecified partial-input surface features $\phi(x)$
such as position, length, or one-sided lexical cues, which may or may not carry a
construct-external component (one that may correlate with $y$ but is not part of what $s$ is
intended to measure); and a \emph{provenance-separated construct measurement} $y(x)$. We call $\phi$ a \emph{surface} channel throughout and reserve \emph{artifact} for the cases where construct-externality is secured by design or by external evidence --- the code audit of Section~\ref{sec:rmcode} and the synthetic models, where the artifact is fixed by construction --- because no observational statistic here settles that question. What \emph{provenance-separated} requires of $y$ is defined in Section~\ref{sec:condition}, next to the assumptions it feeds. We write $A(u)=\mathrm{AUC}(y,u)$ for the construct alignment of a score $u$.

The observational audit uses two fitted models and one reported contrast. The first model asks
where a surface-only view of the data fails: the \emph{surface predictor} $\hat a$ --- a predictor
of the \emph{label} that sees only surface cues --- is the out-of-fold prediction of $y$ from the
putative-artifact block $\phi_a$ of $\phi$ (the four dense cues in the QA settings; all of $\phi$
elsewhere; Table~\ref{tab:phispec}), and the \emph{slice} is the set of examples its thresholded
prediction gets wrong. The second model asks how much of the score the surface features predict:
the \emph{removed component} $\hat g$ is the out-of-fold ridge prediction of $s$ from $\phi$, which
the adjustment subtracts; the adjustment itself always uses all of $\phi$. The reported contrast
$G$ is what changes on the declared slice once $\hat g$ is gone. The box below fixes the remaining
notation. \begin{center}
\fbox{\begin{minipage}{0.955\textwidth}
\small
\textbf{Symbols.} $s$ audited score; $\phi$ prespecified partial-input surface features; $y$ provenance-separated construct measurement; $\hat a$ surface predictor (out-of-fold prediction of $y$ from $\phi_a$, the putative-artifact block of $\phi$); $\hat g$ removed component (out-of-fold prediction of $s$ from $\phi$); $\srep=s-\hat g$ adjusted score; $A(u)=\mathrm{AUC}(y,u)$ construct alignment, with $A_{\mathrm{slice}}$ its restriction to the slice; $G=A_{\mathrm{slice}}(\srep)-A_{\mathrm{slice}}(s)$ the reported slice-restricted gain. The remaining statistics are introduced where they are used: the screen's two components in Section~\ref{sec:condition}, the decorrelation index in Section~\ref{sec:checks}. Three quantities are easy to conflate, so we separate them once:

\begin{center}\small
\begin{tabular}{@{}lll@{}}
\toprule
Quantity & What it uses & What it is for \\
\midrule
$\hat a:\phi_a \to y$ & the putative-artifact block of $\phi$ & defines the error slice \\
$\hat g:\phi \to s$ & all of $\phi$ & the component the adjustment subtracts \\
$G$ & the slice, before and after & tests slice alignment only \\
\bottomrule
\end{tabular}
\end{center}

\textbf{Two pre-adjustment statistics.} The \emph{slice-eligibility check} and the \emph{screen} are both computed \emph{before} adjusting, and they
ask different questions. The screen compares two statistics against gates and can refuse outright. If the slice-eligibility check fails, the outcome is \emph{no usable slice}. We distinguish two reasons with the loading statistic: either the measured channel is weak, or the channel is present but the label-defined slice is not an informative measurement target. The decorrelation diagnostic and the slice-gain test are computed \emph{after}; a \emph{gate} is
the numerical threshold a statistic is compared against. The \emph{slice} is the label-defined set where $\hat a$ errs; the observable \emph{flagged subgroup} ($\hat a>\tau$) and a dataset-defined \emph{subset} such as a HANS heuristic are different populations, and this paper keeps the three apart. 
\end{minipage}}
\end{center}

Adjustment uses cross-fitted residualization,
\begin{equation}\label{eq:op}
\srep(x) \;=\; s(x) \, - \, \hat g_{-k(x)}\!\big(\phi(x)\big),
\end{equation}
with $\hat g_{-k}$ fit on all folds except the one containing $x$ (5-fold; question-grouped for the QA settings, so that candidates of the same question never span a train/held-out boundary). Unless a table is explicitly labeled otherwise, every number in this paper (slice-eligibility check, screen, checks, and gains) is computed under this grouped protocol. Row-level variants and the correspondence between the two protocols are in Appendix~\ref{app:grouped}, and every table reporting them says so in its caption. Cross-fitting removes in-sample fitting bias from the subtraction. It does not guard against feature- or protocol-level adaptivity, which the held-out runs of Section~\ref{sec:positives} address separately.

Residualization is not invariant to monotone transformations of $s$, so the representation is declared per scorer family: the raw cross-encoder logit in the QA settings and the classifier's class probability in the classification settings, with the tested settings' outcomes unchanged under three alternative representations. The primary analysis uses the frozen native-scale ridge specification; rank-Gaussian representation and the conditional-quantile eraser are reported as sensitivity analyses in Appendices~\ref{app:grouped} and~\ref{app:quantile}, and gate and operator robustness in Appendix~\ref{app:gaterobust}. No model is retrained at any point: Eq.~(\ref{eq:op}) is classical partialling-out, and the operator costs one cross-fitted regression.

\subsection{Before any adjustment: a slice-eligibility check and a screen}
\label{sec:condition}

\textbf{Observational slice-eligibility check.} The slice-eligibility check does not test whether a surface artifact
exists, nor whether removing it will help. It asks only whether the errors of the surface-only
predictor define a non-arbitrary population on which a slice change can be measured, and so requires
a $\phi$-only model to predict the construct label at $A(\hat a)\gtrsim0.65$; the appendix tables abbreviate it as \emph{router}. A \emph{measurement target} is the
object on which improvement is measured: a label-defined subset here, or paired label-preserving
contrasts in a designed audit. That is why an eligibility failure and a successful designed audit are not
in tension --- the slice is unusable, but the paired contrasts are a better measurement target
(Section~\ref{sec:rmcode}). The check routes on measurement-target availability rather than predicting harm: on the
observational analysis rows it changes no decision, and it refuses some settings that would have
gained. Shortcuts a linear model cannot recover also fail it, so a refusal
here is not evidence that no artifact exists \citep{feng2019misleading}; the gate value, its
sensitivity, and the refused candidates are in Appendix~\ref{app:screen}.

\textbf{The screen (surface-associated degradation).} The \emph{loading} $R^2_{\phi\to s}$, computed before any adjustment, measures whether the score loads on the measured surface channel at all. The \emph{slice degradation} $\Delta_{\mathrm{slice}}=A(s)-A_{\mathrm{slice}}(s)$ measures whether that loading costs alignment where the surface predictor errs. The screen passes when both are materially positive ($R^2_{\phi\to s}\gtrsim0.05$, $\Delta_{\mathrm{slice}}\gtrsim0.03$). Both components are needed, because they fail in complementary ways: a score can load heavily on the surface channel without being misled by it (QQP), or be degraded where the surface predictor errs without riding the linear channel (MNLI). Their readings are in Table~\ref{tab:scope} (Section~\ref{sec:samelabel}), reported with the settings it scopes in Section~\ref{sec:positives}.

\textbf{The construct measurement's provenance.} We call $y$ \emph{provenance-separated} when the process generating it targets the construct that $s$ claims to measure and has access to neither $s$ nor the annotation pipeline behind it. This is a criterion on where the labels came from, not a statistical one (separately produced labels can still share annotation heuristics), so Table~\ref{tab:indep} (Appendix~\ref{app:settings}) records each setting's provenance and shared-convention risk; throughout, the term carries only this sourcing sense. Statistical independence of $y$'s measurement error from the artifact channel is the separate assumption A2 below.

The slice-eligibility check and the screen are heuristics rather than an identification result: the slice-eligibility check establishes only that the prespecified surface features are predictive enough of the construct label to induce a non-arbitrary error slice; it says nothing about whether the score rides that channel, nor whether the channel carries construct signal. The further inference that removing this channel preserves the construct rests on two assumptions that these statistics cannot check. \textbf{(A1)} $\phi$ is construct-external on the target distribution. Construct-external does not
mean statistically unrelated to $y$: a surface cue can predict the construct label strongly on the
observed distribution without being part of the construct the score is meant to measure, which is
why the slice-eligibility check can require a predictive $\hat a$ while A1 still asks something different. \textbf{(A2)} the measurement error of $y$, where $y$ deviates from the construct it targets, is unrelated to the artifact channel. Where A1 is debatable (lexical overlap in the QA settings is this paper's own borderline case), no observational statistic here resolves A1; Section~\ref{sec:rmcode} establishes it by design in one regime, and Appendix~\ref{app:c2} reports a rule-based overlap intervention on held-out SQuAD: label-preserving edits move the raw score strongly while leaving the residualized score largely insensitive, so, in the manipulated directions, the subtracted component tracks overlap rather than the preserved label.

The numerical gates are heuristic defaults rather than identified constants. On the settings
evaluated here, outcomes are insensitive to the gate values within the measured gap between
positives and negatives; a new domain should re-measure that gap rather than reuse these constants
(Appendix~\ref{app:threshold}).

\subsection{Post-adjustment slice-gain test and decorrelation diagnostic}
\label{sec:checks}

\textbf{Index decorrelation.} To claim index decorrelation, the adjusted score must be
uncorrelated with the prespecified surface index: the adjusted score's absolute linear correlation with the surface predictor's decision
function $\hat c$ (the logit of the surface predictor, $\hat a=\sigma(\hat c)$, and the single
label-predictive surface direction), $R_{\mathrm{lin}}(\srep):=|\mathrm{corr}(\srep,\hat c)|$, must satisfy
$R_{\mathrm{lin}}(\srep)\le0.05$ on the point estimate, with a bootstrap interval reported alongside; failing it withdraws the decorrelation claim rather than the improvement (Section~\ref{sec:protocol}; Appendix~\ref{app:certprobes} defines the index and its two secondary diagnostics). It claims only low linear correlation with one prespecified index: not that every artifact
direction is removed, and not that nonlinear consumers are constrained. Orthogonality is exact only
for the population projection, so the criterion is re-measured per evaluation set; across the seven
positives it is the component that most often qualifies or withdraws a claim.

\textbf{Slice-gain test.} The slice gain $G=A_{\mathrm{slice}}(\srep)-A_{\mathrm{slice}}(s)$ must be positive with its example-level bootstrap interval, clustered by query or problem, excluding zero: the test asks whether alignment improves on the declared slice once the measured surface component is
removed. Passing it establishes a positive change on the declared slice and nothing more: it does
not validate artifact removal or score validity, and Section~\ref{sec:attrdef} shows why the slice
construction already favors a positive result. We therefore read this stage as a test of the slice gain, not as a validation of the adjustment, and use it only to reject adjustments that fail it. The two post-adjustment checks are asymmetric by design: the slice-gain test supports the claim and
so demands affirmative interval evidence, while the gates and the decorrelation diagnostic compare fixed
constants on point estimates, with intervals reported beside them (Appendix~\ref{app:rules} gives
the reason and the strict-interval reading).
\section{A Designed Intervention Audit: a Reward Model on Code}
\label{sec:rmcode}

The designed audit fixes the reference point for the rest of the paper: the labels are mechanical, and only the surface channel is manipulated. The score is a public preference reward model (OpenAssistant DeBERTa-v3-base; \citealp{kopf2023openassistant}) scoring code against a problem statement, and the construct label is produced by executing each candidate against the problem's own unit tests, a mechanically determined $y$ with no annotator in the loop. From 353 MBPP problems \citep{austin2021mbpp} we construct a $2{\times}2$ design per problem: the canonical solution (correct, terse); a single-AST-mutation variant verified to fail the tests (buggy, terse); and comment/docstring-only rewrites of both, whose labels are re-verified by execution; correctness here means passing the benchmark's unit tests, not full semantic correctness.

\begin{center}\small
\begin{tabular}{llrr}
\toprule
Candidate (MBPP 308: cubes of list elements via a lambda) & tests & raw $s$ & residual $\srep$ \\
\midrule
correct, terse (\texttt{x ** 3}, 91 chars) & pass & $-0.984$ & $-1.424$ \\
correct, verbose (docstring and comments added, 361 chars) & pass & $-0.893$ & $-1.425$ \\
buggy, terse (\texttt{x ** 4}) & fail & $-1.131$ & $-1.651$ \\
buggy, verbose & fail & $-0.955$ & $-1.556$ \\
\bottomrule
\end{tabular}
\end{center}

\noindent One problem from the design. The raw reward ranks the commented buggy solution above the terse correct one ($-0.955$ against $-0.984$); after residualization the two correct candidates score alike and both rank above both buggy ones. All four candidates share one AST-normalized code style, so the format channel $\phi$ is label-orthogonal \emph{by construction} (Appendix~\ref{app:rmcode}).

\begin{table}[!ht]
\centering
\small
\caption{Reward-model code audit (353 MBPP problems, $2{\times}2$ cells per problem, $n{=}1412$): within-problem paired effects on the reward, raw vs.\ residualized, with problem-level bootstrap 95\% CIs. Format edits are comment/docstring-only and every label is re-verified by unit-test execution. The construct-margin rows are a design check; the cross-format preference rows show the practical consequence (\texttt{rm\_code\_paired\_change.py}).}
\label{tab:rmcode}
\setlength{\tabcolsep}{3pt}
\renewcommand{\arraystretch}{1.15}
\begin{tabular}{lccc}
\toprule
Paired effect (within problem) & Raw & Residualized & Change \\
\midrule
Format ($+$comments) at fixed correct code & $+0.083$ {\scriptsize$[+0.011,+0.157]$} & $-0.041$ {\scriptsize$[-0.110,+0.031]$}  & -0.124 {\scriptsize$[-0.142,-0.104]$} \\
Format ($+$comments) at fixed buggy code & $+0.154$ {\scriptsize$[+0.078,+0.228]$} & $+0.035$ {\scriptsize$[-0.030,+0.102]$}  & -0.119 {\scriptsize$[-0.135,-0.103]$} \\
Construct (correct$\,-\,$buggy) at fixed terse & $+0.158$ {\scriptsize$[+0.128,+0.187]$} & $+0.160$ {\scriptsize$[+0.130,+0.191]$}  & +0.002 {\scriptsize$[-0.004,+0.009]$} \\
Construct (correct$\,-\,$buggy) at fixed verbose & $+0.087$ {\scriptsize$[+0.069,+0.106]$} & $+0.084$ {\scriptsize$[+0.064,+0.104]$}  & -0.003 {\scriptsize$[-0.010,+0.003]$} \\
\midrule
$P(\text{terse correct} \succ \text{verbose buggy})$ & $0.507$ {\scriptsize$[0.453,0.555]$} & $0.586$ {\scriptsize$[0.535,0.637]$} \\
$P(\text{verbose correct} \succ \text{terse buggy})$ & $0.615$ {\scriptsize$[0.564,0.663]$} & $0.550$ {\scriptsize$[0.499,0.603]$} \\
\bottomrule
\end{tabular}
\end{table}

The paired design isolates format changes at fixed unit-test labels; none of our observational settings supports an equivalent claim. For the construct dimension this audit declares --- functional correctness under the benchmark's tests --- A1 is discharged by the intervention design: the edits alter only comments and docstrings, executable behavior is preserved, and every label is re-verified by unit-test execution, so the format effect and its attenuation are measured by intervention rather than assumed. We do not claim that comments are irrelevant to every dimension of response quality a general preference model may encode; the claim is scoped to the declared dimension. The pre-adjustment statistics also read differently here than on observational data: with $\phi$ orthogonal to the label, $\hat a$ carries no information, the slice-eligibility check statistic reads exactly the chance value the design implies ($A(\hat a)=0.500$ under problem-grouped folds), and the label-defined slice is uninformative because its membership is nearly arbitrary---while the scorer depends strongly on the measured channel all the same ($R^2_{\phi\to s}=0.283$). The measurement target is therefore supplied by design: the paired interventions stand in for the slice. In this branch the paired contrasts
measure the format effect directly---their post-adjustment intervals cover zero
(Table~\ref{tab:rmcode})---so a separate decorrelation diagnostic is unnecessary. Section~\ref{sec:protocol} returns to this configuration as the protocol's designed branch, whose reported outcome is \emph{measured format attenuation}.

Table~\ref{tab:rmcode} gives the decomposition. Adding comments to a solution, changing nothing the unit tests can see, raises the raw reward by $+0.083$ on correct code and $+0.154$ on buggy code, both intervals excluding zero. Residualization shrinks both paired format effects to estimates whose $95\%$ intervals include zero---attenuation, not equivalence or complete removal---while leaving the correct-versus-buggy margins numerically unchanged ($+0.158$ at fixed terse, $+0.087$ at fixed verbose). The residual effects are not statistically distinguishable from zero, whereas the paired raw-to-residualized changes are $-0.124$ $[-0.142,-0.104]$ at fixed correct code and $-0.119$ $[-0.135,-0.103]$ at fixed buggy code (problem-level bootstrap of the within-problem change). Because the residual intervals ($[-0.110,+0.031]$ and $[-0.030,+0.102]$) are too wide to establish equivalence to zero, we claim attenuation rather than removal. The construct margins change by only $+0.002$ $[-0.004,+0.009]$ and $-0.003$ $[-0.010,+0.003]$. Margin preservation has a design-level cause: format is balanced across labels by design, so the subtracted component is nearly label-orthogonal, and subtracting it therefore leaves
the correctness margin nearly unchanged. The preserved margins serve as a design check; the primary evidence is the attenuation of the
paired format contrasts.

Asked to rank a terse correct solution against a commented buggy one, the raw reward model is at a coin flip ($0.507$); the adjusted score prefers the correct one at $0.586$. The two contrasts are sharply asymmetric before adjustment ($0.507$ against $0.615$) and move together after it ($0.586$ against $0.550$). They do not coincide because the adjusted construct margin is itself smaller on verbose code than on terse (non-overlapping intervals): a format$\times$construct interaction of the scorer, which reads correctness less well once code is commented, and which residualization on a format-only $\phi$ neither removes nor claims to. Pooled AUC barely moves ($0.529\to0.530$), because format is balanced across labels by design, and the within-problem paired contrasts above are what register the effect.

Robustness runs replicate the pattern and sharpen it: rescoring with the larger checkpoint of the same family \emph{reverses the sign} of the bias and residualization attenuates the format effect in that direction too, and on HumanEval, where the format effect dwarfs the construct margin, residualization attenuates the channel ($+1.24\to+0.08$) without establishing removal in an equivalence sense (Appendix~\ref{app:rmcode}).

\textbf{Scalar length correction is not enough.} A third probe varies the \emph{type} of semantics-preserving edit, and the raw response is not even monotone in added text: the base model penalizes a single bare comment line ($-0.23$/$-0.20$) while rewarding docstring-style verbosity ($+0.08$/$+0.15$). A scalar length-only residualization baseline---regress the reward on a length variable and
subtract the fitted component, a pointwise analogue of length-controlled evaluation
\citep{dubois2024length}\footnote{The deployed recipe models \emph{pairwise} preferences
with a GLM, and the LLM-judge chat settings it is deployed in are untested here, so we
treat this only as a scalar length baseline.}---leaves every format effect significant and moves them in opposite directions: the docstring effects
grow several-fold ($+0.08\to+0.62$ on correct code, $+0.15\to+0.68$ on buggy) while the
bare-comment penalties shrink (all four effects with intervals: Appendix~\ref{app:rmcode}). Residualization on the multivariate $\phi$ attenuates all four effects to intervals covering zero, because the channel that defeats the scalar correction is comment \emph{style} (docstring presence versus a bare comment line), which no single length variable indexes and whose sign reverses across edit types. The conclusion is not specific to the linear form: none of the nonlinear length-only corrections we
tested attenuates all four paired format effects to intervals covering zero either
(Table~\ref{tab:nonlinlength}).

\textbf{The correction does not transfer across edit styles.} On three held-out edit templates the
same operator overshoots one, attenuates another, and amplifies a third, and fitting on two
templates together makes the remaining two worse rather than better (Appendix~\ref{app:rmcode});
the measured attenuation is bounded to the edit style the operator was fit on. This failure of
transfer motivates the narrower question of the next section: what can observational evidence
establish when no label-preserving intervention is available?

\section{Observational Slice Gains Replicate but Remain Slice-Restricted}
\label{sec:positives}

We now test the observational regime, asking not whether residualization repairs an artifact but
whether a pre-declared slice-conditional response replicates. The primary observational result is the held-out SQuAD
replication (Section~\ref{sec:heldout}): it froze the complete analysis recipe before any score existed, it
survives the strictest reading of the intervals, and it carries a second, disjoint
annotator-derived evaluation. Section~\ref{sec:samelabel} then reports the five developed settings that
motivated it, spanning NLI (SNLI, SICK) and QA answer selection (WikiQA, ASNQ, TriviaQA), and
Section~\ref{sec:boundary-refusals} examines the settings the same statistics refuse. Under the
point-estimate rule used throughout, five observational slices show a measured gain; under a
uniformly conservative interval rule, only TriviaQA and the held-out SQuAD run remain
(Appendix~\ref{app:rules}), so we treat the remaining rows as corroboration.

Because the slice is defined from the same label--feature relation the adjustment uses, a positive
$G$ is favorably conditioned; in the idealized model the slice construction mechanically favors
subtraction (Property 4, Appendix~\ref{app:identifiability}). A positive $G$ therefore cannot by
itself support artifact removal or validity restoration: what replicates
below is the slice-level effect conditional on the declared surface channel, not the
construct-externality of the channel itself (A1, Section~\ref{sec:condition}). Throughout this
section \emph{surface} is therefore descriptive: a \emph{gain} is the slice-restricted change
$G=A_{\mathrm{slice}}(\srep)-A_{\mathrm{slice}}(s)$, and we test whether these slice-level changes
reproduce under the stated conditioning, while Section~\ref{sec:discussion} asks what they identify
(it exhibits a synthetic case with no artifact loading at all that still yields a large positive
$G$). Appendix~\ref{app:accounting} holds the full evidence accounting: which rows are analyzed,
which are refused, and which count toward each recurring total.

\subsection{Held-out replication with a disjoint-annotator evaluation}
\label{sec:heldout}

To separate method development from evaluation, we froze the complete analysis recipe before any scoring
and applied it once, end to end, to a dataset never previously touched: answer-sentence selection built from SQuAD, with context sentences as candidates and the answer-bearing sentence as positive ($n{=}8{,}000$, 1{,}604 questions). We apply the recipe, features, gates, slice rule, and operator exactly as specified in Section~\ref{sec:method}. The scope of that pre-commitment is precise: the slice-eligibility rule and its decision were committed to a separate artifact before the cross-encoder produced a single score, and the screen and the slice-gain test were then evaluated once, in order, with nothing revised between stages. Table~\ref{tab:heldout} collects the three runs in one view.

\begin{table}[!ht]
\centering\small
\caption{Held-out and disjoint-evaluation runs in one view: same-label slice alignment, the slice gain $G$, the disjoint-annotator evaluation where one exists, and the full-population change. Grouped protocol, query-clustered 95\% CIs, conditional on the fitted pipeline (refit variability $\approx\pm0.003$, Appendix~\ref{app:grouped}). SQuAD and DuoRC ran the frozen pipeline with the router committed before scoring; Natural Questions was constructed after all freezes and is a post-hoc replication (its full-population entry is computed from the released caches, with the frozen recipe reproducing the recorded slice gain to four decimals; change $-0.066$, CI $[-0.079,-0.054]$). DuoRC has no disjoint annotation.}
\label{tab:heldout}
\footnotesize
\setlength{\tabcolsep}{3pt}
\renewcommand{\arraystretch}{1.2}
\begin{tabular}{lcccc}
\toprule
Setting & Slice (raw$\to$res.) & Slice gain $G$ & Disjoint-evaluation gain & Full pop.\ (raw$\to$res.) \\
\midrule
SQuAD (held-out) & $0.706\to0.882$ & $+0.176$ {\scriptsize$[+0.155,+0.197]$} & $+0.155$ {\scriptsize$[+0.135,+0.176]$} & $0.932\to0.836$ \\
DuoRC (held-out \#2) & $0.807\to0.863$ & $+0.056$ {\scriptsize$[+0.038,+0.076]$} & --- & $0.951\to0.867$ \\
Natural Questions (post hoc) & $0.402\to0.658$ & $+0.256$ {\scriptsize$[+0.233,+0.279]$} & $+0.220$ {\scriptsize$[+0.196,+0.244]$} & $0.783\to0.717$ \\
\bottomrule
\end{tabular}
\end{table}

Here $\phi$ is a partial-input feature set carrying a TF-IDF block over one side of the pair alongside the dense cues, so A1 is a stronger assumption for this feature set than it would be for length alone (Section~\ref{sec:samelabel} measures how much of the result that assumption carries), and $y$ is the annotator's answer span, which never sees the cross-encoder. Appendix~\ref{app:settings} reports the run stage by stage. The frozen pipeline passed both pre-adjustment gates: the slice-eligibility check at $\phi$-only AUC $0.813$,
and the screen at $R^2_{\phi\to s}=0.491$ and $\Delta_{\mathrm{slice}}=0.226$. Its committed slice-eligibility check decision, recorded before the cross-encoder scored anything, was that an
informative measurement target is available (Appendix~\ref{app:settings} quotes the file's own wording). The grouped ridge
adjustment then brought the surface correlation from $0.627$ to $0.027$, inside the gate. On the
pre-declared slice, construct alignment rose $0.706\to0.882$, a gain of $+0.176$, while
full-population alignment fell $0.932\to0.836$ (Table~\ref{tab:heldout}; the observed trade-off, taken up
in Section~\ref{sec:discussion}). The outcome is therefore reported as a measured slice gain, with the slice-restricted scope defined in Section~\ref{sec:attrdef}.

The SQuAD validation set provides multiple annotator answer spans, which supply the disjoint evaluation the slice protocol otherwise lacks. Defining everything from the first annotator's span and evaluating only against the disjoint remaining annotators' spans, the gain survives nearly intact: $+0.155$ against $+0.176$ same-label (Table~\ref{tab:heldout}). The slice and the screen statistics were defined on the first annotator's labels, so the headline effect is not attributable solely to reusing the exact annotation instances that define the slice. Inter-annotator agreement is $0.973$, so a conditioning artifact confined to the $2.7\%$ disagreement region would be invisible to the check. And both label sets come from the same annotation task, so what the disjoint evaluation breaks is exact-label reuse, not the broader conditioning induced by defining the slice from the predictor's errors; the stronger A2 assumption---artifact-independent measurement error---remains unverified.

The second frozen-pipeline replication probes a different text domain: answer-sentence selection from DuoRC movie-plot summaries \citep{saha2018duorc} ($n{=}8{,}000$, 594 questions), same frozen pipeline, same pre-scoring commitment ($A(\hat a)=0.834$). The gates pass and the slice improves $0.807\to0.863$ (Table~\ref{tab:heldout}), a smaller gain than SQuAD's, in line with its smaller $\Delta_{\mathrm{slice}}$; DuoRC has no disjoint evaluation.

Because the annotator separation rested on a single setting, we built a second disjoint evaluation from the Natural Questions validation set \citep{kwiatkowski2019nq} ($n{=}4{,}929$ over all $946$ eligible questions), added only after all design choices had been frozen and therefore outside the pre-commitment accounting. Every gate passes, and the five-way annotations leave a wider disagreement region than SQuAD's, making this a disagreement-richer disjoint evaluation: the gain survives at $+0.220$, $86\%$ of the same-label magnitude (Table~\ref{tab:heldout}).

\subsection{The same-label settings, and the dose--response underneath them}
\label{sec:samelabel}

\begin{table}[!ht]
\centering
\small
\caption{Outcomes for the five label-defined slice rows with a measured gain; the six refused rows are in Table~\ref{tab:refused}, and the sixth improved row, HANS lexical overlap, is a post-hoc dataset-defined subset reported in Section~\ref{sec:boundary-refusals}. $A(\hat a)$: the slice-eligibility AUC, gate ${\approx}0.65$. Screen: the two statistics of Section~\ref{sec:condition}. Slice effects are same-label evaluations and optimistically biased (Appendix~\ref{app:slicebias}); gains are computed from unrounded endpoints. $^{\dagger}$ decorrelation marginal under the interval reading. $^{\ddagger}$ the only row here that survives the uniformly strict interval rule. All rows are development or supporting settings; the primary confirmatory observational result is held-out SQuAD (Table~\ref{tab:heldout}). Strict-interval reading: Appendix~\ref{app:rules}; full accounting: Appendix~\ref{app:accounting}.}
\label{tab:scope}
\setlength{\tabcolsep}{4pt}
\renewcommand{\arraystretch}{1.15}
\resizebox{\textwidth}{!}{%
\begin{tabular}{llclll}
\toprule
Setting (score) & $A(\hat a)$ & Screen & Slice or subset effect & Full pop. & Outcome \\
\midrule
\multicolumn{6}{l}{\textcolor{passgreen}{\textbf{Slice improved}}\footnotesize: both pre-adjustment checks pass; gains are confined to the label-defined slice}\\
\addlinespace[1pt]
SICK (DistilBERT-MNLI, prospective) & 0.727 \pass & \pass & $0.765\to0.964$ & $0.967\to0.906$ & \vrep \\
WikiQA (MS MARCO CE, screened) & 0.748 \pass & \pass & $0.622\to0.802$ & $0.864\to0.811$ & \vrep$^{\dagger}$ \\
TriviaQA (MS MARCO CE, screened)$^{\ddagger}$ & 0.676 \pass & \pass & $0.473\to0.621$ & $0.706\to0.636$ & \vrep \\
\addlinespace[2pt]
\multicolumn{6}{l}{\textcolor{passgreen}{\textbf{Slice improved, qualified}}\footnotesize: improvement measured, but treated as supporting rather than primary evidence (Section~\ref{sec:positives})}\\
\addlinespace[1pt]
SNLI (DistilBERT-MNLI, off-dist) & 0.708 \pass & \pass & $0.927\to0.970$ {\scriptsize$(+0.044)$} & $0.958\to0.880$ & \vrep$^{\dagger}$ {\scriptsize(illustrative)} \\
ASNQ (MS MARCO CE, screened) & 0.751 \pass & \pass & $0.495\to0.695$ & $0.851\to0.769$ & \vrep\ {\scriptsize(decorr.\ not met)} \\
\midrule
\multicolumn{6}{l}{\textcolor{failred}{\textbf{Refused}} (six rows)\footnotesize: a named gate fails; Table~\ref{tab:refused} (Appendix~\ref{app:screen}) lists them with the exit each takes}\\
\bottomrule
\end{tabular}}
\end{table}

Further observational settings instantiate the same recipe, four of the five selected before scoring by a computed slice-eligibility check reading or a prospective heuristic. Their outcomes appear in Table~\ref{tab:scope}, their before/after slice outcomes in Figure~\ref{fig:real}, and their per-setting details in Appendix~\ref{app:settings}; we summarize what they add.

\textbf{Coverage.} The positives span both task families, NLI (SNLI, SICK) and QA answer selection (WikiQA, ASNQ, TriviaQA), with slice gains of $+0.044$ to $+0.200$ (HANS lexical overlap is a post-hoc dataset-defined \emph{subset} effect, counted apart; Section~\ref{sec:boundary-refusals}). The observational case rests on SQuAD; these developed settings are corroboration, and their evidential weight under the strict interval rule is in Appendix~\ref{app:rules}. Across settings the gains covary descriptively with the loading and the slice degradation rather than separating by task family (Figure~\ref{fig:repairability}, Appendix~\ref{app:c2}).

Two rows carry qualifiers. SNLI is the illustrative first instance and the weakest: its margin survives independently defined slices at full sample size but not the fully-nested protocol, so no empirical claim rests on it. ASNQ is where index decorrelation \emph{fails} on the point estimate under the question-grouped protocol ($R_{\mathrm{lin}}=0.139$), so we report its slice gain but withdraw the decorrelation claim (Section~\ref{sec:checks}). The failure has a mechanism: row-level folds let candidates of one question span the train/held-out boundary, so $\hat g$ explains $s$ partly through within-question information; grouping the folds removes that component, and the surface correlation it had been masking surfaces in the residual ($0.047\to0.139$).

\textbf{The features behind the QA gains.} Splitting $\phi$ and adjusting with each part alone, on the same fixed slices, locates the QA gains in the question--answer overlap features, the answer-only block contributing little, in all three settings (Appendix~\ref{app:c2} gives the numbers and the protocol). The features that produce the gains are ordinary lexical-overlap and answer-only surface features, and Section~\ref{sec:boundary-deploy} states the A1 caveat this leaves open.

\textbf{The loading--gain relationship.} Across the evaluated scorers, a larger loading is associated with a larger slice gain. The cross-pair
correlation is descriptive only, since the pairs share datasets and slices, and is reported with Figure~\ref{fig:repairability}. The sharper reading holds a single dataset
and slice fixed and varies only the scorer: on WikiQA three scorers with grouped loadings $0.016/0.314/0.360$ gain $+0.045/{+}0.181/{+}0.239$ (Appendix~\ref{app:grouped}; a scorer-ordering illustration is in Appendix~\ref{app:reversal}).

\subsection{Four current-generation judge probes}
\label{sec:judgeprobe}

The scorers above are 2023-generation encoders. Swapping only the scorer on the identical WikiQA rows, features, slice recipe, and protocol, an LLM judge (Claude Haiku 4.5; \citealp{anthropic2025haiku}; temperature $0$) is far stronger where the encoders fail: full-population alignment $0.925$ and raw slice alignment $0.861$, against the cross-encoder's $0.864$ and $0.622$. Its loading on the linear surface channel is an order of magnitude weaker than the two strong scorers on this dataset ($R^2_{\phi\to s}=0.031$ against $0.314$ and $0.360$, though above the weak QNLI scorer's $0.016$) and falls below the gate, so the screen refuses. Residualization past that refusal yields a small positive gain, $0.861\to0.911$ on the slice ($+0.049$).\footnote{The value reported in the main run; a re-run of the frozen recipe on the cached scores reads $+0.050$ with a query-clustered interval of $[+0.036,+0.066]$ (\texttt{judge\_gain\_ci.py}).} Results near the $R^2$ threshold---two conservative near-misses and the QNLI replications
erring in both directions---and why the $0.05$ constant stays as committed are reported in
Appendix~\ref{app:c2threshold}. The measured linear loading is smaller for this judge than for the 2023 encoders, and the same pattern holds across every current-generation probe in this paper: the three code-audit probes read $R^2_{\phi\to s}=0.000$ on the balanced sample (Section~\ref{sec:entangled}), so all four probes fall below the loading gate (the one-view table is in Appendix~\ref{app:c2threshold}). These four probes demonstrate non-universality of the measured linear channel in current-generation judges; they are too few to characterize prevalence.

\subsection{Where the same statistics refuse}
\label{sec:boundary-refusals}

The refusals (Table~\ref{tab:refused}, Appendix~\ref{app:screen}) help characterize the scope of the
procedure, because each fails a \emph{named} gate: the surface channel is too weak to be
associated with degradation in a strong in-distribution classifier (toxicity), the scorer leaves no surface-associated
error to remove (QQP at AUC $0.993$), the benchmark was built to remove the channel (ANLI),
or the channel is present in a form a \emph{linear} operator cannot recover (HellaSwag,
SWAG). Appendix~\ref{app:refusals} reads each setting in turn, and records one
non-instance: in the pairwise representation we use, RewardBench supplies no construct
label independent of the preference \citep{lambert2024rewardbench}. A post-hoc analysis of HANS \citep{mccoy2019hans} further shows that the same operator can help one
heuristic subset while harming two others, so a benchmark-level average would report one figure
consistent with no effect; the readings are in Appendix~\ref{app:settings}.

\section{What the Slice Gains Do Not Establish}
\label{sec:discussion}

Section~\ref{sec:positives} established that slice gains replicate under held-out and disjoint-annotator evaluations; this section states what such gains do and do not support. The supported interpretation is narrower than repair (Section~\ref{sec:attrdef}), and every measured quantity outside the declared slice moves the other way (Section~\ref{sec:costs}).

\subsection{The supported reading: a slice-restricted gain}
\label{sec:attrdef}

\textbf{What the adjustment is for.} Residualization is useful here not because the residual score is
better --- every quantity we measure outside the declared slice falls --- but because the
before--after contrast measures how construct alignment changes on a declared failure population
when the fitted surface-associated component is removed. That diagnostic localises a vulnerability of the
score; it does not identify an artifact-specific causal contribution and does not license replacing
the raw score. Nor is it implied by the pre-adjustment summary
$\Delta_{\mathrm{slice}}$ alone: the ratio $G/\Delta_{\mathrm{slice}}$ runs from below $0$ to $1.38$
across our settings, so a report-only summary misjudges in both directions
(Appendix~\ref{app:baseline}). How far $G$ is from an artifact-specific quantity is easiest to see at
the extreme: in the Gaussian model with $\beta=0$, so that the score carries \emph{no} artifact
loading at all, residualization still raises slice alignment from $0.903$ to $0.987$ while
full-population alignment collapses from $0.973$ to $0.748$ (Appendix~\ref{app:identifiability}).

\textbf{Why a same-label gain is not a validity claim, and what the slice-gain test does instead.}
The slice is defined by the errors of a predictor built from the same measured feature channel
later used to fit the subtracted component, so conditioning on those errors can favor subtraction
even when that component is not artifact-specific; once the screen passes, the slice construction
already favors a positive gain (Property 4, Appendix~\ref{app:identifiability}; the idealization
fixes a sign, not a magnitude). Gains are therefore reported as a \emph{slice-restricted alignment
change}: $G$ is a descriptive before--after contrast under the fitted transformation, not a causal
effect, an additive decomposition of error, or an attribution to an artifact, and a positive $G$
establishes neither construct validity nor deployment benefit, even when every diagnostic threshold
is satisfied. The slice-gain test is accordingly used only to reject adjustments that fail it, and
it does discriminate: it fails two replications that clear every gate
(Appendix~\ref{app:c2threshold}; its behavior in the component ablation is in
Appendix~\ref{app:ablation}).

\textbf{What the gains do rest on.} The strongest observational evidence comes instead from the
held-out, pre-committed replication and the disjoint-annotator evaluations of Section~\ref{sec:heldout}, corroborated by the
screen, the decorrelation diagnostic, and the costs reported beside every gain.

Same-label slice gains carry an upward bias, and a synthetic dual measurement quantifies the
inflation (Appendix~\ref{app:slicebias}); four controls examine it. The disjoint evaluations on SQuAD
and Natural Questions show the gains are not due solely to reusing the exact labels that define the
slice (they do not bound the bias, since the second annotator pool still shares annotation
conventions and the feature channel), and the nested split-half protocol, slice defined on one half
and the adjustment evaluated on the other, preserves every positive (Appendix~\ref{app:slicebias}).
The other two controls, a placebo predictor matched to the surface predictor's AUC and slices refit
on proper sub-blocks of $\phi$, are in Appendices~\ref{app:slicebias} and~\ref{app:crossfamily}.
None of the four removes the conditioning that defining the slice from a predictor's errors
introduces; what they establish is that it is not the whole of the measured gains, which is why the
supported reading stays the slice-restricted one.

\subsection{Measured costs outside the slice}
\label{sec:costs}

\begin{table}[!ht]
\centering\small
\caption{Query-level ranking on slice-touching queries, raw $\to$ residualized (\texttt{grouped\_full.py}).}
\label{tab:ranking}
\begin{tabular}{lccc}
\toprule
Setting & slice queries & MRR & mean per-query AUC \\
\midrule
WikiQA & 286 & $0.789\to0.688$ & $0.864\to0.802$ \\
ASNQ & 375 & $0.699\to0.619$ & $0.874\to0.782$ \\
TriviaQA & 2202 & $0.696\to0.646$ & $0.696\to0.638$ \\
SQuAD & 1067 & $0.876\to0.824$ & $0.901\to0.857$ \\
DuoRC & 424 & $0.857\to0.810$ & $0.937\to0.898$ \\
\bottomrule
\end{tabular}
\end{table}

\textbf{The gains do not make a better ranker.} The slice gains are pooled candidate-level AUCs, and they do not translate into better within-question ranking: restricted to queries touching the slice, MRR declines in every setting under residualization (SQuAD $0.876\to0.824$; WikiQA $0.789\to0.688$; the others likewise, Table~\ref{tab:ranking}). This is the operational-scope limit measured at the task level, and it is what the slice-restricted framing predicts: because $G$ is a pooled, candidate-level quantity computed on a label-defined subset, while MRR ranks within a question over all candidates including those the removal costs, an adjusted score can improve pooled alignment on the declared slice without improving
within-question ranking.

\textbf{The supported claim is evaluative and does not extend to deployment.} The slice on which the gain is measured is label-defined, so per-example membership in it is not observable at deployment time. There is a natural observable substitute: restrict to the subgroup the surface predictor flags ($\hat a>\tau$), which is computable without $y$. It does not improve alignment (Table~\ref{tab:threepop}; every entry declines), because the flagged subgroup mixes examples where the surface predictor is right with
those where it is wrong. A pre-declared family of observable $\phi$-only switching and scale-matched
rules does no better: none improves held-out full-population alignment, and the optimum is the
identity (Table~\ref{tab:deployrules}, Appendix~\ref{app:c2}). What the audit can still support operationally is triage: an observable risk flag that changes
no score but ranks examples for re-annotation (Appendix~\ref{app:baseline}).

\begin{table}[!ht]
\centering\small
\caption{Raw$\to$adjusted construct alignment on three populations, per positive setting: the full population, the label-defined slice (the declared target), and the $\phi$-only observable flagged subgroup $\hat a>\tau$ (the candidate deployment rule). Grouped protocol.}
\label{tab:threepop}
\begin{tabular}{lccc}
\toprule
Setting & Full population & Label-defined slice & Observable flagged ($\hat a>\tau$) \\
\midrule
SNLI & $0.958\to0.880$ & $0.927\to0.970$ & $0.946\to0.895$ \\
SICK & $0.967\to0.906$ & $0.765\to0.964$ & $0.953\to0.930$ \\
WikiQA (CE) & $0.864\to0.811$ & $0.622\to0.802$ & $0.942\to0.835$ \\
ASNQ (CE) & $0.851\to0.769$ & $0.495\to0.695$ & $0.733\to0.695$ \\
TriviaQA (CE) & $0.706\to0.636$ & $0.473\to0.621$ & $0.685\to0.650$ \\
SQuAD (CE, held-out) & $0.932\to0.836$ & $0.706\to0.882$ & $0.912\to0.812$ \\
DuoRC (CE, held-out \#2) & $0.951\to0.867$ & $0.807\to0.863$ & $0.952\to0.835$ \\
\bottomrule
\end{tabular}
\end{table}

\textbf{The full-population cost: mandatory report, not exit condition.} The cost's sign cannot separate the expected trade from destruction---the cost appears in every observational positive here, while the designed audit's pooled AUC is flat by construction ($0.529\to0.530$)---and a numerical threshold on its magnitude would be a hidden assumption about $\rho$, the quantity no observational statistic here identifies (Appendix~\ref{app:identifiability}). The reporting protocol of Section~\ref{sec:protocol} instead mandates the pair, slice gain beside full-population cost; a consumer-side stopping rule on the cost is compatible, and belongs there.

\section{Entanglement and Information Retention}
\label{sec:synthetic}

This section examines two independent limits on what an adjustment can establish. The first is
identification: when construct and artifact are entangled, the observable statistics stop deciding
whether the adjustment is valid (Sections~\ref{sec:entangled} and~\ref{sec:boundary-synth}), which we map with an assignment-based probe on real text and the controlled model that reproduces
its qualitative pattern. The two play different roles: the probe demonstrates the harm, and the
controlled model demonstrates that no pre-adjustment gate can be guaranteed to catch it. The second is the consumer class: linear decorrelation is not closure against a
nonlinear reader (Section~\ref{sec:boundary-nonlinear}).

\subsection{A semi-synthetic assignment probe on real text}
\label{sec:entangled}

Construct--artifact entanglement is the one regime for which we lack a natural real-data setting
with an independent check, so we use a semi-synthetic assignment probe, and what it can show is
bounded by how it is built. The two experiments in this section make
different points. The assignment probe exhibits a harmful regime that the committed screen
\emph{happens} to reject; the controlled model of Section~\ref{sec:boundary-synth} then shows why
that rejection is no evidence that the gates detect entanglement: equally harmful configurations
pass the same gates. The probe induces correlation between a fixed format channel and the construct in a real dataset
and sweeps the collinearity.
The assignment couples format with correctness: at $p{=}0.5$ correct and buggy candidates are
verbose equally often, while as $p\to1$ correct candidates are mostly verbose and buggy ones mostly
terse. A scorer that tracks correctness then becomes statistically predictable from format without
any direct dependence on it, and residualizing on that format removes construct signal along with it. Because label-dependent text edits cannot be certified semantics-preserving, it
holds the edits fixed and varies which already-verified candidates enter the evaluation sample (Appendix~\ref{app:bridge}).
The sweep is controlled by an assignment probability $p$, which induces an empirical
construct--artifact collinearity $\rho$; the two are different parameters, and we report gate
readings on the $p$ grid the procedure actually ran on and damage on the $\rho$ scale it induces.

The sweep runs on the code audit with the audited reward model and three instruction-tuned judges whose paired construct signal is strong ($0.83$ to $0.98$ against the reward model's $0.58$). As the collinearity is swept upward, the predicted damage follows: the adjustment is neutral at $\rho \approx 0$ and costs $0.27$ to $0.33$ of paired win rate at perfect collinearity (problem-level bootstrap intervals exclude zero for every judge from $p{=}0.6$ upward), while the judges' raw performance stays flat across the sweep, so the loss is induced by applying the adjustment under this assignment design. One judge's traversal is tabulated beside the full sweep (Table~\ref{tab:bridgehaiku}).

The diagnostics move the other way. The loading measurement $R^2_{\phi\to s}$ rises from $0.000$ to as much as $0.714$ (Opus 5, Table~\ref{tab:bridge}) purely because artifact and construct were made collinear: on the probe's own $\rho$ scale the screen's $R^2$ gate is open for two of the three judges by $\rho \approx 0.4$ and for all three by $\rho \approx 0.6$, scorers with no detected aggregate format dependence; and the slice gain rises as the damage rises, reaching $+0.128$ where the full population's AUC has fallen by $0.196$.

Running the committed procedure over every cell of the sweep shows what it does here: no cell reaches the post-adjustment checks --- every cell exits at the slice-eligibility check or at the screen, on the observed slice-degradation statistic rather than on any estimate of construct--artifact collinearity (per-cell exits and gate values: Table~\ref{tab:bridgeverdicts}, Appendix~\ref{app:bridge}). In the controlled model of Section~\ref{sec:boundary-synth} a heavily loaded scorer passes every pre-adjustment gate with a large positive slice gain exactly where the adjustment damages the full population (Table~\ref{tab:rhoscreen}), which is why Section~\ref{sec:protocol} treats entanglement as a reporting obligation rather than a detectable condition. The naturally entangled case remains missing; the probe is semi-synthetic because it induces collinearity through candidate sampling (Section~\ref{sec:boundary-deploy}).

\subsection{The controlled model behind the probe}
\label{sec:boundary-synth}

\begin{figure}[!ht]
\centering
\includegraphics[width=\textwidth]{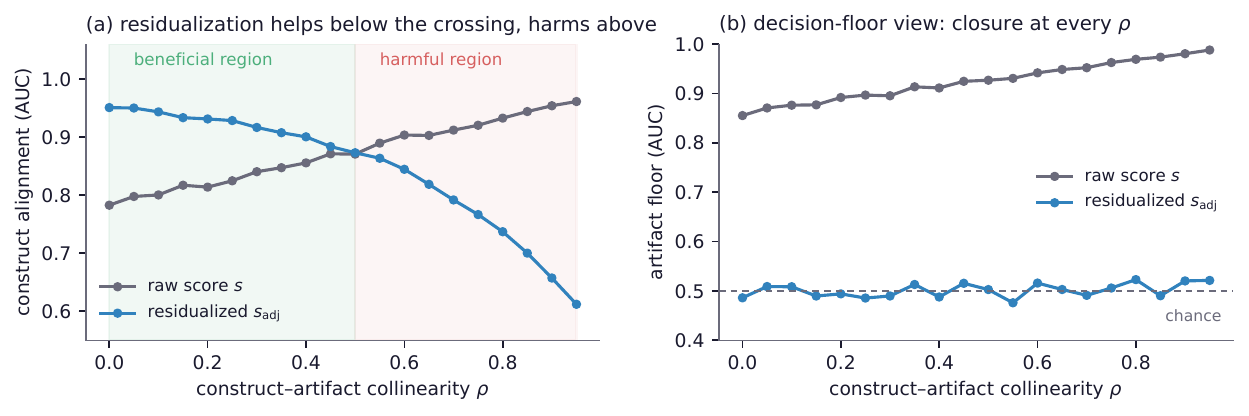}
\caption{Synthetic collinearity sweep ($N{=}4000$ per point; score $=$ construct $+$ $1.2\times$artifact $+$ noise; the validation label thresholds the \emph{latent} construct, so label--artifact independence holds only at $\rho{=}0$). \textbf{(a)} Construct alignment of the raw and residualized score across $\rho$. \textbf{(b)} Decision-level artifact floor: AUC of a probe predicting the artifact decision from the score alone (at $0.5$, artifact predictability has fallen to chance).}
\label{fig:rho}
\end{figure}

A controlled model reproduces and explains this qualitative pattern. Figure~\ref{fig:rho} traces the validity boundary in a model where the construct--artifact
collinearity $\rho$ is known. As $\rho$ grows, the projection of the score onto the artifact
direction increasingly contains construct signal as well, so subtracting that projection removes
both --- which is why the artifact channel can close while construct alignment falls. Cross-fitted adjustment improves construct alignment below the collinearity crossing ($0.783\to0.951$ at $\rho{=}0$) and degrades it beyond ($0.933\to0.737$ at $\rho{=}0.8$, figure values; the Table~\ref{tab:rhoscreen} re-run carries this level, its gain differing by $0.005$), while the artifact floor closes at \emph{every} $\rho$. Channel closure and validity restoration are separate properties here, which is why channel closure and construct alignment must be reported separately; passing the slice-gain test supports only the slice-level claim of Section~\ref{sec:attrdef}. The sweep is a deliberately clean illustration (one linear artifact, Gaussian geometry), and the departures encountered in real data---nonlinear channels, multiple artifacts, and noisy constructs---are stress-tested separately in Appendix~\ref{app:synthstress}, where the same qualitative boundary reappears. This regime is also the boundary case of the reporting protocol: only in this controlled model, where $\rho$ is known by construction, can a setting be declared \emph{non-identifiable from the audit observables}; a real setting's observables do not reveal whether it lies in this regime, and Section~\ref{sec:protocol} states what may be reported there.

In the model, the slice-eligibility check crosses its gate near the collinearity at which full-population adjustment stops helping, so for weakly loaded scorers the damage begins before the slice-eligibility check's gate opens---the pattern the assignment probe measures on three judges (the crossing derivations, the two-parameterization caveat, and the monotonicity assumption behind the ordering claim are in Appendices~\ref{app:identifiability} and~\ref{app:rhoscreen}). The gate therefore does not mark the boundary it would need to mark: on a weakly loaded scorer it opens only once the damage has begun.

\subsection{What linear removal leaves behind}
\label{sec:boundary-nonlinear}

\textbf{Linear scope, and a measured pathology.} Cross-fitted ridge residualization targets the
fitted linear component of the measured channel, and shortcuts a linear operator cannot represent
stay outside its reach (the nonlinear cases, and an oracle recoverability probe that predicts those
refusals in advance, are in Appendix~\ref{app:nonlinear}). The operator also has a measured failure
mode: on SNLI it raises the nonlinear recoverability of the prespecified surface decision from
$0.581$ to $0.827$ rather than destroying it, consistent with relocation into nonlinear structure.
Linear decorrelation therefore does not imply nonlinear closure: passing the diagnostic supports
only a low-linear-association claim for the prespecified index, and where the probe still recovers
the surface decision, the report includes its value. A nonlinear eraser removes this recoverability at the cost of
abandoning the original score scale, and because the same probe detects it, escalating is a diagnosed choice
rather than a default (Appendix~\ref{app:quantile}).

\section{Implications for Auditing: a Reporting Protocol}
\label{sec:protocol}
\label{sec:boundary-deploy}

This section assembles the distinctions of Sections~\ref{sec:rmcode}--\ref{sec:synthetic} into what
an auditor should compute, in what order, and what each outcome permits them to report, as a
conservative reporting procedure; its thresholds are heuristic defaults.

The screen errs deliberately toward refusal and can miss weak but adjustable channels, and refusals
are common rather than nominal: six of the twelve analysis rows fail a named gate, and the checks
catch two more replications afterwards (Sections~\ref{sec:attrdef} and~\ref{sec:boundary-refusals}). Designed interventions take a separate branch whenever paired
label-preserving interventions exist: the contrasts are the measurement target, and the slice-based stages never enter (Section~\ref{sec:rmcode}).

\begin{figure}[!ht]
\centering
\includegraphics[width=\textwidth]{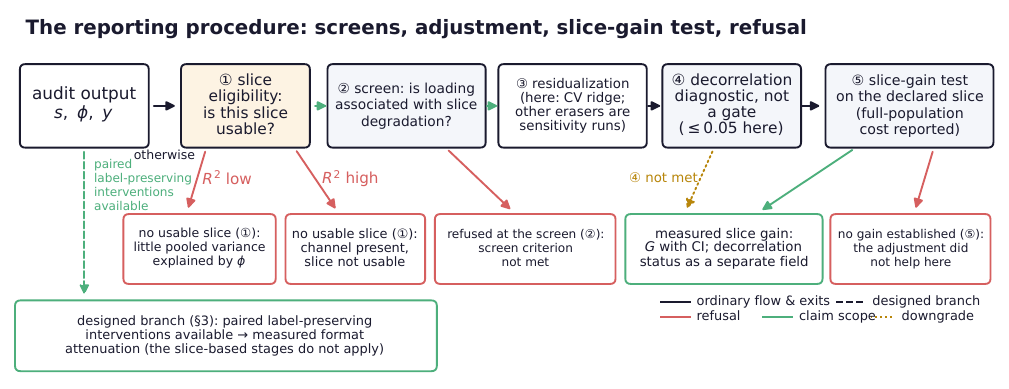}
\caption{The procedure in five stages, consuming what an audit produces ($s$, $\phi$, $y$). The stages apply the statistics of Sections~\ref{sec:condition} and~\ref{sec:checks}; the outcomes are defined in this section; the ordinary path screens a setting before adjustment and tests any resulting slice gain afterwards. The designed branch of Section~\ref{sec:rmcode} enters as a separate fork (dashed): where paired
label-preserving interventions are available they supply the measurement target, so the slice-based stages do not apply and the terminal outcome is measured format attenuation. A failed decorrelation diagnostic (the dotted amber arrow) is recorded as a separate field and does not terminate the slice-gain test.}
\label{fig:workflow}
\end{figure}

One thing the procedure does \emph{not} do is detect entanglement. Where it refuses a harmful
adjustment it does so on the judges' undamaged slice alignment, not on any reading of $\rho$, and in
the controlled model a heavily loaded scorer passes every pre-adjustment gate beyond the crossing
(Section~\ref{sec:synthetic}). If external evidence establishes construct--artifact entanglement,
adjustment should be refused; without such evidence a passing run warrants only the mandatory dual
report, the slice gain beside its full-population cost, never certification of a repaired score.
Figure~\ref{fig:workflow} draws the procedure and Algorithm~\ref{alg:procedure} states it as
pseudocode, including the stage ordering that keeps every post-adjustment outcome uninspected until
the decision to adjust is made.

\begin{table}[h]
\centering\small
\caption{The four outcomes of the observational branch and the claim scope associated with each. A refusal
withholds an adjustment the checks cannot vouch for.}
\label{tab:verdicts}
\begin{tabular}{@{}lp{0.66\textwidth}@{}}
\toprule
Outcome & What it permits \\
\midrule
\emph{No usable slice} & The slice eligibility check fails, or the slice degenerates: either way the slice cannot serve as the object the gain is measured on. \\
\emph{Refused at the screen} & A pre-adjustment gate fails, so no adjusted score is tested on this setting. \\
\emph{No gain established} & The gain test runs and its interval does not lie above zero: this operator, with these features, on this sample, did not demonstrably help. \\
\emph{Measured slice gain} & The gain test passes; the claim is scoped to the declared slice and is reported beside the full-population cost. \\
\bottomrule
\end{tabular}
\end{table}

\textbf{Outcomes.} Table~\ref{tab:verdicts} lists the four terminal outcomes of the ordinary observational branch and
the claim scope associated with each; an outcome's scope is what the reported number permits one to
say. A designed intervention provides a separate terminal outcome,
\emph{measured format attenuation} (Section~\ref{sec:rmcode}).

Beside those outcomes, the index-decorrelation diagnostic is reported as a separate
field, met or not met, rather than as a grade on the gain: a failure at stage
\textcircled{\footnotesize 4} does not end the run, and a slice gain reported with the
criterion unmet carries the same descriptive scope as one reported with it met (the path
ASNQ takes, Table~\ref{tab:scope}). Which diagnosis a no-usable-slice setting carries, and how it
is read from $R^2_{\phi\to s}$, is stated with the pseudocode in Appendix~\ref{app:procedure}.

The stages do not contribute equally: on observational data the selection work is done by the
screen and the slice-gain test, while the slice-eligibility check's contribution is measurable only over the candidate pool (component
ablation, Appendix~\ref{app:ablation}; Table~\ref{tab:threshold}, Appendix~\ref{app:threshold}),
and it remains the conservative selection heuristic of Section~\ref{sec:condition}, not a detector
of harmful adjustment.

\textbf{What to report.} A supported slice-level diagnostic is reported as a pair, the slice gain beside
the full-population cost, never the gain alone (Section~\ref{sec:costs}); the index-decorrelation diagnostic is reported beside it as a separate field, met or not met; and where relocation is measured, the nonlinear-probe value accompanies the report (Section~\ref{sec:synthetic}).

\textbf{Evidence required before reporting a residualization-based diagnostic.} Application of an erasure-style operator to an evaluation metric should be gated on a
provenance-separated construct measurement, which requires no additional labels when $\phi$ and $y$
are already part of the audit. The reporting logic is operator-agnostic in principle, but our
empirical validation centers on cross-fitted residualization; the alternative erasers here are
sensitivity analyses rather than independently validated deployments of the full procedure. Decorrelation cannot carry that burden, as the synthetic sweep shows. No single statistic establishes validity. The slice-gain test is a necessary check; in the
observational regime the strongest evidence instead comes from the held-out, pre-committed
replication and the disjoint-annotator evaluations (Section~\ref{sec:positives}).

\textbf{Requirements on the construct measurement.} Labels are needed to fit the surface predictor, define the slice, and evaluate alignment, at the scale of hundreds to low thousands of items; a new example's label is not needed at deployment time. The gain test was robust in its outcome, though not in its magnitude, to the tested rates of independent label corruption: the slice-gain test is a differential comparison, and in a synthetic experiment where the auditor recomputes the slice from the noisy labels the outcome survives $30\%$ flips (Appendix~\ref{app:multi}). Its measurement error must, however, be \emph{independent} of the artifact, the one property whose failure biases the comparison itself (Appendix~\ref{app:corrnoise}). We cannot verify that statistical independence from observational data. We therefore use
label provenance as an auditable risk indicator, and report shared-convention risk
explicitly; provenance is a proxy for independence, not proof of it.

\section{Related Work}
\label{sec:related}

\textbf{Concept removal and robust training.} INLP \citep{ravfogel2020inlp}, RLACE \citep{ravfogel2022rlace} and LEACE
\citep{belrose2023leace} remove concepts from model \emph{representations}; we apply the
same projectional idea to a scalar evaluation score and ask whether the transformed
score still measures its intended construct. Earlier work showed that attributes can remain
recoverable after apparent removal, whether by a different adversary
\citep{elazar2018adversarial} or by geometry the removal never touched
\citep{gonen2019lipstick}. On a scalar score the phenomenon we measure differs in
direction: the linear channel does close (SNLI linear recovery $0.627\to0.451$, near chance with the fitted orientation reversed), while
a flexible probe reads the surface decision \emph{better} than before ($0.581\to0.827$), because the
subtraction imprints its own estimate into the residual. Appendix~\ref{app:quantile} tests a
distributional closure operator against it.
Appendix~\ref{app:corrnoise} explains why LEACE itself degenerates on a scalar score.
The operator is classical partialling-out \citep{frisch1933partial,lovell1963seasonal},
cross-fitted in the style of double machine learning \citep{chernozhukov2018double}.
Related training-time methods use residual fitting or product-of-experts objectives to
reduce reliance on bias-only predictions \citep{he2019drift,clark2019dont}. Those
methods target model robustness, whereas ours is a post-hoc measurement procedure for
frozen or externally supplied scores. Their observed trade-off between
out-of-distribution robustness and in-distribution accuracy
\citep{utama2020tradeoff,mahabadi2020endtoend} motivates our requirement to report the
full-population cost alongside any slice gain.

\textbf{Post-hoc metric debiasing.} Length-controlled AlpacaEval \citep{dubois2024length},
motivated by length dependence in reward models \citep{singhal2023long}, is the closest
operation in current practice: it adjusts an LLM-judge win rate for response length. We
instead treat adjustment as one stage in a procedure that can test or refuse it, and
Section~\ref{sec:rmcode} shows why scalar correction can be insufficient.
Constraint-based control can fail in the same way: \citet{chen2026judgevalidity} find
that constraining preferred responses to be no longer than rejected ones does not remove the length confound in RewardBench; it inverts it. The same work separates construct-preserving from construct-changing edits, the distinction our designed audit instantiates with unit-test labels. Chatbot
Arena's style control already extends regression adjustment beyond length to markdown
headers, bold text, and lists \citep{li2024style}, and Arena-Hard-Auto adopts the same
multivariate strategy \citep{li2025arenahard}. These methods broaden the adjusted channel; what they do not report is a test of the adjusted score against a provenance-separated construct measurement. 

\textbf{Relation to the closest prior work.} In pairwise judging, \citet{xu2026debias} already supply a trusted-anchor gate, a paired rendering design, and the observation that a correction turns from helpful to harmful as the construct--covariate correlation rises; our gate and paired design are the pointwise analogues of that work, and we introduce no new estimator. What neither that work nor the operator literature settles, and what the measurements here show, is that once a pointwise score has been residualized, surface-effect removal, conditioned slice alignment, construct validity, and deployment utility are distinct claims that come apart on the same data --- one can improve while the others degrade --- and the held-out and disjoint-annotator evidence together with the controlled identifiability boundary characterize what evidence each claim requires. The reporting protocol of Section~\ref{sec:protocol} is that characterization made operational.

\textbf{Other post-hoc approaches.} From the consumer side, \citet{lamparth2026substitution} show that
single-axis mitigation can redirect optimization toward correlated proxies even as audit
statistics improve; accordingly, the decorrelation claim here is restricted to linear consumers of the score, and the slice-alignment claim remains evaluative rather than a license to optimize against the adjusted score. Other post-hoc approaches include reward calibration
\citep{huang2024posthoc}, regression-based calibration of judge scores toward human
ratings \citep{sahoo2026quantjudges}, nonlinear length decoupling
\citep{zhao2025biasfitting}, confounder-aware judge aggregation \citep{zhao2026care},
cross-fitted style residualization for distillation \citep{wu2026lightning}, and
empirical comparisons of mitigation strategies across judges \citep{soumik2026judging}.
 Conditioning is another alternative to subtraction; the conditional-quantile eraser in Appendix~\ref{app:quantile}
implements its continuous analogue at the cost of calibrated score scale.

\textbf{Construct validity and measurement invariance.} Work on measurement validity in
NLP and the social sciences \citep{jacobs2021measurement,cronbach1955construct},
including reviews of LLM benchmarks \citep{measuring2025matters}, diagnoses when
observed scores fail to represent their intended constructs. The artifact evidence such
audits run on comes from partial-input and shortcut analyses---hypothesis-only baselines,
heuristic subsets, judge-bias surveys, and their known failure modes
\citep{gururangan2018annotation,poliak2018hypothesis,mccoy2019hans,geirhos2020shortcut,zheng2023judging,feng2019misleading}. Together they provide the
partial-input features and provenance-separated construct measurements our procedure consumes; we add a decision about what an adjusted score may be claimed to show. Automated
slice-discovery methods such as Domino \citep{eyuboglu2022domino} identify coherent
subsets on which a model fails; our problem is downstream of discovery: it tests an adjustment on a slice already declared, and leaves the search for slices to those methods.
Psychometrically, this is a question of measurement invariance
\citep{meredith1993invariance,millsap2011invariance}, with differential item functioning
as the classical conditional case \citep{holland1993dif}. Our construct check is a
coarse, two-population analogue: unlike latent-variable approaches, it supports only
differential comparisons within the evaluated configuration, and the full-population
cost records the invariance violation introduced by adjustment.

\section{Limitations}
\label{sec:scopelimits}

\textbf{Current applicability.} For any scorer with a prespecified surface feature channel
and a provenance-separated construct measurement, the two pre-adjustment checks cost two cross-fitted
regressions per evaluation set. Where encoder-era scorers remain in production (rerankers, deployed classifiers, cached-score analyses, reward-model checkpoints of the audited generation) the full procedure applies as demonstrated; the dual-report requirement and the risk-flag triage are scorer-agnostic; and for designed audits the recipe is directly reusable: mechanical labels, paired semantics-preserving edits, and a measurement target supplied by design.

\textbf{Other limits.} The observational slice positives are all English sentence-pair tasks scored by small encoder models, and our screens suggest strong \emph{linear} artifacts are rarer than the literature's nonlinear partial-input results imply: on naturally occurring preference data the reward-model length bias is weak or absent as a linear channel (Appendix~\ref{app:screen}). The format attenuation does not transfer predictably across edit templates: on three
held-out templates the same operator overshoots one, attenuates another, and amplifies a
third (Appendix~\ref{app:rmcode}). Both audited reward-model checkpoints are 2023-generation encoders; the LLM-judge probes are four judge$\times$setting probes spanning three judge models and two
evaluation designs, all falling below the committed loading gate (Section~\ref{sec:judgeprobe}),
enough to demonstrate non-universality but insufficient to characterize prevalence. Every setting scores ranking or classification of \emph{given} text, so retrieval over open corpora and free-form generation are untested and would require rethinking what a partial-input $\phi$ is. The decorrelation diagnostic is dataset-specific and should be re-run per evaluation set.

\textbf{What we have not established.} There is no independently validated case here of residualization improving construct validity on real data where construct and artifact are \emph{entangled}. The intervention audit establishes attenuation of the measured format contrasts in a design where the channel is label-orthogonal by construction; the observational regime supports the slice-restricted scope of Section~\ref{sec:attrdef} and nothing wider; the entangled region is covered by the controlled sweep and the semi-synthetic probe of Section~\ref{sec:entangled}, and by nothing drawn from natural data. That probe induces its collinearity by sampling, so what it delivers is a bound on the procedure and no validation of the procedure here. Nor does the paper establish A1 where it matters most for the observational half. The QA gains rest on the question--answer overlap block (Section~\ref{sec:samelabel}), and a sentence that reuses the question's words is a plausible carrier of relevance as well as of surface form. The rule-based intervention of Appendix~\ref{app:c2} shows the raw score following manipulated overlap; it does not show that overlap is construct-external, so the QA positives remain conditional on A1 for that block.

\section{Conclusion}
\label{sec:conclusion}

What exactly has been shown to improve? In the designed code audit, attenuation of the measured
format channel: paired format effects shrink to estimates whose intervals cover zero while the
correctness margins stay essentially unchanged (equivalence not claimed), and every tested
length-only correction fails to do the same. On three held-out edit templates the same operator
overshoots one, attenuates another, and amplifies a third, which bounds that result to the edit
style it was fit on.

In the observational settings, alignment on a declared failure slice: gains that replicate under a frozen pipeline and survive disjoint-annotator evaluations---and nothing else, since full-population alignment, the observable flagged subgroup, and within-question ranking all decline, so the supported reading is a slice-restricted gain, not repair.

At high collinearity, slice gains coexist with full-population losses, and the committed statistics cannot tell entanglement from the observed diagnostics alone; the reporting response is refusal where a gate catches the case, and the mandatory dual report where none does.  The adjusted score is an audit-time companion to the raw score; the observable-subgroup test that would support replacing it fails. The reporting protocol distilled from these results exists so that the next reported adjustment carries its evidence with it.

\subsubsection*{Reproducibility statement}
All analyses run CPU-only on cached scores and are seeded. The cached scores were produced by public checkpoints and, for the judge probes, an API; no base scorer is trained or fine-tuned, and the fitted auxiliary objects are lightweight scikit-learn models---the operator's cross-fitted ridge, the logistic surface and flag probes, the MLP and random-forest diagnostic probes, the isotonic/spline/boosted length-only baselines, the random-forest residualizer of Appendix~\ref{app:nonlinear}, and the conditional-quantile eraser (5-fold cross-fitting, question-grouped for the QA settings and shuffled otherwise; seed 0; bootstrap 1{,}000 resamples). Full checkpoint, dataset and robustness-suite details are in
Appendix~\ref{app:settings}. Code, pre-commitment artifacts and cached scores are at \url{https://github.com/wdi1024/residualization-audit}. We release the operator, the decorrelation and slice-gain protocol, all per-setting scripts, the robustness suites behind Appendices~\ref{app:screen}--\ref{app:synthstress} (candidate screen, threshold sweep, residualizer/fold/noise/predictor variants, nonlinear probes, nonlinear and multi-artifact synthetics, construct-noise checks, the screen-versus-gain sweep of Table~\ref{tab:rhoscreen}, the decorrelation confidence intervals), the four-template generalization run of Appendix~\ref{app:rmcode}, the mechanism
measurements (the SNLI relocation decomposition, the ASNQ slice-conditional correlations, and the assignment-based entangled-regime probe), and cached scores sufficient to reproduce every number without API access (including the LLM-judge probe, whose scores are cached; only re-scoring from scratch would require the judge API).

\subsubsection*{Broader Impact Statement}
This work concerns the measurement layer of machine-learning evaluation: it diagnoses, and where the checks support it adjusts, scores that steer routing, deferral, and benchmark claims. The intended impact is protective: artifact-confounded scores are today read as semantic evidence, and the procedure either decorrelates a prespecified index under a measured criterion with a slice-scoped claim, or refuses and says why. An ``adjusted'' score could be over-trusted; the measured criterion, the slice-scoped claim, and the refusal mechanism exist precisely to bound what it warrants, and the post-adjustment checks are part of the procedure itself. And the machinery could be misapplied to launder an invalid score by skipping the checks; we release the full protocol so that a reported adjustment without its criterion and claim scope is detectable as such.

\bibliography{refs}
\bibliographystyle{tmlr}

\appendix

\section{The Reporting Procedure, Stated Operationally}
\label{app:procedure}

Algorithm~\ref{alg:procedure} states the procedure of Section~\ref{sec:protocol} as pseudocode,
with the stages and their order of Figure~\ref{fig:workflow}. The order matters: $\hat g$ is fitted
first because the screen needs $R^2_{\phi\to s}$, but the adjusted score is formed only after both
pre-adjustment gates have passed, so no post-adjustment outcome is inspected before the decision to
apply the adjustment is made. Which diagnosis a no-usable-slice setting carries is read from
$R^2_{\phi\to s}$, which separates ``the fitted feature model explains little pooled score
variance'' from ``the channel is there but the slice cannot show it, so supply a measurement target
by design'' (Section~\ref{sec:rmcode}).

\begin{algorithm}[t]
\caption{The reporting procedure, stated operationally. Inputs are what an audit already produces on one
evaluation set. Gate constants are the implementation defaults of
Section~\ref{sec:condition}; the outcomes are those defined in Section~\ref{sec:protocol}.}
\label{alg:procedure}
\begin{algorithmic}[1]
\Require score $s$, partial-input features $\phi$, provenance-separated construct measurement $y$,
groups $g$ (question or problem id)
\Ensure an outcome, and on one of them an adjusted score $\srep$ with a scoped claim
\State \textbf{Readout.} \textbf{if} paired label-preserving interventions are available \textbf{then return} \emph{measured format attenuation}:
 report the paired format effects (raw vs.\ residualized, clustered CIs) and the construct-margin
 change; a stronger removal claim requires a pre-declared equivalence tolerance
 (Section~\ref{sec:rmcode})
 \Comment{designed branch: the contrasts are the measurement target, so the slice-based stages below do not apply}
\Statex
\State \textbf{Observational branch.} $\hat a \gets \textsc{CrossFit}(\phi_a \to y;\, g)$
  \Comment{surface predictor, out-of-fold}
\State $\tau \gets \textsc{Quantile}(\hat a,\, 1-\bar{y})$;\;
       $\mathrm{slice} \gets \{x : \mathbf{1}[\hat a(x)>\tau] \neq y(x)\}$
\State \textbf{if} the slice does not contain both label classes \textbf{then return} \emph{no usable slice} (slice undefined)
       \Comment{$A_{\mathrm{slice}}$ undefined; the $p{=}1$ column of Table~\ref{tab:bridgeverdicts} is this case}
\State $\hat g \gets \textsc{CrossFit}(\phi \to s;\, g)$;\;
       $R^2_{\phi\to s} \gets 1 - \|s-\hat g\|^2/\|s-\bar s\|^2$
\Comment{the surface component is fitted here; the adjusted score is not formed yet}
\Statex
\State \textbf{Slice-eligibility check.} \textbf{if} $A(\hat a) < 0.65$ \textbf{then return} \emph{no usable slice}:
       little pooled variance explained if $R^2_{\phi\to s}$ is low (a within-problem channel can
       still be present, Appendix~\ref{app:rmcode}); channel present, measurement target missing if high
\State \textbf{Screen.} \textbf{if} $R^2_{\phi\to s} < 0.05$ \textbf{or}
       $A(s)-A_{\mathrm{slice}}(s) < 0.03$ \textbf{then return} \emph{refused at the screen}
\Statex
\State $\srep \gets s - \hat g$
\Comment{both pre-adjustment gates have passed; only now is the score adjusted}
\State \textbf{if} $|\mathrm{corr}(\srep,\hat c)| > 0.05$ \textbf{then} withdraw the decorrelation claim
       and record the criterion as \emph{not met}; the outcome below is unchanged \Comment{index decorrelation}
\State \textbf{if} the clustered bootstrap interval of $G=A_{\mathrm{slice}}(\srep)-A_{\mathrm{slice}}(s)$ does not lie entirely above zero \textbf{then
       return} \emph{no gain established} \Comment{slice-gain test on the declared slice}
\State \textbf{return} \emph{measured slice gain}: report
       $G = A_{\mathrm{slice}}(\srep)-A_{\mathrm{slice}}(s)$ beside the
       full-population change $A(\srep)-A(s)$; the pair is an audit-time measurement, with the index-decorrelation field from line 9 reported beside it
\end{algorithmic}
\end{algorithm}

\clearpage

\section{Refusals, Setting by Setting}
\label{app:refusals}

This appendix reads each refused setting of Table~\ref{tab:refused} in turn, and records
the one candidate that the pointwise setup does not instantiate.

Toxicity's identity-term artifact is too weak to be associated with measurable degradation in a strong in-distribution classifier. QQP's scorer is evaluated in distribution at AUC $0.993$, leaving no surface-associated error to remove. ANLI was adversarially built to remove the hypothesis-only artifact, so no linear hypothesis-only channel is detected \citep{nie2020anli}. HellaSwag and SWAG carry their famous ending-only artifacts \citep{zellers2018swag,zellers2019hellaswag} in a form a \emph{linear} operator cannot recover; the documented partial-input results there come from finetuned BERT probes.

\textbf{Three outcomes inside one benchmark.} HANS \citep{mccoy2019hans} decomposes NLI failures into three heuristics, and the gate statistics make a per-heuristic prediction. Lexical overlap is a linear, dense artifact misaligned with the construct there, and the adjustment helps ($0.592\to0.697$; the gain, $+0.104$, is computed from the unrounded endpoints $0.5925\to0.6968$). The constituent and subsequence heuristics are syntactic, the raw score's surface loading vanishes on those subsets, and the adjustment hurts ($-0.064$ and $-0.200$). The three signs line up after the fact with the subset-level correlation of the subtracted component with the construct ($-0.28$/$-0.01$/$+0.31$), a correspondence that describes the outcomes without predicting them; Appendix~\ref{app:settings} gives the reading, including the saturation reading of constituent's $-0.064$. A HANS-level average would fold the $+0.104$ lexical-overlap gain together with the $-0.064$ and $-0.200$ losses on the syntactic heuristics into a single figure consistent with no effect. Because that correlation is measurable per subset, the one benchmark yields three distinct outcomes instead of one average.

\textbf{A non-instance by construction.} Reward-model margins are a natural target, but in the pairwise representation we use, RewardBench supplies no construct label independent of the preference, since the chosen response is preferred by construction, so the pointwise setup here is not instantiated on it \citep{lambert2024rewardbench}; a representation with randomized pair orientation and an added construct label would be a different instantiation. The requirement falls on the evaluation design: every check above needs an independent $y$, and the chosen-minus-rejected margins we would have scored supply none. Whether a particular subset could supply one under a different representation is a per-subset question of label provenance we do not settle here.

\section{Evidence Accounting}
\label{app:accounting}

Table~\ref{tab:accounting} fixes the accounting of evidence units (table rows, scorer pairs, pre-scoring selections, and out-of-table reports) used throughout the paper. The seven positives of Section~\ref{sec:checks} are the five slice positives of Table~\ref{tab:scope} plus the two held-out replications (SQuAD, DuoRC), which sit outside the twelve analysis rows along with the designed audits. SNLI is among the five only as a denominator for the decorrelation census; it is marked illustrative in Table~\ref{tab:scope} and we treat it as illustrative only.

The held-out replications sit outside the twelve analysis rows because their pipelines were frozen and applied once, so pooling them with the developed settings would misstate what each contributes. The designed reward-model audits sit outside for a different reason: their screen outcome is fixed by the construction, so counting them among the screen's predictions would double-count the design.

Table~\ref{tab:precommit} lists those six one by one. Five matched outright and one, ASNQ, matched only in the coarse sense recorded there. Nothing else in the paper was predicted in advance; the post-hoc row of Table~\ref{tab:accounting} lists what was not.

\textbf{Supporting and illustrative evidence.} Everything outside the three prespecified endpoints
below is \emph{supporting} --- replications and stress tests that corroborate them --- or
\emph{illustrative}, as SNLI is marked in Table~\ref{tab:scope}. The refusals are results in their
own right.

One statistical note applies throughout the paper: roughly $170$ interval estimates appear across the main text and appendices (setting--scorer--gate combinations and robustness variants), each per-quantity, with no formal multiple-comparison correction. Our main conclusions rely on three prespecified endpoints: the two paired format effects of the designed
audit (Table~\ref{tab:rmcode}), the slice gain of the held-out SQuAD run under its
disjoint-annotator evaluation (Table~\ref{tab:heldout}), and the full-population change across
the pre-specified collinearity levels of the sweep (Figure~\ref{fig:rho}); the remaining
intervals are sensitivity analyses rather than additional tests. The mitigations are
otherwise procedural: gates fixed in advance, screens committed before scoring where
possible, and no main conclusion resting on a single marginal interval. The SICK interval in particular establishes little beyond sign, and is weighted accordingly in Appendix~\ref{app:slicebias}.

\begin{table}[h]
\centering\small
\caption{Evidence accounting: what is counted where.}
\label{tab:accounting}
\begin{tabular}{p{0.30\textwidth}cp{0.52\textwidth}}
\toprule
Unit & Count & Members \\
\midrule
Analysis rows (Table~\ref{tab:scope}) & 12 & Positives: SNLI, SICK, WikiQA, ASNQ, TriviaQA, HANS lexical overlap. Refusals: HANS constituent$+$subsequence (one joint row), MNLI, Toxicity, QQP, ANLI, HellaSwag$+$SWAG (one shared row). \\
(setting, scorer) pairs (Figure~\ref{fig:repairability}) & 13 & The slice-bearing rows with cached scores (SNLI, SICK, TriviaQA, MNLI, QQP, Toxicity) plus scorer replications: WikiQA $\times$\{MiniLM, QNLI, STS-B\}, ASNQ $\times$\{MiniLM, QNLI, STS-B, ELECTRA\}. \\
Selected before scoring & 4 & SICK (prospective heuristic); WikiQA, ASNQ, TriviaQA (computed router readings). \\
Scorer replications outside Figure~\ref{fig:repairability} & 2 & TriviaQA $\times$ \{STS-B, QNLI\}, reported in Section~\ref{sec:positives} and Appendix~\ref{app:grouped} and counted in the ablation of Table~\ref{tab:ablation}. \\
Held-out frozen-pipeline replications (reported in text, outside Table~\ref{tab:scope}) & 2 & SQuAD, DuoRC. \\
Designed reward-model audits (outside Table~\ref{tab:scope}; screen outcome follows from the construction) & 2 & MBPP, HumanEval. \\
\midrule
Predictions committed before scoring & 6 & The four screen selections and the two held-out replications; all six outcomes matched the committed (coarse-grained) predictions. \\
Post-hoc analyses (no advance prediction) & --- & HANS (both rows), scorer replications beyond each setting's primary scorer, the LLM-judge probe, the Natural Questions disjoint-evaluation setting, and the appendix robustness suites. \\
\bottomrule
\end{tabular}
\end{table}

\section{Component Ablation of the Procedure}
\label{app:ablation}

Table~\ref{tab:ablation} measures what the procedure buys over reporting $R^2_{\phi\to s}$ and $\Delta_{\mathrm{slice}}$ and stopping. It answers that for the screen and the slice-gain test, and its population is why it cannot answer it for the slice-eligibility check. The sixteen units are the ten eligibility-valued rows of Table~\ref{tab:scope} together with six scorer replications whose gate readings and gain intervals are on record, and every one of them reached the analysis table, so on this comparison set the two gates overlap: the settings the slice-eligibility check excludes are excluded by the screen as well. The slice-eligibility check's own contribution is measurable only where it does the excluding, over the candidate pool, and Table~\ref{tab:threshold} scores it there, where it separates the pool's accepted candidates from its rejections. The comparison below is therefore between the two components downstream of that routing. Two exclusions apply. The HANS rows are post hoc with no slice-eligibility check reading. WikiQA $\times$ STS-B lacks the \emph{grouped} $\Delta_{\mathrm{slice}}$ its screen outcome would need (its loading passes, was never computed before the freeze, and the row-level value that exists, Table~\ref{tab:c2}, is what Figure~\ref{fig:repairability} plots for that pair), while WikiQA $\times$ QNLI needs no $\Delta_{\mathrm{slice}}$ reading at all, its loading alone settling the screen; that asymmetry is why the two replications are treated differently. The false approvals are ASNQ (QNLI), $+0.007$ $[-0.011,+0.026]$, and TriviaQA (QNLI), $-0.014$ $[-0.030,-0.001]$; the false refusal is WikiQA (QNLI), $+0.045$, refused at $R^2_{\phi\to s}=0.016$. The screen alone approves ten units and is wrong three times: it approves ASNQ and TriviaQA scored by QNLI, whose gain intervals cover and fall below zero respectively, and it refuses WikiQA scored by QNLI, which gains $+0.045$. \emph{Adding the slice-eligibility check changes none of those decisions}, because on this population every setting the slice-eligibility check would refuse is one the screen refuses already. The slice-gain test removes both false approvals and leaves the false refusal, which is the cost of the $R^2_{\phi\to s}$ gate discussed above. One accounting caveat: a false approval is defined by the same gain interval the slice-gain test thresholds, so the zero in the slice-gain row is definitional rather than an independent test of the check; the informative content of the table is what the screen alone gets wrong, and what refusals cost. On observational data, then, the selection work is done by the screen, with the slice-gain test acting as the corrective described above. The slice-eligibility check does not change decisions in this comparison set. The slice-eligibility check asks whether a slice exists to validate on, and its answer routes the setting to a diagnosis before any outcome is reached: where it fails with a low loading the fitted feature model explains little pooled variance, the diagnosis for MNLI and Toxicity, and where it fails with a high loading the channel is present and the measurement target is missing, the diagnosis for the designed audit of Section~\ref{sec:rmcode}, the branch that yields its central result. Where both screen statistics are also on record the diagnosis reads them jointly: QQP fails the slice-eligibility check with a high loading but a null slice degradation, so its label is \emph{no slice degradation}, the channel present but costing nothing, rather than \emph{measurement target missing} (Table~\ref{tab:scope}). Appendix~\ref{app:rhoscreen} traces that split across the collinearity sweep. The disagreement between the two tables about the slice-eligibility check comes from the different sets they are asked about.

\begin{table}[!ht]
\centering\small
\caption{The six predictions committed before scoring, one row each: what fixed the commitment, what was predicted, and what the run returned. ``Coarse'' in the last column marks the one outcome that matched only at the level of the predicted improvement: ASNQ's slice gain was measured while its decorrelation claim was withdrawn under the stricter protocol. $A(\hat a)$ values (the router statistic) are the screen-time (row-level) readings the commit artifacts record, with the later grouped recomputation in parentheses. Commit evidence is the released development artifact named in the chronology of Appendix~\ref{app:settings}; it is not third-party timestamped.}
\label{tab:precommit}
\setlength{\tabcolsep}{4pt}
\begin{tabular}{p{0.15\textwidth}p{0.27\textwidth}p{0.22\textwidth}p{0.26\textwidth}}
\toprule
Setting & What was committed, and when & Prediction & Outcome \\
\midrule
SICK & Prospective heuristic, fixed with the first NLI settings before any QA work & Screen-positive & Gain measured, slice $0.765\to0.964$ \\
WikiQA & Router computed before the scorer ran (screen-time $A(\hat a)=0.747$; grouped recompute $0.748$) & Screen-positive & Gain measured, slice $0.622\to0.802$; decorrelation marginal under the interval reading \\
ASNQ & Router computed before the scorer ran (screen-time $A(\hat a)=0.747$; grouped recompute $0.751$), candidate-structure decision disclosed alongside both screen readings & Screen-positive & Slice gain measured, $0.495\to0.695$; decorrelation claim later withdrawn (coarse) \\
TriviaQA & Router computed before the scorer ran (screen-time $A(\hat a)=0.675$; grouped recompute $0.676$) & Screen-positive & Gain measured, slice $0.473\to0.621$ \\
SQuAD & Entire pipeline frozen and the committed router decision written to a machine-readable artifact before scoring & Screen-positive & Gain measured, slice $0.706\to0.882$; $+0.155$ against a disjoint annotator set \\
DuoRC & Pipeline frozen and router committed before scoring ($A(\hat a)=0.834$) & Screen-positive & Gain measured, slice $0.807\to0.863$ \\
\bottomrule
\end{tabular}
\end{table}

\begin{table}[!ht]
\centering\small
\caption{Component ablation over sixteen units: the ten router-valued rows of Table~\ref{tab:scope} (SICK, WikiQA, TriviaQA, SNLI, ASNQ, MNLI, Toxicity, QQP, ANLI, HellaSwag/SWAG) and six scorer replications (ASNQ $\times$ \{STS-B, ELECTRA, QNLI\}, TriviaQA $\times$ \{STS-B, QNLI\}, WikiQA $\times$ QNLI). Nine units help, four harm, three are null by their gain intervals. ``Approves'' counts units the rule approves; a false approval is an approved unit whose gain interval does not lie above zero, a false refusal a withheld unit whose interval does. Exclusions (the two HANS rows and WikiQA $\times$ STS-B) and the false-approval and false-refusal instances are given in the text of this appendix. Grouped protocol throughout.}
\label{tab:ablation}
\begin{tabular}{lccc}
\toprule
Rule & Approves & False approvals & False refusals \\
\midrule
Screen alone ($R^2_{\phi\to s}$ and $\Delta_{\mathrm{slice}}$) & 10 & 2 & 1 \\
Router $\wedge$ screen & 10 & 2 & 1 \\
Router $\wedge$ screen $+$ slice-gain & 8 & 0 & 1 \\
Refuse always & 0 & 0 & 9 \\
\bottomrule
\end{tabular}
\end{table}

\section{Decision Rules: Point Estimates versus Intervals}
\label{app:rules}

The decorrelation gate is read on the point estimate, with a bootstrap interval reported alongside; the interval does not decide the gate. Two reasons: the gate was fixed on point estimates before the interval machinery existed, so switching rules afterwards would itself be a post-hoc choice; and a CI-upper-bound rule makes the effective gate a function of sample size, so two settings with the same true effect would face different outcomes if one had a smaller slice. Table~\ref{tab:rules} sets the two rules side by side across the seven positives, so a reader who prefers the strict reading can see which settings it would drop.

\begin{table}[h]
\centering\small
\caption{The seven positives under the deployed point-estimate rule vs.\ a uniformly strict interval rule (every gate read by its CI). Marginal = the relevant CI crosses a gate the point estimate clears. Grouped headline values.}
\label{tab:rules}
\begin{tabular}{lll}
\toprule
Setting & Point-estimate rule (deployed) & Strict interval rule \\
\midrule
SNLI & retained (decorr.\ $\dagger$) & dropped (screen and decorrelation marginal) \\
SICK & retained & dropped (screen marginal) \\
WikiQA & retained (decorr.\ $\dagger$) & dropped (decorrelation marginal) \\
ASNQ & decorrelation withdrawn & dropped \\
TriviaQA & retained & retained \\
SQuAD (held-out) & retained & retained \\
DuoRC (held-out) & retained (decorr.\ marginal) & dropped (decorrelation marginal) \\
\bottomrule
\end{tabular}
\end{table}

Under the strict reading the empirical record reduces to TriviaQA and the held-out SQuAD replication. That is a substantial reduction, and it is the reason the main text does not rest its case on the marginal settings: the SQuAD held-out run, which survives both rules and has a disjoint annotator-derived
reading, is the observational result the main text rests on.

\section{An Ordering That Reverses Under Adjustment}
\label{app:reversal}

The supported claim is audit-time measurement, so the question to put to $G$ is whether it changes a conclusion someone would draw. WikiQA supplies one example, an illustration rather than a supported use: the only margin that separates will lean on a refusal the record itself marks as false. All values come from a single paired run, grouped, with intervals bootstrapped on the paired difference; the run draws its slice from the dense surface features alone, which puts $n_{\mathrm{slice}}$ at $630$ against the pipeline's $626$ and leaves slice alignments within $0.002$ of Table~\ref{tab:scope}'s. Its $388$ questions count every question touching this slice, whereas the $286$ of Table~\ref{tab:ranking} counts only ranking-eligible questions, those with a positive answer and at least two candidates. Raw, the slice alignments rank QNLI first at $0.734$, the MS~MARCO cross-encoder second at $0.624$, and STS-B third at $0.447$; the margin, $+0.111$ $[+0.059,+0.162]$, separates. Adjusted, the top two exchange places ($0.803$ against $0.777$), and the loadings explain the swap: QNLI barely rides the channel ($R^2_{\phi\to s}=0.016$ against $0.314$ and $0.360$), so its raw score had the least to recover. Neither reading of the reversal makes the example more than the illustration announced above, for the following reason. Which reversal is supported depends on which adjusted scores are supported, and there are two readings. The first adjusts both scorers; its reversed margin, $+0.027$ $[-0.015,+0.071]$, does not exclude zero. The second follows the gates: the screen refuses QNLI, so the supported comparison sets the cross-encoder's adjusted score beside QNLI's raw one, and that margin, $+0.069$ $[+0.023,+0.120]$, does separate. The second reading, however, leans on the refusal of QNLI, and that refusal is the recorded false refusal (Table~\ref{tab:c2}): the separating margin is exactly as strong as an outcome the paper itself flags. The cross-encoder's advantage is also bought on the slice and not the population ($0.864\to0.811$), and a reader holding the slice's labels could compare raw scores on it directly: what the adjusted comparison adds is the ordering that survives removing the surface channel, a diagnostic inside the audit-time scope. STS-B keeps its rank ($0.447\to0.684$).

\section{A Probe of the Entangled Regime by Assignment}
\label{app:bridge}

Section~\ref{sec:discussion} names the missing case: no real-data setting here has construct and artifact genuinely entangled. This appendix reports a probe of one route to building one, and its limitation.

\textbf{Design.} The code audit already contains four verified candidates per problem, and their edits are comment/docstring-only and were produced without reference to the label. We leave the edits untouched and vary \emph{which} two of the four enter the evaluation sample. For a fraction $p$ of problems we keep the correct-verbose and buggy-terse candidates; for the remaining problems we keep the mirror pair. Every retained candidate is still a semantics-preserving edit whose label was verified by execution, so the certification the intervention rests on is untouched; what changes is only the sampling. Each problem contributes one correct and one buggy candidate, so the natural construct measure is the within-problem paired win rate (does the score rank the correct candidate above the buggy one), which is also the instrument the main audit uses, because pooled AUC is dominated by across-problem scale variation. Ties count as half a win: the reward model emits floats and never ties, but the judges emit integers and tie on up to a fifth of the pairs.

\textbf{Two loadings for one reward model.} The reward model's loading reads $0.178$ here against the $0.283$ of Section~\ref{sec:rmcode} because the two runs score different samples of the same $353$ problems: the audit scores all four cells ($n{=}1412$) and the probe scores the two the assignment selects ($n{=}706$).

Because the audited reward model has almost no construct signal to lose, we repeat the sweep with three instruction-tuned judges (Claude Haiku 4.5, Sonnet 5, Opus 5), each asked for a $0$--$100$ correctness rating on the same $1{,}412$ candidates. The API models are \texttt{claude-haiku-4-5}, \texttt{claude-sonnet-5} and \texttt{claude-opus-5}, each returning a single integer under a JSON output schema; extended thinking is disabled on the latter two, Opus at the low effort setting under which disabling is accepted (\texttt{rescore\_mbpp\_judges.py}). This changes only the scorer; the candidates, edits, labels, and assignment are identical. Table~\ref{tab:bridge} reports all four sweeps. The assignment fraction $p$ sets the measured collinearity $\rho=\mathrm{corr}(\text{format},y)$ shown beside it ($p{=}0.5$ gives $\rho=-0.003$, the value the text rounds to $\rho\approx0$); the grid differs from the evenly spaced synthetic $\rho$ grid of Appendix~\ref{app:rhoscreen}, and rows of the two sweeps do not align one-to-one. $A(\hat a)$ and the slice size are properties of the assignment alone, so they are stated once: $A(\hat a)$ rises $0.419, 0.565, 0.674, 0.789, 0.883, 1.000$ across the six values of $p$, and the slice holds $386, 282, 212, 142, 70, 0$ candidates. The below-chance $0.419$ at $p{=}0.5$ is not the twin-leakage artifact of Appendix~\ref{app:rmcode}, since the problem-grouped folds keep format twins together; a label-permutation null with the same pairing structure is itself centered below chance and wide ($0.43$--$0.54$ over thirty within-problem flips, mean $0.48$), and the observed value sits just below that thirty-draw range.

\begin{table}[!ht]
\centering
\small
\caption{One judge's traversal of the entangled regime, abridged from the run behind Table~\ref{tab:bridge} (Haiku 4.5; $p=0.6$ and $p=0.8$ omitted; the other three scorers' full traversals are in Table~\ref{tab:bridge}). The full-population column reads the same run's per-scorer AUCs, which Table~\ref{tab:bridge} itself does not carry. $p$ is the assignment probability that induces the collinearity and $r_{\phi y}$ the realized correlation between the format channel and the label. At $p=1$ the slice is empty, so no gain is defined. The last column is the paired-win change with a problem-level bootstrap 95\% interval (1{,}000 resamples; \texttt{bridge\_win\_cis.py}).}
\label{tab:bridgehaiku}
\begin{tabular}{ccccccc}
\toprule
$p$ & $r_{\phi y}$ & $R^2_{\phi\to s}$ & Slice gain & Full pop. & Paired win rate & $\Delta$win [95\% CI] \\
\midrule
$0.5$ & $-0.003$ & $0.000$ & $+0.031$ & $0.810\to0.814$ & $0.827\to0.847$ & $+0.020$ $[-0.004,+0.044]$ \\
$0.7$ & $0.399$ & $0.011$ & $+0.080$ & $0.814\to0.756$ & $0.833\to0.725$ & $-0.108$ $[-0.142,-0.074]$ \\
$0.9$ & $0.802$ & $0.191$ & $+0.128$ & $0.811\to0.615$ & $0.822\to0.569$ & $-0.252$ $[-0.295,-0.210]$ \\
$1.0$ & $1.000$ & $0.309$ & --- & $0.810\to0.546$ & $0.820\to0.501$ & $-0.319$ $[-0.361,-0.275]$ \\
\bottomrule
\end{tabular}
\end{table}

\textbf{Findings.} The judges supply the construct margin the reward model lacks (a raw paired win rate of $0.83$ to $0.98$ against the reward model's $0.58$), and with it the sweep reproduces the synthetic prediction in a sharper form. At $p{=}0.5$, the balanced design of Section~\ref{sec:rmcode}, the adjustment is neutral on all four scorers ($-0.006$ to $+0.020$). As the format channel becomes predictive of correctness, the adjusted score loses construct signal monotonically, reaching $-0.27$ to $-0.33$ at perfect collinearity. The reward model's \emph{raw} paired win is not monotone in $p$ ($0.601\to0.637\to0.615$): each level redraws which problems contribute which cell pair, and the level-to-level differences sit inside a per-level binomial $95\%$ half-width of about $0.05$ on $353$ pairs, so we read no structure into them. For the judges this crosses from harmless to severely harmful within a single sweep, which the reward model could not show because its adjustment was already slightly negative at $\rho{\approx}0$.

Problem-level bootstrap intervals over the sweep (1{,}000 resamples, \texttt{bridge\_win\_cis.py}) are unambiguous: every judge's $\Delta$win interval covers zero at $p{=}0.5$ and excludes zero from $p{=}0.6$ upward, while the reward model's excludes zero only from $p{=}0.8$. The judges also pin down the accounting. Their raw win rate is flat across the sweep (Opus 5 scores $0.977, 0.975, 0.975, 0.977, 0.976, 0.979$), so unlike the reward model, which gains from riding the format channel as it becomes predictive through $p{=}0.8$ ($0.578 \to 0.637$), the judges' scores show no aggregate format response, so the entire $\Delta$ is damage done by the adjustment.

\textbf{What the gates read, and what the damage shows.} On the balanced sample the judges' loading is $R^2_{\phi\to s} = 0.000$, far below the $0.05$ gate: the procedure correctly declines to adjust them, the intended behavior on a scorer with no surface channel to remove. The sweep shows the same measurement rising to $0.31$--$0.71$ at $p{=}1$ purely because the artifact and the construct have been made collinear: the loading is induced by $\rho$, the judge shows little measured aggregate format sensitivity in this design, and no measurement available to the practitioner separates the two---the identifiability boundary of Section~\ref{sec:discussion}, derived in Appendix~\ref{app:identifiability}, made concrete on real text. Past $\rho \approx 0.4$ the loading gate opens, but the committed procedure still refuses every cell: the slice-degradation statistic stays below its gate throughout, and Table~\ref{tab:bridgeverdicts} records each cell's exit (\texttt{bridge\_verdicts.py}). The damage visible in Tables~\ref{tab:bridge} and~\ref{tab:bridgehaiku} is therefore what happens when the refused adjustment is applied anyway, a diagnostic of the regime rather than an output of the procedure.

\begin{table}[h]
\centering\small
\caption{Sweeping construct--artifact collinearity by assignment on the MBPP audit ($353$ problems, one correct and one buggy candidate each). ``Win'' is the within-problem paired win rate for the construct; $\Delta_{\text{win}}$ is the adjustment's change in that win rate, and $G$ its AUC change on the label-defined slice (the slice gain of Section~\ref{sec:setup}, distinct from the screen statistic $\Delta_{\mathrm{slice}}$). The slice is empty at $p{=}1$.}
\label{tab:bridge}
\begin{tabular}{llrrrrrr}
\toprule
scorer & $p$ & $\rho$ & $R^2_{\phi\to s}$ & win (raw) & win (rep.) & $\Delta_{\text{win}}$ & $G$ \\
\midrule
reward model & 0.5 & $-0.003$ & 0.178 & 0.578 & 0.572 & $-0.006$ & $+0.006$ \\
             & 0.6 & $+0.201$ & 0.179 & 0.601 & 0.575 & $-0.025$ & $+0.025$ \\
             & 0.7 & $+0.399$ & 0.181 & 0.618 & 0.584 & $-0.034$ & $+0.022$ \\
             & 0.8 & $+0.598$ & 0.182 & 0.637 & 0.558 & $-0.079$ & $+0.028$ \\
             & 0.9 & $+0.802$ & 0.182 & 0.618 & 0.507 & $-0.110$ & $+0.023$ \\
             & 1.0 & $+1.000$ & 0.183 & 0.615 & 0.473 & $-0.142$ & --- \\
\midrule
Haiku 4.5    & 0.5 & $-0.003$ & 0.000 & 0.827 & 0.847 & $+0.020$ & $+0.031$ \\
             & 0.6 & $+0.201$ & 0.000 & 0.837 & 0.807 & $-0.030$ & $+0.051$ \\
             & 0.7 & $+0.399$ & 0.011 & 0.833 & 0.725 & $-0.108$ & $+0.080$ \\
             & 0.8 & $+0.598$ & 0.068 & 0.826 & 0.635 & $-0.191$ & $+0.094$ \\
             & 0.9 & $+0.802$ & 0.191 & 0.822 & 0.569 & $-0.252$ & $+0.128$ \\
             & 1.0 & $+1.000$ & 0.309 & 0.820 & 0.501 & $-0.319$ & --- \\
\midrule
Sonnet 5     & 0.5 & $-0.003$ & 0.000 & 0.936 & 0.935 & $-0.001$ & $+0.010$ \\
             & 0.6 & $+0.201$ & 0.019 & 0.942 & 0.861 & $-0.081$ & $+0.047$ \\
             & 0.7 & $+0.399$ & 0.072 & 0.939 & 0.802 & $-0.137$ & $+0.059$ \\
             & 0.8 & $+0.598$ & 0.180 & 0.938 & 0.762 & $-0.176$ & $+0.071$ \\
             & 0.9 & $+0.802$ & 0.335 & 0.941 & 0.697 & $-0.244$ & $+0.072$ \\
             & 1.0 & $+1.000$ & 0.516 & 0.938 & 0.606 & $-0.331$ & --- \\
\midrule
Opus 5       & 0.5 & $-0.003$ & 0.000 & 0.977 & 0.980 & $+0.003$ & $+0.006$ \\
             & 0.6 & $+0.201$ & 0.026 & 0.975 & 0.929 & $-0.045$ & $+0.020$ \\
             & 0.7 & $+0.399$ & 0.104 & 0.975 & 0.881 & $-0.093$ & $+0.022$ \\
             & 0.8 & $+0.598$ & 0.249 & 0.977 & 0.847 & $-0.130$ & $+0.026$ \\
             & 0.9 & $+0.802$ & 0.468 & 0.976 & 0.805 & $-0.171$ & $+0.047$ \\
             & 1.0 & $+1.000$ & 0.714 & 0.979 & 0.705 & $-0.273$ & --- \\
\bottomrule
\end{tabular}
\end{table}

\begin{table}[h]
\centering\small
\caption{Committed-procedure exit for every cell of the assignment sweep (\texttt{bridge\_verdicts.py}). R: no usable slice, refused at the router ($A(\hat a)<0.65$). $R^2$: refused at the screen's loading gate. $\Delta$: refused at the screen's slice-degradation gate ($\Delta_{\mathrm{slice}}<0.03$). ---: slice empty at $p{=}1$.}
\label{tab:bridgeverdicts}
\begin{tabular}{lcccccc}
\toprule
scorer & $p{=}0.5$ & $0.6$ & $0.7$ & $0.8$ & $0.9$ & $1.0$ \\
\midrule
reward model & R & R & $\Delta$ & $\Delta$ & $\Delta$ & --- \\
Haiku 4.5 & R & R & $R^2$ & $\Delta$ & $\Delta$ & --- \\
Sonnet 5 & R & R & $\Delta$ & $\Delta$ & $\Delta$ & --- \\
Opus 5 & R & R & $\Delta$ & $\Delta$ & $\Delta$ & --- \\
\bottomrule
\end{tabular}
\end{table}

The slice \emph{gain} moves the wrong way. For Haiku 4.5 it climbs $+0.031 \to +0.128$ across the range where the full-population change moves from $+0.004$ to $-0.196$ (reaching $-0.264$ at $p{=}1$, where the slice is empty) and the adjusted score's paired win rate collapses from $0.847$ to $0.569$ ($0.501$ at $p{=}1$). A practitioner watching only the slice would read increasing improvement while the scorer was being destroyed. The refusals here rest on the observed slice-degradation statistic, not on any estimate of construct--artifact collinearity; the full-population cost reported beside every slice gain is the quantity that moves in the right direction throughout. This yields a monotone diagnostic trend across the sweep: the cost falls as the damage rises, so a reader who sees it move can see the trade getting worse. No single value of it separates the expected trade from destruction, and Section~\ref{sec:discussion} accordingly makes the pair a mandatory report and not an exit condition.

\textbf{Status of the probe.} The collinearity is induced by sampling, so the setting remains semi-synthetic: stronger than the Gaussian sweep of Section~\ref{sec:synthetic}, since the text, the scorers, and the labels are all real and the edits are certified, but weaker than a naturally entangled dataset, which we still do not have. The design also cannot produce a scorer that has both a real surface shortcut and a strong construct margin: the two scorer families available to us have one property each, so the entanglement has to be supplied by the assignment.

\section{Falsification Conditions}
\label{app:falsification}

This appendix lists the observations that would count as evidence against the procedure, so that its outcome categories, and refusal in particular, cannot be satisfied by every
result. The conditions were written during manuscript preparation, after the scorer replications were known; they therefore bind future evidence and certify nothing about the record reported here.

\textbf{(i) Router$\wedge$screen pass but the slice gain is zero or negative.} This condition carries two severities. Against a disjoint or intervention-based evaluation it is a direct disconfirmation, because the screen's prediction fails and refusal cannot absorb it; this is currently checkable only on SQuAD and Natural Questions, the settings with such a reading. Against a same-label evaluation it is a weaker disconfirmation, since same-label gains are optimistically biased and their failures are ambiguous.

\textbf{(ii) An outcome flips under a monotone reparameterization of the score.} The operator would then be under-specified.

\textbf{(iii) An effect vanishes under question-grouped cross-fitting.} The original result would then be a leakage artifact.

\textbf{(iv) Designed-intervention attenuation succeeds repeatedly where the slice-eligibility check and screen fail \emph{with a low loading}.} Their scope claim would then be false, and conservatism would not explain it. The qualifier matters: Section~\ref{sec:rmcode} is a case where the slice-eligibility check fails and removal succeeds, but it fails with $R^2_{\phi\to s}=0.283$, which is the branch that sends the setting to a designed measurement target. Success there is what the procedure predicts. Success where the loading is also near zero would falsify it. The large-v2 checkpoint (Appendix~\ref{app:rmcode}) is the closest instance on record: its pooled loading is $0.023$, below the screen, and its paired format effects still attenuate by $+0.281$ $[+0.270,+0.291]$ and $+0.286$ $[+0.274,+0.297]$. Because pooled $R^2$ understates a within-problem channel when between-problem variance dominates, this condition is read on the paired loading for designed audits, and a low pooled loading is not read as ``nothing to remove''.

Conditions (ii) and (iii) have already been exercised (Appendix~\ref{app:grouped}). No outcome flips under three alternative score representations, and the slice gains survive grouped cross-fitting. One decorrelation claim does not survive it: ASNQ's is withdrawn accordingly.

Condition (i) has three recorded instances, all of the weaker kind because all are same-label evaluations. Two are scorer replications that clear both gates: ASNQ (QNLI) returns a null slice gain ($+0.007$, n.s.), and TriviaQA (QNLI) loses, $-0.014$ $[-0.030,-0.001]$. The third comes from the cross-family control of Appendix~\ref{app:crossfamily}: ASNQ on its length-defined slice clears both gates and loses, $-0.019$ $[-0.037,-0.002]$. Both QNLI replications are caught at the slice-gain test (ASNQ's interval covers zero, TriviaQA's lies below it) and the settings end in \emph{no gain established}, but the condition is written about the pre-adjustment stages and is recorded on that basis. We diagnose the first below, since its mechanism generalizes to the others. The direct diagnostic for why is the slice-conditional alignment of the component the operator removes: on the slice, $\mathrm{corr}(\hat g, y)$ is $-0.081$ for the QNLI scorer against $-0.629$ for the MiniLM scorer on the same rows and slice, with full-sample values $+0.084$ and $+0.269$. Subtraction helps on the slice to the extent that the removed component is anti-aligned with the construct there, and for this scorer it barely is. The gates, which read the loading and the degradation, do not consult this correlation, so within the committed procedure the row is a genuine near-miss. The statistic itself, however, is computable before the slice-gain test: it needs only $\hat g$, $\hat a$, and $y$. It is nonetheless not promoted to a third screen: it was identified after the very failure
it would have caught, which is the adaptivity the accounting is designed to exclude, and it
presupposes the declared slice and measures the same conditional geometry the slice-gain test measures, so as a gate it would partly duplicate the slice-gain test rather than corroborate
it---moving the double-use concern of Appendix~\ref{app:slicebias} into the pre-adjustment
stages. We record it as a \emph{prospective} candidate for future audits rather than
retrofitting it here. It stays a refusal-analysis diagnostic, and as a same-label evaluation it does not carry the full condition's force; no direct instance has occurred. The face-value boundary picture is symmetric: the two QNLI replications err in opposite directions (Section~\ref{sec:positives}; Figure~\ref{fig:repairability}).

\section{The Slice-Eligibility Check Across the Collinearity Sweep}
\label{app:rhoscreen}

\textbf{The crossings, in detail.} The sweep also explains a pattern that would otherwise look like an anomaly. The sweep varies $\rho$ while the slice-eligibility check is left to move on its own, and its informativeness and the adjustment's full-population benefit run in opposite directions: at $\rho{=}0$ the artifact is independent of the label, so a $\phi$-only predictor is at chance and the slice-eligibility check fails, precisely where the adjustment helps most on the full population. Measured across the sweep, the slice-eligibility check crosses its gate at $\rho\approx0.4$ and the full-population gain crosses zero at $\rho\approx0.5$; neither location is fundamental, and the ordering is not either. The slice-eligibility check is computed from $\phi$ and $y$ alone, so its crossing does not move with the artifact loading $\beta$; the full-population crossing does, and Appendix~\ref{app:identifiability} places it at the root $\rho^{\ast}(\beta)$ of the condition derived there: $0.489$ at this sweep's $\beta{=}1.2$, but $0.363$ at $\beta{=}0.6$ and $0.281$ at $\beta{=}0.4$, below the slice-eligibility check's crossing. For weakly loaded scorers the damage begins before the slice-eligibility check's gate opens, which is what the assignment probe of Appendix~\ref{app:bridge} measures on three instruction-tuned judges whose balanced-sample loading is $R^{2}_{\phi\to s}=0.000$: the change in their paired construct win rate turns negative by $\rho\approx0.2$, while the slice-eligibility check does not clear its gate until $\rho\in(0.2,0.4)$. The two $\rho$ are different constructions and are not on a common scale: here it is a population parameter of a Gaussian model, and there it is induced by which already-verified candidates enter the sample, so their locations do not carry across.

The ordering of the two crossings carries across only if the two parameterizations are monotonically related in $\rho$, which we assume and do not verify. On its own scale the probe establishes that the damage begins before the slice-eligibility check's gate opens, and that is the claim we rest on. This sweep fixes one ordering at one loading, and the entangled regime is where that ordering fails. Meanwhile from $\rho{=}0.2$ upward the slice gain stays near $+0.55$ to $+0.58$ (Table~\ref{tab:rhoscreen}); at $\rho{=}0$ itself the slice has degenerated and the gain collapses to $+0.061$. The screen selects for slice informativeness, not for adjustment benefit, which is why every real positive's full-population number declines, from $0.958\to0.880$ on SNLI to $0.951\to0.867$ on DuoRC, and why the cost is a mandatory co-report (Section~\ref{sec:discussion}). The designed audit of Section~\ref{sec:rmcode} occupies the other regime, $\rho\approx0$ with the measurement target supplied by design.

Section~\ref{sec:synthetic} states that the slice-eligibility check's gate opens roughly where full-population adjustment stops paying. Table~\ref{tab:rhoscreen} is the measurement behind that statement: it reports both screening statistics next to both gains at each collinearity level of the same sweep as Figure~\ref{fig:rho}.

\begin{table}[h]
\centering\small
\caption{The two screening statistics beside both adjustment gains at each collinearity level; same generative model, $N{=}4000$, and seed as Figure~\ref{fig:rho}, re-run with the screening statistics added, which offsets values from the figure's grid by up to $0.01$. At $\rho{=}0$ the slice degenerates: $\hat a$ is at chance, so the slice is an arbitrary half of the data ($n_{\mathrm{slice}}{=}1988$ of $4000$) and $\Delta_{\mathrm{slice}}$ is negative.}
\label{tab:rhoscreen}
\begin{tabular}{lccccccc}
\toprule
construct--artifact collinearity $\rho$ & 0.0 & 0.2 & 0.4 & 0.5 & 0.7 & 0.8 & 0.9 \\
\midrule
$A(\hat a)$ (router gate $\approx0.65$) & 0.473 & 0.581 & 0.669 & 0.715 & 0.811 & 0.863 & 0.917 \\
$R^2_{\phi\to s}$ (gate $\approx0.05$) & 0.563 & 0.643 & 0.724 & 0.765 & 0.847 & 0.888 & 0.930 \\
$\Delta_{\mathrm{slice}}$ (gate $\approx0.03$) & $-0.100$ & $+0.403$ & $+0.443$ & $+0.469$ & $+0.506$ & $+0.557$ & $+0.606$ \\
\midrule
residualization gain, full population & $+0.168$ & $+0.115$ & $+0.042$ & $-0.003$ & $-0.117$ & $-0.191$ & $-0.288$ \\
residualization gain, declared slice & $+0.061$ & $+0.550$ & $+0.561$ & $+0.568$ & $+0.563$ & $+0.583$ & $+0.583$ \\
\bottomrule
\end{tabular}
\end{table}

The two crossings quoted in Section~\ref{sec:synthetic} (the eligibility gate at $\rho\approx0.4$, the full-population zero at $\rho\approx0.5$) are read from this table. Meanwhile the slice gain sits between $+0.550$ and $+0.583$ at every $\rho\ge0.2$; the $\rho{=}0$ column is the degenerate case described in the caption. Read together with the identification boundary of Appendix~\ref{app:identifiability}, the sweep instantiates with measured values the two branches an eligibility failure can take. A low $R^2_{\phi\to s}$ means the fitted feature model explains little of the pooled score variance, so subtraction is dominated by estimation noise with room for harm, the diagnosis behind the MNLI and toxicity refusals; the designed audits show the pooled statistic can also understate a within-item channel, which is why that condition is read on the paired loading there (Appendix~\ref{app:rmcode}). A high $R^2_{\phi\to s}$ means the channel is present but label-orthogonal, so only the measurement target is missing, which is the reward-model regime of Section~\ref{sec:rmcode} ($R^2_{\phi\to s}=0.283$ with problem-grouped $A(\hat a)=0.500$). The practitioner's next step differs accordingly: stop, versus design a measurement target the slice cannot supply.

\section{The Report-Only Baseline}
\label{app:baseline}

The natural baseline for the observational regime is to skip adjustment entirely and
report the raw slice alignment together with the degradation $\Delta_{\mathrm{slice}}$:
the diagnosis is kept, the adjustment is withheld, and $y$ is still consumed.
Table~\ref{tab:baseline} shows why that summary is not sufficient: the ratio of the
measured gain $G$ to $\Delta_{\mathrm{slice}}$ ranges from below $0$ to $1.38$.
Replications with nontrivial $\Delta_{\mathrm{slice}}$ yield nothing, and SNLI's small
$\Delta_{\mathrm{slice}}$ understates its measured gain, so the report-only
baseline misjudges in both directions. The ratio is not a share, so it is not bounded by
$1$; on SNLI it exceeds $1$.

\begin{table}[h]
\centering\small
\caption{The report-only baseline vs.\ the measured slice gain: $\Delta_{\mathrm{slice}}$ (available without adjustment) beside the measured gain $G$ and their ratio, grouped protocol.}
\label{tab:baseline}
\begin{tabular}{lccc}
\toprule
Setting & $\Delta_{\mathrm{slice}}$ (report-only) & Measured gain $G$ & $G/\Delta_{\mathrm{slice}}$ \\
\midrule
SNLI & $0.032$ & $+0.044$ & $1.38$ \\
SICK & $0.202$ & $+0.199$ & $0.99$ \\
WikiQA & $0.242$ & $+0.181$ & $0.75$ \\
ASNQ & $0.356$ & $+0.200$ & $0.56$ \\
TriviaQA & $0.233$ & $+0.148$ & $0.64$ \\
SQuAD (held-out) & $0.226$ & $+0.176$ & $0.78$ \\
DuoRC (held-out) & $0.144$ & $+0.056$ & $0.39$ \\
ASNQ (QNLI scorer) & $0.077$ & $+0.007$ (n.s.) & $0.09$ \\
TriviaQA (QNLI scorer) & $0.046$ & $-0.014$ & $<0$ \\
\bottomrule
\end{tabular}
\end{table}

The limit of this argument: $G$ is computed on the same label-defined slice, so it
establishes the insufficiency of the report-only summary without externally validating
$G$. That external anchor exists only where a disjoint evaluation does, namely SQuAD
($+0.155$) and the post-hoc Natural Questions setting ($+0.220$).

\textbf{The observable baseline: the risk flag on its own.} A practitioner who wants
something usable at deployment time has one more option that needs no residualization and
no five-stage procedure: fit the slice-membership flag and act on its ranking.
Table~\ref{tab:flagbaseline} places that option between the two endpoints on the three
settings where the frozen flag is evaluated out of sample.

\begin{table}[!ht]
\centering\small
\caption{The risk flag as a baseline in its own right, on the three non-development settings. Rows are what a practitioner can do; columns are construct alignment on the label-defined slice and on the full population. The first row covers the raw score and both no-change baselines at once: reporting $\Delta_{\mathrm{slice}}$ and computing the flag leave the score untouched, so they share its alignment. All endpoints are recomputed inside a single run (\texttt{risk\_flag\_switch}; row-level cross-fitting, as in Table~\ref{tab:switch}), which is why they can differ from Table~\ref{tab:threepop} in the third decimal.}
\label{tab:flagbaseline}
\setlength{\tabcolsep}{4.5pt}
\begin{tabular}{llcccccc}
\toprule
& & \multicolumn{2}{c}{SQuAD} & \multicolumn{2}{c}{DuoRC} & \multicolumn{2}{c}{TriviaQA} \\
\cmidrule(lr){3-4}\cmidrule(lr){5-6}\cmidrule(lr){7-8}
Baseline & Score & Slice & Full & Slice & Full & Slice & Full \\
\midrule
raw; $\Delta_{\mathrm{slice}}$-only; flag-only & unchanged & 0.706 & 0.932 & 0.808 & 0.951 & 0.474 & 0.706 \\
flag used as a switch (top $50\%$) & changed & 0.807 & 0.887 & 0.841 & 0.911 & 0.487 & 0.701 \\
full procedure & changed & 0.883 & 0.835 & 0.869 & 0.871 & 0.626 & 0.637 \\
\bottomrule
\end{tabular}
\end{table}

The flag is genuinely informative about \emph{where} the score fails: frozen after
fitting on the development settings, it predicts slice membership at AUC $0.809$
$[0.800,0.817]$ on SQuAD, $0.913$ $[0.905,0.922]$ on DuoRC and $0.739$ $[0.731,0.747]$ on
TriviaQA, concentrating slice members in its top quintile at $2.3\times$, $4.4\times$
and $1.8\times$ base rate. Acting on that ranking does not escape the trade: switching
to the adjusted score on the flagged half buys slice alignment at a full-population cost
in roughly the proportion the two endpoints imply, so the flag repositions the
slice-versus-population trade without removing it. The full procedure adds a \emph{quantity} the flag lacks: $G$, the change in slice AUC produced by subtracting the fitted surface
channel, which on SQuAD carries the disjoint-annotator anchor of
Section~\ref{sec:heldout}. The flag ranks examples but does not say how much of the error the channel accounts for; the procedure measures the alignment change induced by subtracting the fitted
component, and refuses where it cannot.
Conversely the flag is the one component that survives to deployment;
Section~\ref{sec:costs} reports it there; here it remains an evaluation-time instrument.

\section{Per-Setting Details}
\label{app:settings}

\textbf{Checkpoints and datasets.} Scorers are public checkpoints (DistilBERT-MNLI, BERT-SNLI/QQP, toxic-BERT, and the MS~MARCO MiniLM-L6, QNLI DistilRoBERTa, STS-B RoBERTa, and ELECTRA cross-encoders \citep{reimers2019sentencebert}; their architectures are BERT \citep{devlin2019bert}, DistilBERT \citep{sanh2019distilbert}, RoBERTa \citep{liu2019roberta}, MiniLM \citep{wang2020minilm} and ELECTRA \citep{clark2020electra}, and the reward models are DeBERTa-v3 \citep{he2021deberta}; the QNLI and STS-B tasks are GLUE's \citep{wang2019glue}, the latter built on SemEval-2017 Task 1 \citep{cer2017stsb}); datasets are public (SNLI \citep{bowman2015snli}, SICK \citep{marelli2014sick}, WikiQA, ASNQ, TriviaQA \citep{joshi2017triviaqa}, MNLI \citep{williams2018mnli}, HANS, ANLI, QQP, CivilComments \citep{borkan2019civil}, HellaSwag, SWAG; the held-out SQuAD \citep{rajpurkar2016squad}, DuoRC, and Natural Questions; the MBPP \citep{austin2021mbpp} and HumanEval \citep{chen2021humaneval} code audits; and the screened SHP \citep{ethayarajh2022shp}, hh-rlhf \citep{bai2022hhrlhf}, BoolQ \citep{clark2019boolq}, HateCheck \citep{rottger2021hatecheck}, SciQ \citep{welbl2017sciq}).

\textbf{The SQuAD walkthrough, stage by stage.} We walk through the run stage by stage, numbering the stages as in Figure~\ref{fig:workflow}; the disjoint-annotator evaluation that follows provides the strongest evidence, because stage \textcircled{\footnotesize 5} is favored by the slice conditioning once the slice-eligibility check and screen pass.

\emph{Ingredients.} The score $s$ is an MS~MARCO cross-encoder's relevance score for a candidate sentence; the artifact features $\phi$ are answer-only surface cues plus question--candidate lexical overlap. $\phi$ is a \emph{partial-input} feature set: in the text settings it carries a TF-IDF block over one side of the pair alongside the dense cues (Table~\ref{tab:phispec}), so it sees content, but only from one side. A1 therefore asks more of $\phi$ here than it would of length alone; Section~\ref{sec:samelabel} splits $\phi$ and measures how much of the QA result that assumption carries. The construct measurement $y$ is the annotator's answer span, which comes from SQuAD's own annotation and never sees the cross-encoder. \emph{Stage \textcircled{\footnotesize 1}, the slice-eligibility check:} a $\phi$-only model predicts $y$ at AUC $0.813$, comfortably above the gate, so the set of examples where that model errs is an informative slice; the slice-eligibility check passes, and its outcome was written to a separate file before the cross-encoder scored anything. That file's own wording for this outcome is ``improvement expected''. The artifact used this shorthand; operationally the committed variable was slice-eligibility check eligibility, not the sign of $G$. \emph{Stage \textcircled{\footnotesize 2}, the screen:} the score loads on the channel ($R^2_{\phi\to s}=0.491$) and is degraded where the surface model errs ($\Delta_{\mathrm{slice}}=0.226$), so both components clear their gates and the procedure proceeds to adjust. \emph{Stage \textcircled{\footnotesize 3}, adjustment:} one cross-fitted ridge regression of $s$ on $\phi$, with question-grouped folds, produces $\srep=s-\hat g$.

\emph{Stage \textcircled{\footnotesize 4}, decorrelation:} the adjusted score's correlation with the surface index falls from $0.627$ to $0.027$, inside the gate. \emph{Stage \textcircled{\footnotesize 5}, the slice-gain test:} on the declared slice, construct alignment rises $0.706\to0.882$, a gain of $+0.176$ (query-clustered CI $[+0.155,+0.197]$), while on the full population it falls, $0.932\to0.836$, a cost that is expected and that we report alongside the gain. The \emph{outcome} is a measured slice gain: on this evaluation set, subtracting the
fitted surface component increases alignment on the declared error slice.

\textbf{The Natural Questions replication, in full.} Because the annotator separation rested on a single setting, we built a second: answer-sentence selection from the Natural Questions validation set \citep{kwiatkowski2019nq}, whose five-way annotations supply the same structure ($n{=}4{,}929$ over all $946$ eligible questions). It was constructed after every freeze, so it sits outside the paper's pre-commitment accounting; everything downstream of construction is this paper's recipe unchanged. Every gate passes ($A(\hat a)=0.677$, $R^2_{\phi\to s}=0.442$, $\Delta_{\mathrm{slice}}=0.381$), and the adjustment lifts the slice from $0.402$ to $0.658$. The two annotator pools agree on $95.2\%$ of candidates, a wider disagreement region than SQuAD's, so this check discriminates more; the gain survives against the disjoint evaluation at $+0.220$ $[+0.196,+0.244]$, $86\%$ of the same-label magnitude.

\textbf{The HANS constituent subset, in detail.} The sign correspondence of Section~\ref{sec:boundary-refusals} accounts for the sign only where the subset-level correlation is materially non-zero, and for the magnitude only where the score is not saturated. Constituent is the subset with neither: its subtracted-component correlation with the construct is $-0.01$, so no directional term operates, and the raw score is saturated there (median $0.992$, tenth percentile $0.759$), so the subtracted component (whose standard deviation, $0.037$, is a sixth of the score's) reorders the saturated bulk without a preferred direction. The measured $-0.064$ is that reshuffle; a subset scored near ceiling has little to gain from a directionless perturbation and its ordering to lose, which is consistent with the negative sign without determining it. So a near-zero correlation does not, on a saturated subset, predict a near-zero effect.

\begin{figure}[h]
\centering
\includegraphics[width=\textwidth]{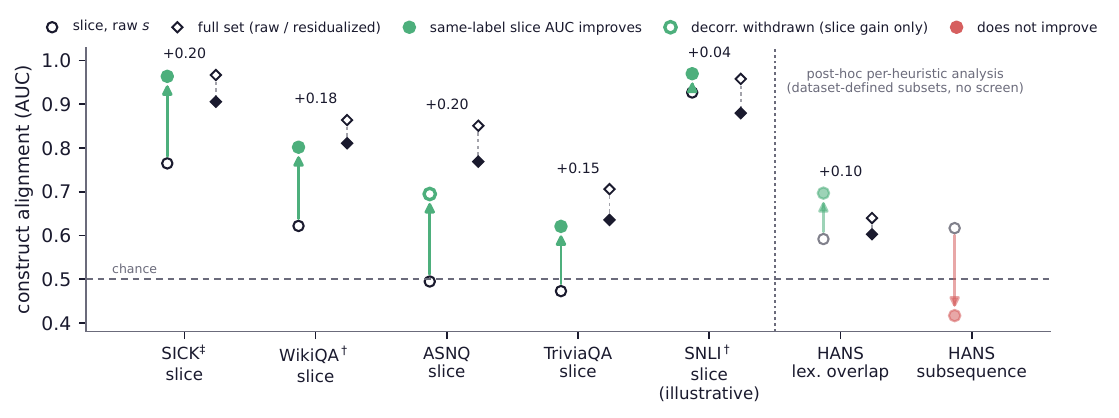}
\caption{Same-label slice outcomes on real scores (construct alignment before $\to$ after; QA values question-grouped; same-label evaluations optimistically biased, Appendix~\ref{app:slicebias}). Arrows: slice alignment, raw (open circle) $\to$ residualized (filled). Open green circle: index-decorrelation diagnostic not met (ASNQ). $^{\dagger}$: decorrelation marginal under the interval reading. $^{\ddagger}$: wide-interval row, see text. Strict-interval reading: Appendix~\ref{app:rules}. Muted markers right of the divider: post-hoc HANS analysis, no router reading. Diamonds: full-population outcome (open raw, filled residualized), always reported alongside. Red: does not improve.}
\label{fig:real}
\end{figure}

One row of Figure~\ref{fig:real} needs a weight note: SICK's slice is small ($n{=}455$) and its $\Delta_{\mathrm{slice}}$ interval correspondingly wide ($[0.002,0.429]$), so the row corroborates the pattern weakly and we use it only as supporting evidence (the
$\ddagger$ mark).

\textbf{Provenance and independence risk, per setting.} ``Independent $y$'' is a provenance criterion that admits degrees; the right unit is a per-setting risk assessment. Mechanical labels (string containment) carry the lowest dependence risk; crowd labels sharing task conventions with the scorer's training data carry more.

\textbf{Chronology of design freezes.} As recorded in the released development artifacts (not third-party timestamped): (1) the $\phi$ recipes, operator (ridge $\alpha{=}1$, 5-fold, seed 0), and slice rule were fixed with the first NLI settings, before any QA work; (2) the slice-eligibility check/screen/decorrelation gates were fixed after the first five settings and applied unchanged afterwards; (3) WikiQA, ASNQ, and TriviaQA were selected by the slice-eligibility check before their scorers ran, with ASNQ's candidate-structure decision disclosed alongside both $A(\hat a)$ values; (4) the SQuAD run froze the entire pipeline and wrote its committed slice-eligibility check decision to a machine-readable artifact before scoring. Not in this list: the falsification conditions of Appendix~\ref{app:falsification} were written during manuscript preparation, after all main results (including the ASNQ (QNLI) near miss) were known; they bind future evidence; the results reported here fall outside them.

\begin{table}[h]
\centering\small
\caption{Provenance decomposition and shared-convention risk for the construct measurement $y$, per setting.}
\label{tab:indep}
\resizebox{\textwidth}{!}{%
\begin{tabular}{llll}
\toprule
Setting & Scorer provenance & $y$ provenance & Dependence risk \\
\midrule
SNLI & DistilBERT fine-tuned on MNLI & SNLI crowd gold (5 annotators) & medium (shared NLI conventions) \\
SICK & same & expert-constructed labels & medium--low \\
WikiQA & MS~MARCO cross-encoder & crowd relevance labels & medium--low \\
ASNQ & same & mechanical span mapping from NQ & low--medium \\
TriviaQA & same & mechanical answer containment & low \\
SQuAD (held-out) & same & mechanical answer containment & low \\
DuoRC (held-out) & same & mechanical answer containment & low \\
NQ (post hoc) & same & mechanical span mapping, five-way annotations & low--medium \\
MBPP (RM audit) & OpenAssistant preference RM & mechanical unit-test execution & none \\
HumanEval (RM audit) & same & mechanical unit-test execution & none \\
\bottomrule
\end{tabular}}
\end{table}

\begin{table}[!ht]
\centering\small
\caption{Specification of the artifact features $\phi$ per setting: composition and dimension. TF-IDF blocks are 1--2 grams with the \texttt{max\_features} cap shown, fit once on the evaluation sample; the cross-fitted objects are the ridge $\hat g$ (fit on all of $\phi$) and the logistic probe $\hat a$ (fit on $\phi_a$: the four dense cues in the QA rows, the full $\phi$ elsewhere); the vectorizer is not among them. The cap is reached in the settings whose rows are cached (ASNQ, TriviaQA, NQ); elsewhere the stated figure is the cap.}
\label{tab:phispec}
\setlength{\tabcolsep}{5pt}
\begin{tabular}{llc}
\toprule
Setting & $\phi$ composition & $\dim\phi$ \\
\midrule
SNLI, SICK, MNLI (cross-domain) & hypothesis-only TF-IDF & 8000 \\
HellaSwag, SWAG & ending-only TF-IDF & 8000 \\
WikiQA, ASNQ, TriviaQA, SQuAD, DuoRC, NQ & answer/sentence-only TF-IDF $+$ 4 dense & 8004 \\
QQP & second-question TF-IDF $+$ 4 dense & 8004 \\
HANS & 4 dense $+$ hypothesis TF-IDF (cap 300) & 304 \\
Toxicity (CivilComments) & 28 identity-term counts $+$ length $+$ count sum & 30 \\
MBPP, HumanEval (reward-model audit) & 9 format-only features & 9 \\
\bottomrule
\end{tabular}
\end{table}

Table~\ref{tab:phispec} gives the composition and dimension of $\phi$ in each setting. The four dense features in the QA rows are question--answer token overlap, answer length, question length, and their difference; the code audit's nine are character, line and token counts, comment count and density, docstring presence, blank-line count, and mean and maximum line length. The released artifact records, per setting: feature definitions ($\phi$), scorer checkpoints, sample sizes, slice constructions, and full before/after tables for each of the twelve settings, including the HANS per-heuristic decomposition, the fold-shuffling control (an ordered-file leakage artifact we discovered and fixed in the HANS pipeline; SNLI results are unchanged by it), and the RewardBench diagnostics.

\begin{table}[h]
\centering\small
\caption{The six refused analysis rows, with the columns of Table~\ref{tab:scope}: a named gate fails and the outcome column names which exit. Refusal basis: the raw full-population alignment where the score is uncorrupted, or ``broken'' where that alignment is near chance on the domain.}
\label{tab:refused}
\setlength{\tabcolsep}{4pt}
\renewcommand{\arraystretch}{1.15}
\resizebox{\textwidth}{!}{%
\begin{tabular}{llccllll}
\toprule
Setting (score) & $A(\hat a)$ & Screen & Refusal basis & Slice or subset effect & Full pop. & Outcome \\
\midrule
HANS constituent/subseq. & ---$^{\P}$ & \fail & $R^2_{\phi\to s}{\approx}0$ & $-0.064$ / $-0.200$ & --- & \vnv{nonlinear} \\
MNLI (SNLI-BERT, off-dist) & 0.617 \fail & \fail & raw $0.923$ & $-0.034$ $[-0.05,-0.02]$ & --- & \vscr{weak artifact} \\
Toxicity (CivilComments) & 0.557 \fail & \fail & raw $0.953$ & $+0.008$ n.s. & --- & \vscr{weak artifact} \\
QQP (in-distribution BERT) & 0.636 \fail & \fail & raw $0.993$ & $-0.045$ $[-0.06,-0.04]$ & --- & \vscr{no slice degradation} \\
ANLI (artifact removed by design) & 0.556 \fail & --- & broken (0.446) & $-0.043$ & --- & \vscr{no linear channel} \\
HellaSwag / SWAG (ending-only) & 0.59/0.56 \fail & --- & broken ($\approx$0.57) & $-0.016$ / $-0.013$ n.s. & --- & \vscr{nonlinear} \\
\bottomrule
\end{tabular}%
}
\end{table}

\section{Candidate Selection by the Slice-Eligibility Check}
\label{app:screen}

Table~\ref{tab:screen} reports \emph{every} candidate considered in the non-NLI candidate selection of
Section~\ref{sec:positives}, to make the selection auditable: in the original candidate selection WikiQA was the single candidate whose artifact channel
cleared the slice-eligibility check, determined before any model scoring (candidate selection costs one cross-validated logistic fit per candidate and touches no scorer); ASNQ ($0.747$, with the candidate-structure protocol note of Section~\ref{sec:positives}) and TriviaQA ($0.675$) were screened subsequently under the same procedure, cleared, and adjusted as predicted. Their rows record eligibility-stage refusals; the recall measurement below scores seven of them afterwards and runs the frozen pipeline to an outcome. Three of the rejections cluster directly below the gate ($0.637$--$0.647$), in the same band as the scored negatives of Table~\ref{tab:scope}, additional evidence that the gate sits in a real gap. Separately, the two preference datasets show that the reward-model length bias, though real, is weak as a \emph{linear} channel ($0.647$ on SHP) or absent ($0.513$ on hh-rlhf, whose paired responses are length-matched by construction of the sampling), which is why preference scores remain outside the operator's demonstrated scope.

\textbf{Why the per-question candidate protocol (ASNQ).} ASNQ's raw release pairs each question with every sentence of a Wikipedia page ($\sim$275 candidates per question, positive rate $0.005$); on such naive rows the artifact features reach $A(\hat a)=0.643$. That is not the distribution any answer-selection deployment scores. The evaluation pool was fixed instead to the field's standard AS2 unit, per-question lists, all positives plus at most 19 sampled negatives, matching WikiQA's shape and base rate \citep{yang2015wikiqa,garg2020tanda}, \emph{before} any scorer ran, with the naive value reported alongside. The protocol was frozen before any scorer output existed, its per-question unit is the
datasets' own intended evaluation shape, and applying it unchanged to TriviaQA returns a
different value ($0.675$). The pair ($0.643$/$0.747$) demonstrates a real property of the slice-eligibility check: linear recoverability is relative to the candidate distribution, so candidate selection has to be run on the same distribution the adjustment will face.

\begin{table}[h]
\centering\small
\caption{The complete non-NLI candidate selection ($A(\hat a)$ = the router's surface-only AUC, row-level cross-validation as computed at selection time; gate $\approx$0.65). WikiQA and ASNQ read $0.747$ at selection time on different data; the agreement to three decimals is coincidental and they separate under the deployed protocol ($0.748$ and $0.751$). Under the question-grouped protocol adopted for all headline numbers the third decimal shifts slightly (ASNQ $0.747\to0.751$, WikiQA $0.747\to0.748$, TriviaQA $0.675\to0.676$; Table~\ref{tab:scope}); the selection decisions are unchanged. Only candidates that cleared the gate were subsequently scored; rejected rows are router-stage refusals. Adjustment outcomes quoted in the last column are the grouped headline gains.}
\label{tab:screen}
\resizebox{\textwidth}{!}{%
\begin{tabular}{lllcl}
\toprule
Dataset & Task & Artifact features & $A(\hat a)$ & Selected? \\
\midrule
WikiQA & QA answer ranking & overlap $+$ lengths & \textbf{0.747} & \pass\ (strongest linear channel) \\
WikiQA & --- & answer-only TF-IDF & 0.677 & (secondary channel, included in $\phi$) \\
ASNQ & QA answer ranking & overlap $+$ lengths & \textbf{0.747} & \pass\ screened later; adjusted; slice gain $+0.200$ \\
TriviaQA & QA answer ranking & overlap $+$ lengths & \textbf{0.675} & \pass\ screened later; adjusted; slice gain $+0.148$ \\
SciQ & science MCQ & overlap $+$ lengths & 0.504 & \fail\ no linear artifact (well-designed distractors) \\
SHP & human preference & length features & 0.647 & \fail\ below gate \\
BoolQ & boolean QA & question-only TF-IDF & 0.640 & \fail\ below gate \\
HateCheck & hate-speech tests & identity terms $+$ length & 0.637 & \fail\ below gate \\
hh-rlhf & human preference & length features & 0.513 & \fail\ no linear artifact \\
\bottomrule
\end{tabular}}
\end{table}

\textbf{Recall of the candidate selection: running the pipeline on refused candidates.} A
selection rule can only be credited with refusals that would have failed anyway, so we scored seven of the refused candidates
afterwards and ran the frozen pipeline on each to an outcome (\texttt{screen\_recall\_run.py}; the
scorer paired with each candidate is recorded in its committed artifact, and each run subsamples
$2{,}000$--$4{,}000$ rows). Table~\ref{tab:screenrecall} reports the result. Four refusals are
correct: the pipeline reaches no measured gain on BoolQ, hh-rlhf, PAWS or RTE. Three are false
negatives---SHP, HateCheck and MNLI hypothesis-only would have produced measured slice gains of
$+0.147$, $+0.108$ and $+0.036$---and all three sit in the band immediately below the gate
($A(\hat a)=0.616$--$0.647$ against the $0.65$ constant). The ordering the slice-eligibility check induces is
therefore informative on this pool, while the constant is conservative: every miss is a near-miss,
and none of the four candidates far below the gate would have passed the slice-gain test. This is the recall
counterpart to the threshold sweep of Appendix~\ref{app:threshold}, which varies the constant on
the analysis rows; neither run supports the slice-eligibility check as a detector of harmful adjustment.

\begin{table}[h]
\centering\small
\caption{Screen recall: the frozen pipeline run on candidates the router refused. $A(\hat a)$ is the
screen-time reading; the gate is $0.65$. ``FN'' marks a false negative, a refused candidate whose gain the
full pipeline would have measured. Slice gains carry group-clustered $95\%$ intervals where a
grouping exists. Scorers differ by candidate (preference-margin, toxicity, and NLI/QA cross-encoders)
and are recorded per run.}
\label{tab:screenrecall}
\setlength{\tabcolsep}{6pt}
\renewcommand{\arraystretch}{1.15}
\begin{tabular}{llccl}
\toprule
Candidate & $A(\hat a)$ & Screen & Slice gain & Outcome \\
\midrule
SHP & 0.647 & \pass & $+0.147$ {\scriptsize$[+0.127,+0.170]$} & gain measured (FN) \\
HateCheck & 0.637 & \pass & $+0.108$ {\scriptsize$[+0.099,+0.118]$} & gain measured (FN) \\
MNLI (hyp.-only) & 0.616 & \pass & $+0.036$ {\scriptsize$[+0.028,+0.045]$} & gain measured (FN) \\
\addlinespace[2pt]
BoolQ & 0.640 & \fail & $+0.024$ {\scriptsize$[+0.014,+0.034]$} & refused correctly (screen) \\
PAWS & 0.538 & \fail & $+0.008$ {\scriptsize$[-0.014,+0.033]$} & refused correctly (screen) \\
RTE & 0.524 & \fail & $+0.025$ {\scriptsize$[+0.013,+0.037]$} & refused correctly (screen, decorr.\ unmet) \\
hh-rlhf & 0.513 & \fail & $-0.013$ {\scriptsize$[-0.023,-0.004]$} & refused correctly (screen) \\
\bottomrule
\end{tabular}
\end{table}

\section{The Reward-Model Code Audit: Construction and Protocol}
\label{app:rmcode}

\textbf{The row-level screen reading.} The screen value committed to disk before the reward model produced a single score was measured under row-level folds and read $0.398$; the below-chance reading is a mechanical artifact of row-level folds splitting format-identical twins with opposite labels: the held-out twin has the same $\phi$ as its trained partner and the opposite $y$, so the probe reproduces the partner's label and is wrong every time, which drives the reading below chance. The screen decision is the same under either protocol.

\textbf{Robustness runs.} Rescoring the same candidates with the larger checkpoint of the same family \emph{reverses the sign} of the bias: DeBERTa-v3-large-v2 penalizes comments, its mirrored cross-format contrast falls below chance raw ($0.462$), and residualization attenuates the format effect in this direction too and lifts the failing contrast to $0.620$. HumanEval replicates the routing refusal under the same frozen recipe, with $A(\hat a)=0.360$ committed before scoring. There the base model's format loading is extreme ($R^2_{\phi\to s}=0.418$, a format effect of $+1.24$ against a construct margin of $+0.04$), and raw, it prefers a commented buggy solution to a terse correct one $98.6\%$ of the time. Residualization attenuates the format channel and returns both cross-format contrasts to $0.466$ and $0.527$, intervals covering $0.5$, the level the construct margin supports.

\textbf{Length-only corrections in detail.} A correction \emph{nonlinear} in length is not bound by a single slope, so it could in principle absorb a non-monotone response. We refit the length-only correction as an isotonic regression, a cubic spline basis, and a gradient-boosted regressor on the same design: all four format effects remain separated from zero in every case, whereas the multivariate $\phi$ closes all four (Table~\ref{tab:nonlinlength}). The isotonic fit (increasing direction, the scikit-learn default) explains none of the score (cross-validated $R^2=-0.001$) while the linear fit explains $0.069$; despite that difference in predictive power, neither closes a single paired effect. The non-monotone response itself is measured by the edit-type comparison above, where the bare comment is penalized and the docstring rewarded; this table measures something narrower, that the tested length-only corrections do not close the resulting effects. Flexibility shrinks the damage without removing it, the docstring effect on correct code falling from $+0.62$ under the linear correction to $+0.48$ under the spline and $+0.33$ under boosting, against a raw $+0.08$. A length covariate cannot separate a bare comment line from a docstring of the same size, and that is the distinction the reward model is responding to.

\textbf{Construction (\texttt{rm\_code\_build.py}).} From the MBPP test split we keep problems whose canonical solution passes its own unit tests (all but 2). The buggy variant applies one AST-level mutation, a comparison-operator swap ($<\,\leftrightarrow\,\leq$, $=\,\leftrightarrow\,\neq$, \dots), an arithmetic swap ($+\leftrightarrow-$), or a small-integer increment, and is kept only if it compiles and \emph{fails} the tests (145 problems yield no failing single mutant and are dropped). The verbose variants add a docstring, inline comments, and a trailing comment line, and change nothing else; their labels are nevertheless re-verified by execution (correct-verbose must pass, buggy-verbose must fail). Every execution runs in a subprocess with a timeout. One construction pitfall matters for replication: mutants are regenerated through \texttt{ast.unparse}, which normalizes code style, and in a first build the canonical solutions kept their original formatting; the style difference alone let the format features detect mutation status ($A(\hat a)$ inflated to $0.638$). The fix is to round-trip \emph{all} candidates through the same parse/unparse normalization, after which the screen-time $A(\hat a)$ falls to $0.398$ (row-level folds; exactly $0.500$ under problem-grouped folds). Designed audits of this kind should expect and check for such channel leakage in the construction itself.

\begin{table}[!ht]
\centering
\small
\caption{Length-only corrections against the multivariate $\phi$ on the edit-type design, all cross-fitted on the same folds. $R^2$ is the share of the reward the correction explains. The four columns are the paired format effects, each a semantics-preserving edit against its terse twin, with problem-level bootstrap intervals; the last column counts effects whose residual interval covers zero, which is non-significance, not equivalence. The multivariate row's $R^2=0.285$ is computed on this design's sample, bare-comment cells included, which is why it differs from the audit's $0.283$ on the $2{\times}2$ cells alone (Section~\ref{sec:rmcode}). (\texttt{nonlinear\_length\_baseline.py}.)}
\label{tab:nonlinlength}
\resizebox{\textwidth}{!}{%
\begin{tabular}{lrllllc}
\toprule
Correction & $R^2$ & Bare comment, correct & Docstring, correct & Bare comment, buggy & Docstring, buggy & Residual CI covers 0 \\
\midrule
none (raw) & --- & $-0.227$ $[-0.268,-0.190]$ & $+0.083$ $[+0.004,+0.159]$ & $-0.199$ $[-0.239,-0.163]$ & $+0.154$ $[+0.083,+0.229]$ & $0/4$ \\
\midrule
length, linear & $0.069$ & $-0.170$ $[-0.208,-0.132]$ & $+0.620$ $[+0.544,+0.694]$ & $-0.141$ $[-0.179,-0.104]$ & $+0.682$ $[+0.607,+0.751]$ & $0/4$ \\
length, isotonic & $-0.001$ & $-0.226$ $[-0.266,-0.189]$ & $+0.084$ $[+0.009,+0.163]$ & $-0.198$ $[-0.236,-0.161]$ & $+0.151$ $[+0.081,+0.225]$ & $0/4$ \\
length, spline & $0.111$ & $-0.079$ $[-0.118,-0.041]$ & $+0.477$ $[+0.399,+0.561]$ & $-0.049$ $[-0.085,-0.011]$ & $+0.541$ $[+0.464,+0.619]$ & $0/4$ \\
length, boosted & $0.179$ & $-0.128$ $[-0.180,-0.077]$ & $+0.334$ $[+0.255,+0.418]$ & $-0.084$ $[-0.136,-0.034]$ & $+0.415$ $[+0.339,+0.491]$ & $0/4$ \\
\midrule
multivariate $\phi$ & $0.285$ & $-0.018$ $[-0.055,+0.018]$ & $-0.035$ $[-0.104,+0.042]$ & $+0.018$ $[-0.016,+0.052]$ & $+0.035$ $[-0.034,+0.109]$ & $\mathbf{4/4}$ \\
\bottomrule
\end{tabular}}
\end{table}

\textbf{Why the committed row-level $A(\hat a)$ reads below chance.} Each problem contributes a correct and a buggy candidate that are closely matched on the measured format features (the mean paired difference in character count is about $0.02$ characters) but carry \emph{opposite} labels. Under row-level folds one twin can land in training and the other in evaluation, so the probe learns that a format pattern means ``correct'' from one twin and is then scored on its sibling, where the same pattern means ``buggy.'' Systematically predicting the wrong label for the held-out twin drives the AUC below $0.5$. Grouping by problem keeps twins together, the anti-prediction disappears, and the probe reads chance, which is what the label-orthogonal design implies. The below-chance excursion is row-level twin leakage, an artifact of the fold assignment, and the routing decision is the same under either protocol.

\textbf{Why preservation is near-arithmetic.} That margin preservation is close to arithmetic has a one-line derivation. Within a format cell, the paired construct margin of any score $u$ is $m(u)=\mathbb{E}[u\mid\text{correct}]-\mathbb{E}[u\mid\text{buggy}]$, and residualization changes it by exactly the label contrast of the subtracted component: $m(\srep)=m(s)-\big(\mathbb{E}[\hat g(\phi)\mid\text{correct}]-\mathbb{E}[\hat g(\phi)\mid\text{buggy}]\big)$. The $2\times2$ design pairs every correct candidate with a buggy sibling in the same format cell whose format features are nearly identical (mean paired character-count difference $\approx0.02$ characters; problem-grouped $A(\hat a)$ exactly $0.500$), so the $\phi$-distribution is essentially label-invariant within cells, the label contrast of \emph{any} function of $\phi$ is $\approx0$, and the margin is preserved up to that residual difference plus cross-fitting noise. Margin preservation is due to the design rather than to the operator, so the audit's strongest evidence is the removal results and the cross-format rankings.

\textbf{Protocol (\texttt{rm\_code\_audit.py}).} $\phi$ is nine format-only features (character/line/token counts, comment count and density, docstring presence, blank lines, mean and max line length); A1 holds by construction here, since within every problem format is manipulated at fixed semantics. The router and its outcome were committed to a separate artifact before the reward model scored a single candidate, as in the held-out QA replications (Section~\ref{sec:positives}). The reward model scores (problem statement, code) pairs at 512 tokens, a limit no candidate reaches: the longest input in the design is 328 tokens (median 102, $95$th percentile 211), so nothing truncates, no correct/buggy pair collapses to identical model input, and the paired effects on the subset that provably fits are the reported ones (\texttt{truncation\_check.py}). The mutation is therefore inside the scoring window for every candidate. Residualization is the paper's unchanged cross-fitted ridge. Beyond Table~\ref{tab:rmcode}: the full-population AUC is $0.529$ raw vs.\ $0.530$ residualized (problem-clustered CI on the difference $[-0.001,+0.005]$), and within-cell preference accuracy (correct vs.\ buggy at \emph{matched} format) moves $0.752\to0.739$ under row-level folds. That $-1.3$-point cost is specific to those folds. Measuring the paired contrast directly (\texttt{paired\_g\_contrast.py}): $341$ of $353$ twin pairs have \emph{identical} $\phi$, so the subtraction is exactly neutral within them, and the fitted contrast $\hat g(\phi_{\text{correct}})-\hat g(\phi_{\text{buggy}})$ has mean $-0.0000$ and standard deviation $0.0008$ under problem-grouped folds. Under those folds the construct margins are unchanged to four decimals ($+0.1578$ terse, $+0.0874$ verbose), matched-format accuracy reads $0.752\to0.754$, and one pair of $706$ changes sign. The row-level drop is therefore the twin-leakage artifact of the fold assignment described above, not a cost of the operator.

\textbf{Robustness runs (\texttt{rm\_code\_extra.py}).} Table~\ref{tab:rmextra} reports the same objects for (i) the identical MBPP candidates rescored with OpenAssistant DeBERTa-v3-large-v2 (the router reading is a property of $(\phi,y)$ and carries over) and (ii) the HumanEval replication built by the same recipe ($146$ of $164$ problems, the $18$ dropped failing one of the recipe's mechanical requirements: a canonical solution failing its own tests, no verified failing single mutant, or a rewrite that broke label re-verification; $A(\hat a)=0.360$ committed before scoring). The format bias is checkpoint-specific in \emph{sign} (the base model rewards comments, the large model penalizes them), while residualization is sign-agnostic: it attenuates whichever loading is present and leaves the construct margins numerically unchanged, and the two cross-format contrasts, wildly asymmetric raw, converge after adjustment in both runs. Second, pooled $R^2_{\phi\to s}$ can understate a within-problem channel when between-problem score variance dominates (large-v2: pooled $R^2=0.023$ against a paired format effect of $-0.34$, attenuated by $+0.281$ $[+0.270,+0.291]$ after residualization); the paired contrasts are the right lens for a designed within-problem channel. On HumanEval the base model's construct margin is an order of magnitude smaller than its format effect, and the residualized contrasts settle near the level that margin supports, consistent with the adjustment attenuating the channel and adding no signal; the post-adjustment format effects ($+0.082$ and $+0.081$, intervals covering zero) remain imprecise relative to the $+0.034$ construct margin, so removal in an equivalence sense is not established here.

\textbf{Edit-type response (\texttt{rm\_code\_dose.py}).} Extending the design with an intermediate edit, a single bare comment line inserted after the function signature, labels re-verified by execution (all 353 problems retained), was intended as a dose-response check and instead revealed opposite-signed responses to bare comments and docstrings: the base model's raw format response is \emph{non-monotone across edit types}. The lone comment is penalized ($-0.227$ $[-0.268,-0.190]$ on correct code, $-0.199$ $[-0.239,-0.163]$ on buggy), while the docstring-style heavy edit is rewarded ($+0.083$/$+0.154$; Table~\ref{tab:rmcode}): what the model prices is format \emph{register} (docstring presence vs.\ bare comments), and text volume alone does not predict the response. We then measured the scalar alternative directly (\texttt{length\_only\_baseline.py}): residualizing on character count alone, a length-only analogue of the length-controlled-evaluation recipe \citep{dubois2024length} (which itself models pairwise preferences with a GLM), leaves every paired effect with a CI excluding zero and \emph{amplifies} the heavy-dose effect several-fold ($+0.083\to+0.620$ on correct code, $+0.154\to+0.682$ on buggy; token count in place of characters reads $+0.730$/$+0.794$), because the single fitted slope is a compromise across a non-monotone response and overcorrects the
cells it should leave alone. We report absolute changes rather than ratios throughout, because a
ratio is unstable when the baseline effect is small: the raw docstring effect on correct code is
$+0.083$ with an interval reaching $+0.004$. Residualization on the multivariate $\phi$ (which carries docstring presence and comment density as separate coordinates) flattens all four paired effects simultaneously to CIs covering zero ($-0.018$, $-0.035$, $+0.018$, $+0.035$); construct margins are unchanged under every residualizer, as the label-orthogonal design dictates. The residualizer inherits whatever structure $\phi$ measures, so the audit's feature inventory is where domain knowledge enters.

\textbf{A held-out edit template.} The verbose variants above all come from one comment/docstring
template, so the correction could be specific to that generation style. We built a second
template---a module-level comment before the definition, an \texttt{Args}/\texttt{Returns}
docstring, and a comment before the \texttt{return}, with no trailing banner---applied it to the same
$353$ problems' terse candidates, and re-verified every label by execution ($706$ candidates, none
dropped). The measured surface features barely move between the two templates (mean characters $474$
against $466$, lines $16.78$ against $16.78$, two comments and a docstring in both), but the reward
model's response reverses sign: the original template raises the reward by $+0.083$ on correct code,
the held-out template \emph{lowers} it by $-0.159$ $[-0.249,-0.071]$ ($-0.095$ $[-0.177,-0.013]$ on
buggy code). Applying the operator fit on the original template, without refitting, does not close
these effects; it overshoots them to $+0.195$ $[+0.114,+0.275]$ and $+0.259$ $[+0.184,+0.335]$, while
the construct margin is left unchanged ($+0.094$ before and after; \texttt{template\_holdout.py}).
Since $\phi$ does not separate the registers and the scorer responds to them in different
directions, no $\phi$-only corrector can be right about all of them: the same non-monotonicity that defeats
the scalar length baseline bounds the multivariate one as soon as the edit template changes. The attenuation reported in Section~\ref{sec:rmcode} is a statement about the template the
operator was fit on, not about comment/docstring edits in general.

\textbf{Two further templates.} A single held-out template leaves the reading resting on one
comparison, so we pre-specified and ran two more (\texttt{template\_generalization.py}). The templates,
the three analyses, and three predictions were fixed in a written note before any scoring;
that note is recorded with the code and is not third-party timestamped, the same status as
the other commitments in Appendix~\ref{app:accounting}. Template C attaches an end-of-line comment to each statement and adds no docstring;
template D adds a module-level docstring and blank-line padding and no comments at all. Both
move $\phi$ in directions the first two do not: C leaves the comment-line count and the
docstring indicator at zero while raising mean line length by $32.5$ characters, and D leaves
the comment count at zero while adding four blank lines. All $706$ candidates per template
survived label re-verification.

The raw effects differ in sign and by an order of magnitude across the four templates:
$+0.083$ on correct code under A, $-0.159$ under B, $-1.150$ under C, and $-0.340$ under D.
Table~\ref{tab:tmplgen} reports what the operator does to each held-out template. Fit on A
alone---the operator of Section~\ref{sec:rmcode}---it overshoots B, attenuates C from
$-1.150$ to $-0.185$, and \emph{amplifies} D from $-0.340$ to $-0.695$. Fitting on A and B
together makes both C and D worse rather than better. In every comparison the
raw-to-residualized change agrees to three decimals on correct and on buggy code, so what the
operator applies to an unseen template is close to a constant shift: $\phi$ separates the
templates from one another more than it tracks what the scorer responds to within them. The
construct margin moves by $+0.000$ in all nine comparisons, as the balanced design implies.
Two of the three pre-specified predictions held; the exception is recorded in
\texttt{notes/template\_generalization.json}.

\begin{table}[h]
\centering\small
\caption{Format effect on a held-out edit template, correct-code cell, raw and after applying
an operator that was never refit on it (problem-level bootstrap 95\% CIs, $n{=}353$ problems
per template). ``Fit A'' is the operator of Section~\ref{sec:rmcode}. Leave-one-out fits on
the other three templates. A dash marks a template that is inside the training pool for that
column.}
\label{tab:tmplgen}
\resizebox{\textwidth}{!}{%
\begin{tabular}{lcccc}
\toprule
Held-out template & Raw & Fit A & Fit A$+$B & Leave-one-out \\
\midrule
A (main run) & $+0.083$ {\scriptsize$[+0.011,+0.159]$} & --- & --- & $-0.144$ {\scriptsize$[-0.211,-0.069]$} \\
B & $-0.159$ {\scriptsize$[-0.249,-0.071]$} & $+0.195$ {\scriptsize$[+0.114,+0.275]$} & --- & $-0.049$ {\scriptsize$[-0.133,+0.032]$} \\
C & $-1.150$ {\scriptsize$[-1.233,-1.062]$} & $-0.185$ {\scriptsize$[-0.329,-0.035]$} & $-0.578$ {\scriptsize$[-0.703,-0.452]$} & $-0.200$ {\scriptsize$[-0.342,-0.056]$} \\
D & $-0.340$ {\scriptsize$[-0.408,-0.281]$} & $-0.695$ {\scriptsize$[-0.763,-0.635]$} & $-1.024$ {\scriptsize$[-1.095,-0.961]$} & $-0.574$ {\scriptsize$[-0.639,-0.519]$} \\
\bottomrule
\end{tabular}}
\end{table}

\begin{table}[!ht]
\centering
\small
\caption{Robustness runs for the reward-model code audit: a second checkpoint on the same MBPP candidates (left) and a second dataset with the main-run checkpoint (right). Same paired objects as Table~\ref{tab:rmcode}; problem-level bootstrap 95\% CIs.}
\label{tab:rmextra}
\setlength{\tabcolsep}{3.5pt}
\renewcommand{\arraystretch}{1.15}
\resizebox{\textwidth}{!}{%
\begin{tabular}{lcccc}
\toprule
& \multicolumn{2}{c}{MBPP $\times$ DeBERTa-v3-large-v2} & \multicolumn{2}{c}{HumanEval $\times$ DeBERTa-v3-base} \\
\cmidrule(lr){2-3}\cmidrule(lr){4-5}
Paired effect (within problem) & Raw & Residualized & Raw & Residualized \\
\midrule
Format ($+$comments), correct & $-0.335$ {\scriptsize$[-0.426,-0.248]$} & $-0.055$ {\scriptsize$[-0.142,+0.033]$} & $+1.241$ {\scriptsize$[+1.138,+1.347]$} & $+0.082$ {\scriptsize$[-0.017,+0.182]$} \\
Format ($+$comments), buggy & $-0.245$ {\scriptsize$[-0.327,-0.162]$} & $+0.041$ {\scriptsize$[-0.041,+0.125]$} & $+1.238$ {\scriptsize$[+1.130,+1.346]$} & $+0.081$ {\scriptsize$[-0.021,+0.185]$} \\
Construct (correct$\,-\,$buggy), terse & $+0.241$ {\scriptsize$[+0.195,+0.294]$} & $+0.240$ {\scriptsize$[+0.188,+0.289]$} & $+0.039$ {\scriptsize$[+0.021,+0.058]$} & $+0.034$ {\scriptsize$[+0.013,+0.059]$} \\
Construct (correct$\,-\,$buggy), verbose & $+0.150$ {\scriptsize$[+0.117,+0.182]$} & $+0.144$ {\scriptsize$[+0.113,+0.178]$} & $+0.041$ {\scriptsize$[+0.024,+0.061]$} & $+0.034$ {\scriptsize$[+0.014,+0.056]$} \\
\midrule
$P(\text{terse correct} \succ \text{verbose buggy})$ & $0.683$ {\scriptsize$[0.635,0.731]$} & $0.569$ {\scriptsize$[0.518,0.618]$} & $0.014$ {\scriptsize$[0.000,0.034]$} & $0.466$ {\scriptsize$[0.384,0.555]$} \\
$P(\text{verbose correct} \succ \text{terse buggy})$ & $0.462$ {\scriptsize$[0.408,0.513]$} & $0.620$ {\scriptsize$[0.569,0.671]$} & $0.980$ {\scriptsize$[0.952,1.000]$} & $0.527$ {\scriptsize$[0.445,0.610]$} \\
\bottomrule
\end{tabular}}
\end{table}

\section{Identifiability: Why \emph{Perfect} Collinearity Is Unrepairable}
\label{app:identifiability}

This appendix derives the collinearity boundary behind the screening heuristic of Section~\ref{sec:condition} in the linear-Gaussian population model that also generates Figure~\ref{fig:rho}, making precise why the crossing exists and why refusal at high collinearity is forced by the geometry.

\textbf{Model.} Let $c$ (construct) and $a$ (artifact channel) be standardized latent variables with $\mathrm{corr}(c,a)=\rho\in[0,1)$, and let the score be
\begin{equation}
s \;=\; \gamma c + \beta a + \varepsilon, \qquad \gamma>0,\ \varepsilon\perp(c,a),\ \mathrm{Var}(\varepsilon)=\sigma^2 .
\end{equation}
\textbf{Generative model, stated explicitly.} Four objects are distinct: the latent construct $c$; the artifact channel $a$ (correlation $\rho$ with $c$); the score $s=\gamma c+\beta a+\varepsilon$ (whatever benchmark labels trained the scorer enter only through this provenance); and the \emph{evaluation} measurement $y$, a noisy reading of $c$ whose error is assumed independent of $a$ (A2). Under independent noise every correlation with $y$ below is an attenuated version of the corresponding correlation with $c$, leaving all signs and crossings intact, and Appendix~\ref{app:corrnoise} measures what happens when that independence fails. Assume first that the artifact features recover the channel exactly: $\hat a(\phi)=a$ (idealized recoverability, $r{=}1$; relaxed below). We analyze the \emph{ideal population OLS reference}: residualization that subtracts the best linear predictor of $s$ from $a$. This is an idealization chosen for its exact algebra. It is the large-sample limit of the deployed ridge only in a regime this paper does not operate in: at fixed $\alpha=1$ and fixed dimension the effective penalty $\alpha/n$ vanishes and the estimator converges to the OLS projection, whereas the deployed $\phi$ has thousands of coordinates at $n$ in the thousands, so the fitted operator stays visibly shrunken toward zero. That gap, not an asymptotic argument, is why the deployed check must be an empirical criterion (Section~\ref{sec:method}):
\begin{equation}
\srep \;=\; s - \frac{\mathrm{Cov}(s,a)}{\mathrm{Var}(a)}\,a
\;=\; s - (\gamma\rho+\beta)\,a
\;=\; \gamma\,(c-\rho a) + \varepsilon .
\end{equation}
Decomposing $c=\rho a+\sqrt{1-\rho^2}\,c_\perp$ with $c_\perp\perp a$ gives the central identity:
\begin{equation}
\boxed{\;\srep \;=\; \gamma\sqrt{1-\rho^2}\;c_\perp \;+\; \varepsilon\;}
\label{eq:identity}
\end{equation}

\textbf{Property 1 (index decorrelation is unconditional).} $\mathrm{Cov}(\srep,a)=0$ at \emph{every} $\rho$: the linear artifact floor of the residualized score drops to chance regardless of whether the adjustment preserved anything. This is the formal reason the decorrelation diagnostic alone cannot justify adjustment (Figure~\ref{fig:rho}b) and the slice-gain test is required.

\textbf{Property 2 (construct retention decays as $1-\rho^2$).} From Eq.~(\ref{eq:identity}),
\begin{equation}
\mathrm{corr}(\srep,c) \;=\; \frac{\gamma(1-\rho^2)}{\sqrt{\gamma^2(1-\rho^2)+\sigma^2}} ,
\end{equation}
which is strictly decreasing in $\rho$ and reaches $0$ at $\rho=1$: the residualized score becomes pure noise. The construct-carrying variance that survives adjustment is exactly the component of $c$ orthogonal to the artifact; collinearity leaves nothing to survive.

\textbf{Property 3 (the boundary is an identification failure).} At $\rho=1$ the construct and artifact directions coincide ($c=a$), so the raw score $s=(\gamma+\beta)a+\varepsilon$ is itself correlated with the construct; what fails at the boundary is \emph{separation}, and construct information is still present. By the identity $c=a$, any functional of $(s,\phi)$ is exactly as correlated with the construct as with the artifact: shedding the linear artifact correlation sheds the linear construct correlation with it, and a functional made fully \emph{independent} of $a$ is independent of $c$ and carries no construct alignment at all. Between those poles, a nonlinear functional can be uncorrelated with $a$ yet still carry construct-ordering information (the relocation phenomenon of Appendix~\ref{app:certprobes} is the measured analogue), so the impossibility claim is stated for linear correlation and for full independence, not for every nonlinear functional. The deeper point is identification: at $\rho=1$ nothing observable separates construct from artifact, so no procedure can \emph{certify} preservation while shedding the channel. Refusal is therefore the correct outcome at the boundary.

\textbf{Property 4 (raw vs.\ adjusted, and the crossing).} Raw alignment is governed by $\mathrm{corr}(s,c)=(\gamma+\beta\rho)/\mathrm{sd}(s)$: the artifact term $\beta\rho$ \emph{helps} the raw score wherever the artifact agrees with the construct, which is why raw full-population alignment can exceed the residualized score's (the trade-off reported for SNLI and SICK). On a target slice where the surface-only predictor errs, the artifact--construct correlation is locally negative, and in this model it follows directly: with the label $y=\mathbf{1}[c>0]$ and the (perfect-recovery) surface decision $\mathbf{1}[a>0]$ thresholded at the common median, the predictor errs exactly when $\operatorname{sign}(a)\neq\operatorname{sign}(c)$, i.e.\ on the event $\{ac<0\}$, so the product $ac$ is negative pointwise on the slice. By the sign symmetry of the zero-mean bivariate Gaussian under that event, the conditional means of $a$ and of $c$ both remain zero, so $\mathbb{E}[ac \mid ac<0]<0$ is itself the conditional covariance, and conditioning on the error event forces a negative conditional correlation for \emph{any} unconditional $\rho<1$ (imperfect recovery and off-median thresholds blur the event's boundary; we do not prove that the sign is preserved there, and the claim below is stated for the idealized event). This is the formal content of the claim, used throughout, that conditioning on surface-predictor errors favors subtraction. On that slice the raw score pays $\beta$ as a penalty while $\srep$, whose \emph{unconditional} covariance with $a$ is zero, does not; conditioning on the slice does not preserve that orthogonality, which the next paragraph takes up. Comparing the two correlations gives the crossing explicitly. Write the raw score's artifact contribution as $\beta\rho/\mathrm{sd}(s)$ with $\mathrm{sd}(s)=\sqrt{\gamma^2+\beta^2+2\gamma\beta\rho+\sigma^2}$. The adjustment improves full-population construct alignment exactly when
\begin{equation*}
\frac{\gamma(1-\rho^{2})}{\sqrt{\gamma^{2}(1-\rho^{2})+\sigma^{2}}}>\frac{\gamma+\beta\rho}{\sqrt{\gamma^{2}+\beta^{2}+2\gamma\beta\rho+\sigma^{2}}},
\end{equation*}
so, squaring both sides (both are positive on the relevant range), the crossing $\rho^{\ast}$ is, for the parameter regime used in our sweep, the unique root in $(0,1)$ of $\gamma^{2}(1-\rho^{2})^{2}(\gamma^{2}+\beta^{2}+2\gamma\beta\rho+\sigma^{2})=(\gamma+\beta\rho)^{2}(\gamma^{2}(1-\rho^{2})+\sigma^{2})$. At the sweep's parameters ($\gamma{=}1$, $\beta{=}1.2$, $\sigma{=}0.3$) this gives $\rho^{\ast}=0.489$, matching the measured crossing of Figure~\ref{fig:rho}a and Table~\ref{tab:rhoscreen}. Because the slice-eligibility check is computed from $\phi$ and $y$ alone, its own crossing does not move with $\beta$, so the two crossings are ordered differently for different loadings: $\rho^{\ast}$ falls to $0.363$ at $\beta{=}0.6$ and $0.281$ at $\beta{=}0.4$, below the slice-eligibility check's.

\textbf{What a positive $G$ does not identify.} The component this model subtracts is $(\gamma\rho+\beta)a$, not the artifact term $\beta a$: whenever the construct correlates with the artifact it removes construct-derived variance as well. The extreme case makes the consequence concrete. Set $\beta=0$, so the score carries no artifact loading at all ($s=\gamma c+\varepsilon$), and keep $\gamma=1$, $\rho=0.8$, $\sigma=0.3$. Residualization still subtracts $\gamma\rho\,a$, which raises the score of slice positives ($c>0$, $a<0$) and lowers it for slice negatives, so slice alignment rises ($0.903\to0.987$ at $N=5{\times}10^5$) while full-population alignment falls ($0.973\to0.748$; \texttt{g\_identification\_counterexample.py}). A positive $G$ is therefore not an artifact-specific quantity even here, which is why Section~\ref{sec:attrdef} reports it as a descriptive change in slice AUC and nothing more.

\textbf{Imperfect recovery (one route to an eligibility failure).} Write $\tilde a = r\,a + \sqrt{1-r^2}\,u$ with $u$ independent of $(a,c,\varepsilon)$ and $\mathrm{corr}(\tilde a,a)=r<1$. This $\tilde a$ is \emph{not} the pipeline's $\hat a$: it is an idealized estimate of the latent artifact channel, used only to study imperfect channel recovery, whereas $\hat a$ is the label-mediated predictor of $y$ defined in Section~\ref{sec:setup}, so the slice-eligibility check statistic $A(\hat a)$ additionally depends on how the recovered channel predicts $y$. Replacing $a$ by $\tilde a$ above shows residualization removes only the recovered component: the floor closes only against predictors of $\tilde a$, and under this recovery model a fraction $1-r^2$ of the artifact variance survives in $\srep$. As $r\to0$ the population component shrinks toward zero while the finite-sample fit still subtracts an estimated one; the operator then subtracts mostly estimation noise (a finite-sample statement, since the population coefficient on an unrelated feature is zero), cannot help, and, through the variance it removes, can hurt. This is one of the two routes to an eligibility failure, and the only one that indicts the \emph{score}: the slice-eligibility check can equally fail with $r$ large when the recovered channel is orthogonal to the label ($\rho\approx0$), where removal is warranted and what is missing is the label-defined measurement target (Section~\ref{sec:synthetic}). $R^2_{\phi\to s}$ separates the two. It is also the mechanism behind the nonlinear-artifact refusals (HANS syntactic heuristics, Appendix~\ref{app:nonlinear}).

\textbf{The full-population cost's role.} A negative full-population change is the arithmetic consequence of subtracting a component that partly agrees with the construct, and it appears in every observational positive in this paper (the designed audit's pooled AUC is flat by construction, $0.529\to0.530$); a gate on the sign would refuse settings where the paired contrasts, not the pooled AUC, carry the effect. The magnitude that would separate the expected trade from destruction depends on the construct--artifact collinearity, which no observational statistic in this paper identifies, so a numerical threshold on the cost would be a hidden assumption about $\rho$. The consumer who holds the loss function alone can price the trade, which is why we report the pair together and leave any stopping rule on the consumer's side.

\section{Screen Operationalization and Deployment-Rule Negatives}
\label{app:c2}

Table~\ref{tab:c2} reports the two screen statistics of Section~\ref{sec:condition} for the (setting, scorer) pairs where the declared slice is well defined and the cached scores permit the computation, using the identical operator recipe everywhere (plain ridge, $\alpha{=}1$, 5-fold shuffled cross-fitting). The conjunction separates the positives from the negatives, and the two failure modes are instructive: QQP's in-distribution score is strongly $\phi$-predictable ($R^2_{\phi\to s}=0.287$) yet loses almost nothing where the artifact is wrong ($\Delta_{\mathrm{slice}}=0.005$): the artifact loading is construct-aligned, so there is no artifact-associated error to remove; MNLI's off-distribution score is degraded on the declared slice ($\Delta_{\mathrm{slice}}=0.030$) but barely rides the linear channel ($R^2_{\phi\to s}=0.007$): its errors are not linearly $\phi$-attributable, so residualization has nothing it is warranted to remove. The margins are finite (toxicity sits at $0.040/0.025$ against gates of ${\approx}0.05/0.03$), so the screen remains a heuristic: the binding decision is always the slice-gain test, which it merely predicts. \emph{Chronology:} the gate values were fixed from the first five settings (the NLI family and toxicity) and applied unchanged to WikiQA, which was selected and scored afterwards, an out-of-sample application in time. \emph{Leave-one-setting-out:} removing any single setting and re-choosing gates anywhere in the remaining settings' gaps reproduces every outcome, since no setting's statistics sit between the positives and the negatives of the others; the toxicity margin above is the binding one, and a gate anywhere in $(0.040,0.100)\times(0.025,0.032)$ gives identical results.

\begin{figure}[t]
\centering
\includegraphics[width=0.8\textwidth]{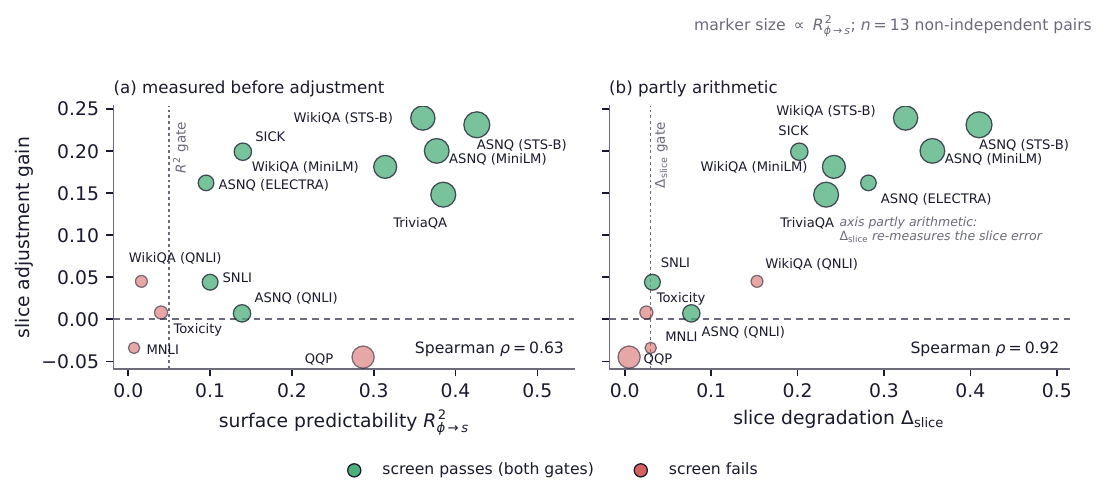}
\caption{Observed slice gain (AUC) against each screen statistic for the thirteen (setting, scorer) pairs with a well-defined slice, question-grouped protocol. \textbf{(a)} against the loading $R^2_{\phi\to s}$, measured before adjustment. \textbf{(b)} against the slice degradation $\Delta_{\mathrm{slice}}$; the tighter correlation is partly arithmetic (in-panel note). Marker size: $R^2_{\phi\to s}$. Green: screen pass; red: screen fail. The WikiQA and ASNQ scorer replications share their dataset and slice, so points are not independent samples.}
\label{fig:repairability}
\end{figure}

\textbf{The relationship behind the gates is continuous.} Figure~\ref{fig:repairability} plots the observed adjustment gain against each screen statistic in turn: panel (a) against $R^2_{\phi\to s}$, panel (b) against $\Delta_{\mathrm{slice}}$, with marker size encoding $R^2_{\phi\to s}$ in both for all settings and scorers under the grouped protocol (WikiQA (STS-B)'s $\Delta_{\mathrm{slice}}$ is the row-level value, the one on record). The predictive relationship is gain vs.\ $R^2_{\phi\to s}$, Spearman $\rho=0.63$ over the thirteen pairs and $0.55$ collapsed to one scorer per setting (descriptive in both forms; the replications share datasets and slices, and eight independent points carry no significance). Gain vs.\ $\Delta_{\mathrm{slice}}$ reads $\rho=0.92$, but part of that is arithmetic ($\Delta_{\mathrm{slice}}$ re-measures the slice error the gain is computed on), so we treat it as internal consistency only. The separation of screen-passing from screen-failing points has two marginal exceptions, the two QNLI replications, and they err in opposite directions. WikiQA (QNLI) falls below the $R^2$ gate under the grouped protocol ($R^2_{\phi\to s}=0.016$) yet retains a small positive slice gain ($+0.045$, CI $[+0.027,+0.062]$): a conservative refusal of a mildly beneficial adjustment. ASNQ (QNLI) passes both gates yet gains nothing ($+0.007$, n.s.): the removal happens (the fitted component reaches an out-of-fold predictive $R^2_{\phi\to s}=0.139$\footnote{Numerically coincident with the ASNQ primary scorer's residual index correlation of $0.139$ (Section~\ref{sec:samelabel}); the two are different quantities, of different scorers, that round to the same value.}) but buys nothing on the slice, because the subtracted component is not misaligned with the construct where the surface predictor errs (raw index correlation $R_{\mathrm{lin}}(s)=0.053$; the two can diverge because $R^2_{\phi\to s}$ measures loading on all of multi-dimensional $\phi$ while $R_{\mathrm{lin}}$ measures correlation with the single label-predictive direction $\hat c$, Section~\ref{sec:setup}). The gates are thus a coarse, pre-specified binarization of a continuous relationship, used only because a decision needs a yes/no; they err in both directions at the margin, and the continuous statistics carry the predictive content. The scorer replications add two controlled comparisons: with everything but the scorer fixed, the gain rises with $\Delta_{\mathrm{slice}}$, the primary axis; ELECTRA is the instructive off-diagonal point ($R^2_{\phi\to s}=0.095$ yet $\Delta_{\mathrm{slice}}=0.282$, gaining $+0.162$). The relationship comes with cautions. Part of it is near-arithmetic: given a screen pass, de-loading raises slice alignment mechanically, and $\Delta_{\mathrm{slice}}$ re-measures the same slice degradation, so the dose--response speaks to internal consistency and carries no out-of-sample prediction. The non-mechanical part is whether de-loading survives construct entanglement: the synthetic crossing and the refused settings (QQP loads on the channel yet its slice gain is negative) show that loading alone does not produce a gain: the gain's magnitude records the AUC change on the slice associated with subtracting the fitted component.

Table~\ref{tab:threepop} makes the three-population picture explicit for every positive: the label-defined slice is the only population that improves, which is the operational-deployment negative result of Section~\ref{sec:positives} in tabular form.

\textbf{A rule-based overlap intervention (label fixed by construction).} The A1 question for the QA settings (is the overlap channel construct-external, or genuine relevance signal?) cannot be settled observationally, so we intervene. On the held-out SQuAD candidates, two deterministic edits manipulate overlap while the containment label is fixed \emph{by construction}: \emph{overlap-down} replaces question-overlapping words in answer-bearing sentences with WordNet synonyms, never touching the answer span (the answer remains verbatim, so $y{=}1$ is preserved; $n{=}503$, mean overlap change $-0.075$); \emph{overlap-up} appends a short clause of question content words to non-answer sentences (the answer string remains absent, so $y{=}0$ is preserved; $n{=}599$, $+0.062$). The frozen operator (ridge fit once on the original candidates) is applied to original and edited text alike. The raw cross-encoder follows the manipulation strongly, mean score change $-5.02$ (SE $0.16$) under overlap-down and $+8.57$ under overlap-up, despite the label being unchanged: a direct, interventional demonstration that the score rides the overlap channel as such. The residualized score's sensitivity is largely removed under the naturalistic down-edit ($-0.66$, an $87\%$ attenuation) and partially removed under the cruder up-edit ($+4.24$, $51\%$; the injected words also enter the answer-side TF-IDF and the edit changes phrasing beyond the linear overlap feature, so full attenuation is not expected). This does not prove A1 globally; it establishes that, \emph{in the manipulated direction}, the component the operator subtracts behaves like overlap and not like the preserved label (\texttt{overlap\_intervention.py}). What the edits hold fixed is the containment proxy, not naturalness. A blinded automatic spot-check (120 edited sentences and their originals, presented in random order to an LLM judge that never sees the labels; \texttt{overlap\_edit\_llm\_check.py}) agrees with the construction: the judge's answer-containment outcome is unchanged by the edit on 93\% of overlap-down cases and 98\% of overlap-up cases. It also records what the construction does not hold fixed: judged fluency falls from 4.8 to 2.7 on a 1--5 scale under the synonym substitutions and from 4.1 to 1.4 under the appended clause, so the raw-versus-residual contrast on edited text is read as an overlap-plus-fluency
manipulation, and ``label fixed'' means the containment label. The intervention therefore
confirms that the score is sensitive to an overlap-changing edit at a fixed containment
label; because fluency also changes substantially, it does not isolate lexical overlap as
the causal factor. This is an automatic check, not a human adjudication.

\textbf{A systematic (negative) search for an observable deployment rule.} We pre-declared a twelve-member family of $\phi$-only rules: global shrinkage $s-\lambda\hat g$ ($\lambda\in\{0,.25,.5,.75,1\}$); switches applying the residualized score where $|\hat g-\mathrm{med}|$ exceeds a percentile threshold ($\{50,70,80,90\}$); and switches applying it in extreme $\hat c$ bands ($\{10,20,30\}\%$). Each family's parameter was selected on WikiQA$+$ASNQ by full-population delta. The selected members were then evaluated frozen on SQuAD and DuoRC. The shrinkage family's optimum is $\lambda{=}0$ (the raw score); the best switch members still cost $0.075$ to $0.082$ of full-population AUC on the held-out settings, with query-clustered CIs entirely below zero (\texttt{obs\_rule\_search.py}). Within this family, no $\phi$-only rule improves an observable population, the operational-deployment negative of the main text, now with a search behind it.

\begin{table}[!ht]
\centering\small
\caption{Every frozen $\phi$-only deployment rule on the held-out settings: change in full-population AUC against the raw score, with query-clustered 95\% intervals. Parameters were selected on the development settings (WikiQA, ASNQ) and then frozen; nothing was re-selected after seeing these columns. Rows 1--2 are the original switch family, rows 3--6 the scale-matched family. The shrinkage family's optimum is $\lambda{=}0$, the raw score itself, and is omitted. All values use question-grouped cross-fitting (\texttt{deployment\_grouped.py}, \texttt{rule\_search\_rescaled.py}).}
\label{tab:deployrules}
\setlength{\tabcolsep}{6pt}
\renewcommand{\arraystretch}{1.15}
\begin{tabular}{llcc}
\toprule
Frozen rule & Parameter & SQuAD $\Delta$AUC & DuoRC $\Delta$AUC \\
\midrule
$|\hat g|$-switch & $t{=}90$ & $-0.079$ {\scriptsize$[-0.087,-0.069]$} & $-0.073$ {\scriptsize$[-0.086,-0.061]$} \\
$\hat c$-band switch & $q{=}10$ & $-0.084$ {\scriptsize$[-0.096,-0.074]$} & $-0.083$ {\scriptsize$[-0.099,-0.069]$} \\
\addlinespace[2pt]
$|\hat g|$-switch, rescaled & $t{=}90$ & $-0.063$ {\scriptsize$[-0.072,-0.054]$} & $-0.056$ {\scriptsize$[-0.071,-0.041]$} \\
$\hat c$-band switch, rescaled & $q{=}10$ & $-0.069$ {\scriptsize$[-0.078,-0.060]$} & $-0.058$ {\scriptsize$[-0.072,-0.044]$} \\
calibrated blend & $\lambda{=}0.25$ & $-0.001$ {\scriptsize$[-0.003,+0.001]$} & $-0.009$ {\scriptsize$[-0.013,-0.005]$} \\
query-conditioned & $0.5$ & $-0.017$ {\scriptsize$[-0.021,-0.014]$} & $-0.018$ {\scriptsize$[-0.023,-0.012]$} \\
\bottomrule
\end{tabular}
\end{table}

\textbf{An observable risk flag: triage without adjustment.} The same search motivates a weaker but achievable deployment deliverable: leave the score unchanged and predict \emph{where it is artifact-suspect}. A logistic flag over deployment-observable, per-setting-standardized features ($\hat c$ and its extremity, $\hat g$ and its extremity, $s$, $s{-}\hat g$, $\hat a$) is fit on the pooled development settings (WikiQA$+$ASNQ; slice labels used only here) and frozen. On the three non-development settings it predicts true slice membership at AUC $0.809$ (SQuAD; query-clustered 95\% CI $[0.800,0.817]$), $0.913$ (DuoRC; $[0.905,0.922]$), and $0.739$ (TriviaQA; $[0.731,0.747]$); flagging the top quintile concentrates slice members at precision $0.422$ $[0.401,0.447]$ vs.\ base rate $0.184$ (SQuAD), $0.375$ $[0.348,0.403]$ vs.\ $0.085$ (DuoRC), and $0.546$ $[0.527,0.565]$ vs.\ $0.296$ (TriviaQA), a $1.8$--$4.4\times$ lift, with top-decile behavior heterogeneous across settings ($1.2\times$--$6.0\times$). The flag inputs are label-free \emph{for the example}: $\hat c$ and $\hat a$ are cross-fitted, so no candidate's own label enters its features, the same epistemic status as the observable $z{=}1$ subgroup used throughout, while the per-setting apparatus is label-fit as elsewhere in the paper. We therefore use the flag only for triage: it prioritizes examples for re-annotation or dual reporting, and makes no claim about the score's value on the flagged examples (\texttt{obs\_risk\_flag.py}).

\textbf{The flag as a switch: a measured negative.} The one natural rule missing from the twelve-member family above is to use the flag itself as a \emph{switch}: score with $\srep$ on the top flagged quantile and with raw $s$ elsewhere ($u_q=\srep$ if the flag probability is in the top $q\%$, else $s$). Acting on that localization is the natural next step, and Table~\ref{tab:switch} measures it on the held-out settings, with the frozen dev-fit flag and $q\in\{10,20,30,50\}$ (\texttt{risk\_flag\_switch.py}; cross-fitting is row-level throughout this appendix's rule-search suite as recorded, while all quoted CIs remain query-clustered: the two are independent protocol choices, and the frozen artifacts are re-evaluated under question-grouped folds below). Every tested switch reduces full-population AUC relative to raw, and none reaches full residualization's slice AUC: on the full population it is strictly worse than raw at every quantile (e.g.\ SQuAD $0.932\to0.911$ at $q{=}10$), and on the true slice it recovers only a fraction of full residualization's gain; at small $q$ it can fall \emph{below} raw even there (DuoRC $q{=}10$: $0.808\to0.772$). Two plausible contributors follow from the construction: the flagged quantile mixes artifact-right with artifact-wrong examples (the same mixture that defeats the $z{=}1$ subgroup), and the switch concatenates two scores that are not scale-commensurate ($\srep$ is a residual, $s$ a logit), so cross-boundary comparisons are corrupted, a cost paid by every switch rule we measured, though per-population standardization of $s$ and $\srep$ could in principle reduce it. The switch the flag suggests is worse than raw on the full population at every quantile and never matches full residualization on the slice; the flag remains suitable only for triage.

\textbf{Grouped re-evaluation of the frozen artifacts.} The suites above were recorded with row-level cross-fitting. Re-evaluating them under question-grouped folds---the rule parameters frozen at their recorded values, the flag's feature specification frozen and its logistic fit re-run once on the pooled development settings---reproduces every negative result: the frozen switch rules lose $0.073$--$0.084$ full-population AUC on SQuAD and DuoRC and $0.027$--$0.034$ on TriviaQA, all query-clustered intervals below zero (TriviaQA gains an interval this suite had not recorded), and the flag reads AUC $0.809/0.911/0.737$ with top-quintile lift $2.3/4.4/1.8$ on SQuAD/DuoRC/TriviaQA. The row-level pipeline first reproduced the recorded numbers to the fourth decimal (\texttt{deployment\_grouped.py}).

\begin{table}[h]
\centering\small
\caption{The risk flag used as a switch ($u_q$: residualized score on the top flagged $q\%$, raw elsewhere), held-out settings, full population / true label-defined slice (\texttt{risk\_flag\_switch.py}; row-level cross-fitting, as in the flag suite, so endpoint values differ from the grouped headline at the third decimal).}
\label{tab:switch}
\begin{tabular}{lcccccc}
\toprule
 & raw & $q{=}10$ & $q{=}20$ & $q{=}30$ & $q{=}50$ & full $\srep$ \\
\midrule
SQuAD full & $0.932$ & $0.911$ & $0.910$ & $0.905$ & $0.887$ & $0.835$ \\
SQuAD slice & $0.706$ & $0.755$ & $0.754$ & $0.764$ & $0.807$ & $0.883$ \\
DuoRC full & $0.951$ & $0.939$ & $0.930$ & $0.921$ & $0.911$ & $0.871$ \\
DuoRC slice & $0.808$ & $0.772$ & $0.781$ & $0.803$ & $0.841$ & $0.869$ \\
TriviaQA full & $0.706$ & $0.703$ & $0.701$ & $0.702$ & $0.701$ & $0.636$ \\
TriviaQA slice & $0.474$ & $0.501$ & $0.481$ & $0.466$ & $0.487$ & $0.626$ \\
\bottomrule
\end{tabular}
\end{table}

\textbf{Clustered uncertainty, per-query metrics, and the nested protocol (QA settings).} Because candidates within a question are dependent, we recompute the slice gain CIs with a \emph{query-clustered} bootstrap (resampling questions, 1{,}000 resamples): WikiQA $[+0.148,+0.207]$ (792 queries), ASNQ $[+0.222,+0.270]$ (390), TriviaQA $[+0.137,+0.166]$ (2{,}816). These bracket the \emph{row-level} gains recorded at selection time (this appendix suite is row-level throughout); the grouped headline gains differ (ASNQ $+0.200$, CI $[+0.180,+0.219]$) and carry their own clustered CIs in Section~\ref{sec:positives}. No conclusion changes under either protocol. Per-query ranking metrics show the same full-population trade-off as pooled AUC: per-query AUC and MRR fall from raw to adjusted on the full sets (e.g.\ WikiQA CE $0.877\to0.797$ AUC, $0.812\to0.691$ MRR), which is the expected cost of removing a channel that partially agrees with the construct on the full distribution; the claim adjustment earns is slice-scoped only (Section~\ref{sec:discussion}). The nested split-half protocol of Appendix~\ref{app:slicebias}, extended to the QA settings (slice defined on one half, adjustment cross-fitted and evaluated on the other, symmetrized; an independent implementation with its own random split, hence WikiQA's per-half values differ from the Appendix~\ref{app:slicebias} run), preserves every positive: WikiQA $+0.181/{+}0.154$, ASNQ $+0.239/{+}0.220$, TriviaQA $+0.150/{+}0.161$.

\textbf{Feature-split ablation: the QA channel is overlap-led.} Splitting $\phi$ and adjusting with each part alone (same fixed slice) shows the QA gains are carried overwhelmingly by the question--answer overlap features, with the answer-only surface contributing little: overlap-only adjustment recovers $+0.171$/$+0.272$/$+0.172$ (WikiQA/ASNQ/TriviaQA CE) against $+0.019$/$+0.039$/$+0.091$ for answer-only adjustment, with the full-$\phi$ gains at $+0.177$/$+0.247$/$+0.152$ (row-level fits; the grouped full-$\phi$ values are the main text's, Appendix~\ref{app:grouped}). This sharpens the scope of the QA positives, and one caveat applies in reading it: because the slice is defined by the \emph{same} predictor's errors, ``overlap is misleading on the slice'' is true by definition, so the result cannot by itself establish that overlap is construct-external (assumption A1); that would be circular. What the ablation does establish is descriptive and still useful: the measured gain is a \emph{conditional} improvement in agreement with the human relevance labels on the label-defined slice, and it is the overlap channel whose removal produces it. The precise reading of the QA positives is this: a post-hoc, slice-conditional correction of an overlap-riding score, reported with its full-population cost, under an A1 that remains an assumption. (Script: \texttt{review\_r2\_experiments.py}.)

\begin{table}[h]
\centering\small
\caption{Screen statistics under the uniform operator recipe (row-level fits, as elsewhere in this appendix; the grouped main-pipeline values are in Table~\ref{tab:grouped}). Gates: $R^2_{\phi\to s}\gtrsim0.05$, $\Delta_{\mathrm{slice}}\gtrsim0.03$. $^{\S}$WikiQA (QNLI): $A(\hat a)$ and $R^2_{\phi\to s}$ shown grouped (authoritative; the row-level fit read $0.082$, a pass: the one outcome the protocol change flips); $\Delta_{\mathrm{slice}}$ remains row-level as recorded. The gain is positive under both protocols ($+0.045$ grouped, CI $[+0.027,+0.062]$), the counterexample discussed at Figure~\ref{fig:repairability}.}
\label{tab:c2}
\begin{tabular}{lccccl}
\toprule
Setting & $A(\hat a)$ & $R^2_{\phi\to s}$ & $\Delta_{\mathrm{slice}}$ & Screen & Observed outcome \\
\midrule
SNLI (off-dist) & 0.708 & 0.100 & 0.032 & \pass & adjusted; slice gain $+0.044$ \\
SICK (off-dist, prospective) & 0.727 & 0.140 & 0.202 & \pass & adjusted; slice gain $+0.199$ \\
WikiQA (off-dist, screened) & 0.747 & 0.435 & 0.240 & \pass & adjusted; slice gain $+0.181$ \\
WikiQA (QNLI scorer) & 0.748 & 0.016 & 0.153$^{\S}$ & \fail & refused; gain $+0.045$ \\
WikiQA (STS-B scorer) & 0.747 & 0.463 & 0.324 & \pass & adjusted; slice gain $+0.238$ \\
ASNQ (off-dist, screened) & 0.747 & 0.514 & 0.363 & \pass & adjusted; slice gain $+0.247$ \\
TriviaQA (off-dist, screened) & 0.675 & 0.418 & 0.232 & \pass & adjusted; slice gain $+0.152$ \\
MNLI (off-dist) & 0.617 & 0.007 & 0.030 & \fail & refused ($-0.034$) \\
QQP (in-dist) & 0.636 & 0.287 & 0.005 & \fail & refused ($-0.045$) \\
Toxicity (in-dist) & 0.557 & 0.040 & 0.025 & \fail & refused (n.s.) \\
\bottomrule
\end{tabular}
\end{table}

\section{The Question-Grouped Protocol, Score Representation, and Query-Level Ranking}
\label{app:grouped}

\paragraph{Full-procedure stability.} The intervals in this paper hold the fitted pipeline fixed and resample evaluation rows or queries. Refitting the whole grouped pipeline twenty times under different group-to-fold assignments (\texttt{repeat\_fit\_stability.py}) moves the slice gain by a standard deviation of $0.002$--$0.003$ (SQuAD $0.175\pm0.003$, range $[0.171,0.182]$; Natural Questions $0.253\pm0.002$; DuoRC $0.057\pm0.003$, range $[0.052,0.065]$; WikiQA $0.176\pm0.003$; ASNQ $0.202\pm0.003$; TriviaQA $0.147\pm0.002$), every refit positive, and the full-population change by $0.001$--$0.002$. On the code audit the residual format effects range over $[-0.045,-0.029]$ and $[+0.031,+0.044]$ across twenty fold seeds. Fitting variability is an order of magnitude below the clustered interval widths, which remain the reported uncertainty; the two are conditional and full-procedure statements respectively.

\textbf{Why grouping.} In the QA settings a question contributes several candidate sentences. Row-level $k$-fold cross-fitting can place some of a question's candidates in a training fold and others in the corresponding held-out fold, so the surface predictor and the residualizer see, at fit time, other candidates of the very question they are later applied to, a leak that question--answer overlap features make material. Query-clustered bootstrap intervals do not address this: they correct the \emph{uncertainty} of estimates and leave the fitting-time leakage in place. The corrected protocol groups every fitted object by question (GroupKFold), and the main text reports grouped numbers throughout; Table~\ref{tab:grouped} shows what changed. Slice gains move by at most $0.05$ and no outcome changes; $R^2_{\phi\to s}$ drops throughout, which is the leaked component, and one decorrelation value crosses the gate: ASNQ fails ($0.139$) and its decorrelation claim is withdrawn (Section~\ref{sec:samelabel}).

\begin{table}[h]
\centering\small
\caption{Row-level vs.\ question-grouped cross-fitting on the five QA settings (\texttt{grouped\_full.py}; row-level values reproduce the paper's original records before trusting the deltas). Appendix robustness suites that predate the change remain row-level and are labeled as such where cited; the scorer replications were \emph{rescored} and re-fit under the grouped protocol (\texttt{rescore\_scorers.py}, \texttt{scorer\_grouped.py}), with row-level reproductions matching the original records first.}
\label{tab:grouped}
\begin{tabular}{lcccccc}
\toprule
 & \multicolumn{2}{c}{slice gain} & \multicolumn{2}{c}{$R^2_{\phi\to s}$} & \multicolumn{2}{c}{$R_{\mathrm{lin}}(\srep)$} \\
Setting & row & grouped & row & grouped & row & grouped \\
\midrule
WikiQA & $+0.177$ & $+0.181$ & 0.435 & 0.314 & 0.029 & 0.042 \\
ASNQ & $+0.247$ & $+0.200$ & 0.514 & 0.377 & 0.047 & \textcolor{failred}{\textbf{0.139}} \\
TriviaQA & $+0.152$ & $+0.148$ & 0.418 & 0.385 & 0.008 & 0.008 \\
SQuAD & $+0.177$ & $+0.176$ & 0.493 & 0.491 & 0.025 & 0.027 \\
DuoRC & $+0.061$ & $+0.056$ & 0.318 & 0.283 & 0.035 & 0.035 \\
\bottomrule
\end{tabular}
\end{table}

\textbf{Scorer replications, grouped.} Rescoring the replication scorers and re-fitting under the grouped protocol preserves both dose--responses: ASNQ STS-B $+0.231$ $[+0.209,+0.255]$ ($R^2=0.426$), ELECTRA $+0.162$ $[+0.130,+0.194]$ ($R^2=0.095$), QNLI $+0.007$ n.s.; WikiQA STS-B $+0.239$ $[+0.208,+0.272]$ ($R^2=0.360$), QNLI $+0.045$ $[+0.027,+0.062]$ ($R^2=0.016$); the WikiQA primary scorer reads $R^2=0.314$ grouped. These three grouped loadings are the values quoted in Section~\ref{sec:positives}. The TriviaQA replication, which the main text does not quote, also reproduces: STS-B $+0.162$ $[+0.146,+0.178]$ and the small negative QNLI effect ($-0.014$ $[-0.030,-0.001]$). On ASNQ, the residuals of all three channel-riding scorers stay above the decorrelation gate (STS-B $0.120$, ELECTRA $0.076$, MiniLM $0.139$): the withdrawal of Section~\ref{sec:positives} is dataset-level.

\textbf{Score representation.} Residualization is not invariant to monotone transformations of $s$, so the representation is part of the operator's specification (Section~\ref{sec:method}). Re-running every grouped QA setting on four representations of the same score, the scorer's logit (the paper's choice), its sigmoid probability, a rank-Gaussian transform, and a winsorized $z$-score, preserves the signs of both AUC changes in all twenty cells (slice gain positive, full-population change negative, Table~\ref{tab:scale}). Magnitudes are representation-dependent, the probability scale compresses gains (DuoRC $+0.056\to+0.022$), and the rank-Gaussian representation is strongest in all five settings, which makes it the recommended default when a score has no natural scale: it removes the arbitrariness the criterion would otherwise inherit. The decorrelation diagnostic travels with the representation, so we report it there too: under rank-Gaussian the raw loadings read $0.42$--$0.59$ and the adjusted scores fall to $R_{\mathrm{lin}}(\srep)\in[0.008,0.039]$ in four of the five settings; ASNQ again fails ($0.117$), so its withdrawal is representation-robust as well as $\alpha$-robust (Table~\ref{tab:alpha}).

\begin{table}[h]
\centering\small
\caption{Score-representation sensitivity under the grouped protocol (\texttt{scale\_sensitivity.py}): slice gain / full-population gain per representation.}
\label{tab:scale}
\begin{tabular}{lcccc}
\toprule
Setting & logit & probability & rank-Gaussian & winsorized $z$ \\
\midrule
WikiQA & $+0.181$ / $-0.054$ & $+0.101$ / $-0.091$ & $+0.216$ / $-0.056$ & $+0.179$ / $-0.058$ \\
ASNQ & $+0.200$ / $-0.082$ & $+0.153$ / $-0.169$ & $+0.245$ / $-0.094$ & $+0.200$ / $-0.082$ \\
TriviaQA & $+0.148$ / $-0.070$ & $+0.147$ / $-0.090$ & $+0.151$ / $-0.077$ & $+0.148$ / $-0.071$ \\
SQuAD & $+0.176$ / $-0.096$ & $+0.126$ / $-0.110$ & $+0.187$ / $-0.134$ & $+0.175$ / $-0.096$ \\
DuoRC & $+0.056$ / $-0.083$ & $+0.022$ / $-0.110$ & $+0.062$ / $-0.131$ & $+0.055$ / $-0.085$ \\
\bottomrule
\end{tabular}
\end{table}

\textbf{Query-level ranking.} Answer-sentence selection is a per-question ranking task, and pooled candidate-level AUC does not measure it. Table~\ref{tab:ranking} reports MRR and mean per-query AUC, raw vs.\ residualized, restricted to queries that touch the declared slice. Residualization \emph{lowers} both, in every setting: the slice gains are measurements on a pooled, label-defined subset, and they do not produce a better within-question ranker, the task-level form of the operational-scope limit that Section~\ref{sec:discussion} states.

\section{Gate and Operator Robustness}
\label{app:gaterobust}
\subsection{Slice-eligibility threshold sensitivity}
\label{app:threshold}

The slice-eligibility check's gate value $0.65$ is a heuristic. Table~\ref{tab:threshold} sweeps it over $\{0.55,\dots,0.75\}$ across the eleven eligibility-valued settings of Table~\ref{tab:scope} (eleven settings over the ten eligibility-valued rows, HellaSwag and SWAG sharing a row; HANS excluded: post-hoc, footnote $\P$). The counts are retrospective operating characteristics on the development sample: candidates refused at the slice-eligibility check were not scored, so their outcomes are unknown, and the table records gate agreement rather than predictive accuracy. The sweep is over the candidate pool, in which the slice-eligibility check still does the excluding; on the analysis rows alone it changes no outcome that the screen does not change already (Table~\ref{tab:ablation}). The sweep shows the \emph{joint} condition eligibility$\wedge$screen is insensitive over $[0.55,0.65]$, because every setting that fails the slice-eligibility check marginally also fails the screen; at $0.70$ the gate loses TriviaQA ($0.675$), and at $0.75$ only ASNQ ($0.751$) survives. The slice-eligibility check alone needs the gate inside the observed gap between the strongest rejected channel (SHP, $0.647$) and the weakest positive (TriviaQA, $0.675$); BoolQ ($0.640$) and HateCheck ($0.637$) fall below it. The prediction is carried by the measured gap; the constant merely marks a point inside it. Chronologically, both screen cutoffs were fixed on the NLI settings before any non-NLI candidate was screened; WikiQA, ASNQ, and TriviaQA were evaluated against the pre-set values unchanged, a de facto leave-domain-out test of the thresholds.

\begin{table}[h]
\centering\small
\caption{Decision stability as the router gate varies (11 router-valued settings; 5 true positives; grouped-protocol $A(\hat a)$ values for the QA settings). ``Pred.+'' = settings predicted screen-positive.}
\label{tab:threshold}
\begin{tabular}{lcccc|cccc}
\toprule
& \multicolumn{4}{c|}{Router $\wedge$ screen (the paper's condition)} & \multicolumn{4}{c}{Router alone} \\
Gate & Pred.+ & Prec. & Rec. & Agree. & Pred.+ & Prec. & Rec. & Agree. \\
\midrule
0.55 & 5 & 1.00 & 1.00 & 1.00 & 11 & 0.45 & 1.00 & 0.45 \\
0.60 & 5 & 1.00 & 1.00 & 1.00 & 7 & 0.71 & 1.00 & 0.82 \\
0.65 & 5 & 1.00 & 1.00 & 1.00 & 5 & 1.00 & 1.00 & 1.00 \\
0.70 & 4 & 1.00 & 0.80 & 0.91 & 4 & 1.00 & 0.80 & 0.91 \\
0.75 & 1 & 1.00 & 0.20 & 0.64 & 1 & 1.00 & 0.20 & 0.64 \\
\bottomrule
\end{tabular}
\end{table}

\subsection{Screen threshold sensitivity}
\label{app:c2threshold}

\textbf{The boundary record.} WikiQA scored by QNLI ($R^2_{\phi\to s}=0.016$) and this LLM judge ($0.031$) sit at the boundary of the $R^2$ gate alone. Both err in the same conservative direction, refusing a channel that turns out to be weakly adjustable, which suggests the gate's value may sit high for weak-but-real channels. We keep it at $0.05$ regardless, for the reason Section~\ref{sec:setup} gives for the operator: a constant lowered after seeing the two rows it misses is no longer pre-committed, so the misses are recorded instead. The sweep below covers both screen constants and reports how far each may move before any outcome changes. For the full eligibility$\wedge$screen conjunction the QNLI replications err in both directions. WikiQA (QNLI) fails a gate yet gains; ASNQ (QNLI) passes both gates yet does not gain ($+0.007$, interval covering zero); and TriviaQA (QNLI) passes all three gates ($R^2_{\phi\to s}=0.069$, $\Delta_{\mathrm{slice}}=0.046$, $A(\hat a)=0.676$) while the adjustment \emph{costs} alignment on the slice, $-0.014$ with a grouped interval $[-0.030,-0.001]$ that excludes zero. That last one is a screen false positive of the kind the post-adjustment checks exist to catch, and the procedure does catch it: the slice-gain test fails and the outcome is \emph{no gain established}. It is also protocol-sensitive, and we state both readings: under row-level folds the same pair gives $-0.012$ $[-0.025,+0.004]$, which does not separate.

\textbf{The four current-generation judge probes, in one view.} The WikiQA probe is the pointwise run of Section~\ref{sec:judgeprobe}; the three code-audit rows are the balanced-sample readings of Section~\ref{sec:entangled}.

\begin{center}\small
\begin{tabular}{llcc}
\toprule
Judge & Setting (design) & $R^2_{\phi\to s}$ & Screen \\
\midrule
Haiku 4.5 & WikiQA (answer selection, pointwise) & $0.031$ & below gate; screen refuses \\
Haiku 4.5 & MBPP code (paired, balanced sample) & $0.000$ & below gate$^{*}$ \\
Sonnet 5 & MBPP code (paired, balanced sample) & $0.000$ & below gate$^{*}$ \\
Opus 5 & MBPP code (paired, balanced sample) & $0.000$ & below gate$^{*}$ \\
\bottomrule
\end{tabular}
\end{center}

\noindent $^{*}$Reference readings: the observational screen itself does not apply on the designed branch (Section~\ref{sec:rmcode}).

The slice-eligibility check sweep above leaves the two screen constants fixed, and Section~\ref{sec:positives} concedes that the $R^2_{\phi\to s}$ gate may sit high for weak-but-real channels, so we sweep them as well. The outcomes are recomputed by re-thresholding the grouped statistics already reported, with no new fitting. Among the eleven eligibility-valued settings, five clear the slice-eligibility check and are the only ones whose outcome a screen constant can move: SNLI ($R^2_{\phi\to s}=0.100$, $\Delta_{\mathrm{slice}}=0.032$), SICK ($0.140$, $0.202$), TriviaQA ($0.385$, $0.233$), WikiQA ($0.314$, $0.242$), and ASNQ ($0.377$, $0.356$). Toxicity, the nearest refusal on both statistics ($0.040$, $0.025$), fails the slice-eligibility check first at $0.557$.

The outcome set is unchanged for $R^2_{\phi\to s}$ anywhere in $[0.03,0.10]$ and for $\Delta_{\mathrm{slice}}$ anywhere in $[0.02,0.03]$, so neither committed constant sits on a knife edge. The lower ends of those ranges are the sweep's grid edges, not binding values: on this population no outcome changes below the committed gates either, since every eligibility-passer clears both statistics with room to spare and every screen-refused setting fails the slice-eligibility check first, so the binding side is the upper one. Both margins are set by the same setting. SNLI is the binding case on each: it leaves at $R^2_{\phi\to s}>0.10$ and at any $\Delta_{\mathrm{slice}}$ gate above its own $0.032$, while the next-tightest, SICK, survives until $R^2_{\phi\to s}=0.14$. After SNLI the $\Delta_{\mathrm{slice}}$ values jump from $0.032$ to $0.202$, so that gate is insensitive across the whole range $[0.03,0.20]$ once SNLI is set aside. This is the quantitative version of a decision the main text already took on other grounds: SNLI is treated as illustrative only in Table~\ref{tab:scope}, and it is also the one positive whose outcome a small change in either screen constant would remove.

\subsection{Operator robustness}
\label{app:robustness}

All experiments in this appendix use the cached scores and the exact pipelines of the main results (SNLI $n{=}2400$, SICK $n{=}2400$); the quantity reported is the slice gain $A_{\mathrm{slice}}(\srep)-A_{\mathrm{slice}}(s)$ unless noted.

\textbf{Residualizer choice (Table~\ref{tab:resid}).} Every \emph{regularized} linear residualizer reproduces the positives; the conclusion does not depend on ridge specifically. Unregularized OLS fails on SNLI ($-0.140$): with $p{=}8000$ TF-IDF features and $n{=}2400$ (Table~\ref{tab:phispec}), the unregularized fit is ill-posed and the cross-fitted ``artifact prediction'' is mostly interpolated noise, whose subtraction destroys signal. The operative ingredient is regularization itself, whichever penalty supplies it, consistent with the operator being a shrinkage projection.

\begin{table}[h]
\centering\small
\caption{Slice gain under alternative residualizers (row-level suite).}
\label{tab:resid}
\begin{tabular}{lcccc}
\toprule
 & OLS & Ridge(1.0) & Lasso($10^{-3}$) & ElasticNet($10^{-3}$, 0.5) \\
\midrule
SNLI & $-0.140$ & $+0.044$ & $+0.022$ & $+0.032$ \\
SICK & $+0.022$ & $+0.199$ & $+0.128$ & $+0.184$ \\
\bottomrule
\end{tabular}
\end{table}

\textbf{Ridge strength (Table~\ref{tab:alpha}).} The paper fixes $\alpha=1$; Table~\ref{tab:alpha} sweeps $\alpha\in\{0.1,1,10,100\}$ over the five grouped QA settings and both NLI settings (\texttt{review\_r4\_experiments.py}). Slice gains sit near their maximum at $\alpha=1$; in three settings the gain alone reads slightly higher at $\alpha=10$ (TriviaQA $+0.159$ vs.\ $+0.148$; SNLI and SICK by $+0.001$) at the cost of a far worse decorrelation reading, so $\alpha=1$ is the gain--decorrelation compromise rather than a gain optimum. The extremes cost on both sides: underfitting at $\alpha=100$ removes too little of the channel (the residual criterion $R_{\mathrm{lin}}(\srep)$ climbs to $0.16$--$0.46$ as the under-removed component survives), while $\alpha=0.1$ begins to overfit the fold-wise channel estimate. And ASNQ's decorrelation failure is \emph{not} a shrinkage artifact: its $R_{\mathrm{lin}}(\srep)$ is minimized at the paper's $\alpha=1$ ($0.139$) and is worse at every other strength ($0.182/0.240/0.457$), and it persists under the rank-Gaussian representation as well ($0.117$; Appendix~\ref{app:grouped}). The withdrawal of ASNQ's decorrelation claim is a property of the setting; the operator's tuning does not produce it.

\textbf{Screening-statistic uncertainty (Table~\ref{tab:screenci}).} The screening statistics themselves carry sampling uncertainty, and two outcomes sit near their gates at the third decimal. Table~\ref{tab:screenci} attaches evaluation-resampling bootstrap CIs (query-clustered for the QA settings, example-level otherwise; predictions held fixed, the evaluation sample resampled, the same convention as every other CI in the paper) to $A(\hat a)$, $R^2_{\phi\to s}$, and $\Delta_{\mathrm{slice}}$ for the eight settings reproduced by \texttt{review\_r4\_experiments.py}. The intervals that matter are SNLI's, SICK's, and toxicity's. SNLI's $\Delta_{\mathrm{slice}}=0.032$ has CI $[0.019,0.045]$, crossing the $0.03$ heuristic cutoff; its screen pass is marginal in the interval sense, consistent with SNLI's role as the illustrative (not evidential) instance. SICK's $\Delta_{\mathrm{slice}}$ interval, $[0.002,0.429]$, crosses the same gate far more dramatically (its small slice makes the statistic imprecise), so SICK's screen pass is interval-marginal too, the accounting that already assigns SICK corroborating weight (Appendix~\ref{app:slicebias}). Applied uniformly, the interval-marginality convention (the analogue of the decorrelation $\dagger$) marks SNLI and SICK on the screen; under a uniformly strict interval reading across every gate, the fully clean positives are TriviaQA and SQuAD. Toxicity's two screen intervals also graze their gates from the failing side ($R^2$ upper bound $0.051$; $\Delta_{\mathrm{slice}}$ upper bound $0.040$), but its refusal does not rest on them: the router fails decisively (interval entirely below $0.65$), and the condition is conjunctive. The five QA settings clear all three gates with their entire intervals.

\begin{table}[h]
\centering\small
\caption{Ridge-strength sweep under each setting's main-text protocol (grouped for QA, row-level for NLI): slice gain / residual decorrelation $R_{\mathrm{lin}}(\srep)$ at each $\alpha$. Gates: gain CI excluding zero; $R_{\mathrm{lin}}\le0.05$.}
\label{tab:alpha}
\begin{tabular}{lcccc}
\toprule
Setting & $\alpha{=}0.1$ & $\alpha{=}1$ & $\alpha{=}10$ & $\alpha{=}100$ \\
\midrule
WikiQA & $+0.141$ / $0.045$ & $+0.181$ / $0.042$ & $+0.151$ / $0.145$ & $+0.055$ / $0.309$ \\
ASNQ & $+0.126$ / $0.182$ & $+0.200$ / $0.139$ & $+0.170$ / $0.240$ & $+0.049$ / $0.457$ \\
TriviaQA & $+0.124$ / $0.013$ & $+0.148$ / $0.008$ & $+0.159$ / $0.039$ & $+0.131$ / $0.157$ \\
SQuAD & $+0.166$ / $0.001$ & $+0.176$ / $0.027$ & $+0.164$ / $0.187$ & $+0.078$ / $0.454$ \\
DuoRC & $+0.041$ / $0.001$ & $+0.056$ / $0.035$ & $+0.053$ / $0.191$ & $+0.017$ / $0.374$ \\
SNLI & $+0.026$ / $0.101$ & $+0.044$ / $0.015$ & $+0.045$ / $0.169$ & $+0.033$ / $0.276$ \\
SICK & $+0.169$ / $0.094$ & $+0.199$ / $0.026$ & $+0.200$ / $0.198$ & $+0.164$ / $0.398$ \\
\bottomrule
\end{tabular}
\end{table}

\begin{table}[h]
\centering\small
\caption{Screening statistics with evaluation-resampling 95\% bootstrap CIs (\texttt{review\_r4\_experiments.py}; grouped protocol for QA, row-level otherwise). Gates: $A(\hat a)\gtrsim0.65$, $R^2_{\phi\to s}\gtrsim0.05$, $\Delta_{\mathrm{slice}}\gtrsim0.03$. $^{*}$post-hoc additions (the LLM-judge probe and the NQ dual-evaluation setting, Section~\ref{sec:positives}), outside the evidence accounting. DuoRC's $A(\hat a)$ reads $0.835$ here (grouped, resampled) vs.\ $0.834$ at screen commitment (KFold); the protocols differ in the third decimal.}
\label{tab:screenci}
\begin{tabular}{lccc}
\toprule
Setting & $A(\hat a)$ & $R^2_{\phi\to s}$ & $\Delta_{\mathrm{slice}}$ \\
\midrule
WikiQA & $0.748$ {\scriptsize$[0.721,0.771]$} & $0.314$ {\scriptsize$[0.286,0.340]$} & $0.242$ {\scriptsize$[0.205,0.279]$} \\
ASNQ & $0.751$ {\scriptsize$[0.732,0.769]$} & $0.377$ {\scriptsize$[0.353,0.400]$} & $0.356$ {\scriptsize$[0.319,0.390]$} \\
TriviaQA & $0.676$ {\scriptsize$[0.664,0.686]$} & $0.385$ {\scriptsize$[0.368,0.401]$} & $0.233$ {\scriptsize$[0.218,0.249]$} \\
SQuAD & $0.813$ {\scriptsize$[0.800,0.824]$} & $0.491$ {\scriptsize$[0.474,0.508]$} & $0.226$ {\scriptsize$[0.202,0.251]$} \\
DuoRC & $0.835$ {\scriptsize$[0.815,0.853]$} & $0.283$ {\scriptsize$[0.250,0.316]$} & $0.144$ {\scriptsize$[0.117,0.173]$} \\
SNLI & $0.708$ {\scriptsize$[0.688,0.727]$} & $0.100$ {\scriptsize$[0.074,0.127]$} & $0.032$ {\scriptsize$[0.019,0.045]$} \\
SICK & $0.727$ {\scriptsize$[0.702,0.750]$} & $0.140$ {\scriptsize$[0.107,0.172]$} & $0.202$ {\scriptsize$[0.002,0.429]$} \\
Toxicity & $0.557$ {\scriptsize$[0.539,0.574]$} & $0.040$ {\scriptsize$[0.028,0.051]$} & $0.025$ {\scriptsize$[0.012,0.040]$} \\
WikiQA (LLM judge)$^{*}$ & $0.748$ {\scriptsize$[0.721,0.771]$} & $0.031$ {\scriptsize$[0.003,0.059]$} & $0.064$ {\scriptsize$[0.041,0.088]$} \\
NQ dual-evaluation$^{*}$ & $0.677$ {\scriptsize$[0.658,0.695]$} & $0.442$ {\scriptsize$[0.418,0.465]$} & $0.381$ {\scriptsize$[0.349,0.413]$} \\
\bottomrule
\end{tabular}
\end{table}

\textbf{Fold count and feature noise.} With $k\in\{3,5,10\}$ folds the SNLI/SICK gains are $+0.041/{+}0.044/{+}0.041$ and $+0.176/{+}0.199/{+}0.205$ ($k$ immaterial); Gaussian noise at $10/20/30\%$ of each feature's sd degrades the gains gracefully ($+0.043/{+}0.041/{+}0.038$ and $+0.202/{+}0.197/{+}0.196$ vs.\ clean $+0.044$/$+0.199$).

\textbf{Artifact predictor.} A linear SVM in place of the logistic probe leaves $A(\hat a)$ essentially unchanged (SNLI $0.708\to0.694$; SICK $0.727\to0.713$) and every outcome identical; gain \emph{magnitudes} shift because the slice is defined by the predictor's decisions (SNLI $+0.037$; SICK $+0.039$ on a larger, higher-scoring slice), but both predictors give strictly positive, CI-excluding-zero gains; sign and outcome are threshold-robust.

\textbf{Artifact feature ablation (HANS).} In the HANS pipeline, whose $\phi$ mixes dense surface features (overlap, lengths) with hypothesis TF-IDF, ablating one block at a time localizes the lexical-overlap adjustment: the dense overlap$+$length block carries the entire $+0.104$ effect, while hypothesis TF-IDF alone is inert (global gain $+0.000$). The channel that responds there is the overlap geometry rather than word identity, consistent with lexical overlap being the one HANS heuristic with a dense linear surface form.

\section{Scope of the Decorrelation Diagnostic: What Is and Is Not Measured}
\label{app:certprobes}

Nonlinear probes of the adjusted score still recover the surface decision at AUC $0.83$--$0.94$ in the NLI settings, while the QA settings sit near the probe's permutation floor. Across the seven positives, the five slice positives of Table~\ref{tab:scope} and the two held-out replications, this criterion is the component that most often qualifies or withdraws a claim: three pass cleanly, three are marginal under a strict interval reading, and one (ASNQ) does not meet it, and its slice gain is reported with the criterion recorded as unmet.

The decorrelation diagnostic of Section~\ref{sec:method} is a claim about the \emph{linear recoverability of the continuous surface index}, and this appendix delimits it empirically in both directions. The values here are from the original row-level fits; Appendix~\ref{app:grouped} reports the question-grouped shift, under which ASNQ's index correlation fails the gate.

\textbf{The channel-level criterion (the measured direction).} Table~\ref{tab:certcorr} reports $|\mathrm{corr}(u,\hat c)|$ (the linear channel index, the criterion's object) and $|\mathrm{corr}(u,\hat g)|$ before and after adjustment on the three original positives. The index correlation collapses ($0.30$--$0.50\to0.015$--$0.029$), meeting the closure criterion ($\le0.05$) everywhere on the point estimate (intervals, and which settings are marginal under the stricter reading, in Section~\ref{sec:method}); the probability-transform diagnostic gives materially identical values ($0.30$--$0.46\to0.014$--$0.031$), and TriviaQA closes the same way ($0.455\to0.008$; \texttt{review\_r3\_experiments.py}); ASNQ closed at $0.602\to0.047$ under these row-level fits but \emph{fails} the gate under the question-grouped protocol ($0.139$; Section~\ref{sec:positives}, Appendix~\ref{app:grouped}). Residual correlation with the removed component $\hat g$ does not vanish (SICK $-0.14$): the cross-fitted ridge is an approximation of the population projection, as stated in Section~\ref{sec:method}, and $\hat g$ also carries construct-correlated surface variance, so $\mathrm{corr}(\srep,\hat g)=0$ is neither guaranteed nor required by the criterion, whose object is $\hat c$.

\begin{table}[h]
\centering\small
\caption{Channel-level decorrelation: linear recoverability of the channel index $\hat c$ (and the removed component $\hat g$) from the score, before and after adjustment; the probability transform $\hat a=\sigma(\hat c)$ is a secondary diagnostic.}
\label{tab:certcorr}
\begin{tabular}{lcccc}
\toprule
 & $|\mathrm{corr}(s,\hat c)|$ & $|\mathrm{corr}(\srep,\hat c)|$ & $|\mathrm{corr}(s,\hat g)|$ & $|\mathrm{corr}(\srep,\hat g)|$ \\
\midrule
SNLI & 0.297 & \textbf{0.015} & 0.324 & 0.073 \\
SICK & 0.444 & \textbf{0.026} & 0.396 & 0.140 \\
WikiQA & 0.496 & \textbf{0.029} & 0.659 & 0.001 \\
\bottomrule
\end{tabular}
\end{table}

\textbf{The decision-level diagnostic (not guaranteed, and not always closed).} Probing the thresholded artifact \emph{decision} $z=\mathbf{1}[\hat a>\tau]$ (the flag at the base-rate threshold $\tau$) from the score is strictly harder to close: $z$ is a nonlinear (thresholded) function of $\hat a$, so linear orthogonality to $\hat a$ does not bound a probe's AUC on $z$. Measured values after adjustment: SNLI $0.451$, SICK $0.648$, WikiQA $0.559$ (from $0.627/0.878/0.843$ raw). SICK's $0.648$ illustrates that the decision-level floor is a diagnostic and carries no closure claim, so the paper's closure claim is defined on the channel, with the decision-level numbers reported alongside. Nor is the residual an argument for switching to the exact projection: in the $p\gg n$ regime of these settings the unregularized OLS projection \emph{does} close even the decision floor (measured: $0.489/0.481/0.494$ on SNLI/SICK/WikiQA, all bootstrap CIs containing $0.5$), but it does so by interpolation; the score it returns has construct alignment $0.534/0.664$ on SNLI/SICK, down from $0.958/0.967$ raw: the channel closes only because the score itself is destroyed. This is a decorrelation--validity trade-off, and the shrinkage operator sits deliberately on its useful side: an approximate, channel-level criterion with the construct intact.

\textbf{Nonlinear probes (the excluded direction).} We probe the residualized score with cross-fitted logistic (linear), MLP ($2{\times}32$), and random-forest classifiers predicting the artifact decision from the score alone (Table~\ref{tab:probes}). Linear recovery falls sharply after adjustment (SNLI $0.627\to0.451$, below chance; SICK $0.878\to0.648$). Nonlinear recovery, in the NLI settings measured here, does not: it rises on SNLI (RF $0.581\to0.827$, MLP similarly) and stays high on SICK (RF $0.929\to0.895$). Subtracting $\hat g(\phi)$ from $s$ imprints the artifact estimate into the residualized score's nonlinear structure (the map $s\mapsto\srep$ is artifact-dependent), so a probe that can invert non-monotone relationships reads it back. Erasure by residualization relocates the linearly recoverable part of the surface decision rather than destroying it (Section~\ref{sec:synthetic}). A \emph{nonlinear} closure statement therefore requires a different primitive (e.g., quantile-matching or optimal-transport erasure), which Appendix~\ref{app:quantile} instantiates; what the linear criterion establishes (exactly for the ideal OLS reference, and as the measured criterion of Table~\ref{tab:certcorr} for the deployed operator) is that linear consumers of the residualized score recover essentially nothing of the \emph{prespecified one-dimensional index} $\hat c$; other directions of $\phi$, and any nonlinear consumer, are outside this statement.

\textbf{Why subtraction can \emph{raise} nonlinear recoverability.} That SNLI result, removing the linear channel while a flexible probe reads the artifact decision better than before ($0.581\to0.827$), reproduces from the released cache ($0.582\to0.828$ here). We hypothesize that score
saturation explains the relocation; two diagnostics support that interpretation. First, the
subtraction does what it claims on the \emph{monotone} part: the mean of the score differs by $+0.250$ between the two artifact-decision groups before adjustment and by $-0.007$ after, and the group standard deviations barely move ($1.07\to1.09$ as a ratio). The recovery is not explained by these mean and spread summaries. Second, the audited score is close to saturated ($53.7\%$ of its values fall below $0.05$ and $39.8\%$ above $0.95$). Subtracting a smooth predictor $\hat g(\phi)$ from a near-binary $s$ makes the residual a \emph{many-to-one} function of the pair $(s,\hat g)$. A residual near $-0.7$ arises when $s\approx0$ while $\hat g\approx0.7$; a residual near $+0.3$ arises when $s\approx1$ while $\hat g\approx0.7$. Both are cases where the surface predictor was confident, so both carry $z{=}1$, while the intermediate residuals are dominated by cases where it was not. And the pathology is absent in the QA settings, whose cross-encoder logits are not saturated and where the linear operator already leaves the probe near its floor (Table~\ref{tab:quantile}). Saturation is the mechanism consistent with both measurements. Isolating it directly, by excluding the saturated region and re-probing, is infeasible here: only $6.5\%$ of the sample lies outside that region.

The resulting relationship is non-monotone, which is precisely the shape a linear reader cannot see and a random forest can: the monotone reading misses it (linear AUC $0.44$, $\approx0.56$ with the sign reversed) while the interval-based probe reads it at $0.83$ (Table~\ref{tab:relocation}).

\begin{table}[h]
\centering\small
\caption{Artifact-decision rate by residual band on SNLI after linear adjustment (row-level suite).}
\label{tab:relocation}
\begin{tabular}{lrr}
\toprule
Residual band $\srep$ & $n$ & $P(z{=}1)$ \\
\midrule
$[-1.10,-0.50)$ & 319 & 0.931 \\
$[-0.50,-0.15)$ & 878 & 0.260 \\
$[-0.15,+0.15)$ & 196 & 0.265 \\
$[+0.15,+0.50)$ & 532 & 0.883 \\
$[+0.50,+1.10)$ & 475 & 0.345 \\
\bottomrule
\end{tabular}
\end{table}

This also says why the conditional-quantile eraser below removes the pathology while the linear operator cannot. The linear operator targets the conditional \emph{mean} and succeeds there; the information that survives lives in which interval of the residual an example lands in. An operator that equalizes the whole conditional distribution within strata of $\hat c$ has no such remainder to leave behind, at the price of returning a within-stratum rank in place of a score on the original scale.

\begin{table}[h]
\centering\small
\caption{Probe AUC recovering the artifact decision from the score alone (5-fold cross-fitted, row-level suite).}
\label{tab:probes}
\begin{tabular}{lcccccc}
\toprule
 & \multicolumn{3}{c}{raw $s$} & \multicolumn{3}{c}{adjusted $\srep$} \\
 & linear & MLP & RF & linear & MLP & RF \\
\midrule
SNLI & 0.627 & 0.627 & 0.581 & \textbf{0.451} & 0.881 & 0.827 \\
SICK & 0.878 & 0.874 & 0.929 & \textbf{0.648} & 0.939 & 0.895 \\
\bottomrule
\end{tabular}
\end{table}

\section{A Nonlinear-Closure Instantiation: the Cross-Fitted Conditional-Quantile Eraser}
\label{app:quantile}

The linear operator's decorrelation diagnostic covers linear consumers only; Appendix~\ref{app:certprobes} shows that in the NLI settings nonlinear probes still recover the artifact decision from the linearly residualized score (on SNLI the subtraction \emph{relocates} information, raising RF recoverability from $0.581$ to $0.827$; in the QA settings linear adjustment already leaves the RF probe near its floor, Table~\ref{tab:quantile}, so the stronger primitive is needed where relocation occurs). Here we instantiate the stronger primitive the main text points to. \textbf{Construction.} Bin the channel index $\hat c$ into $\approx20$ equal-count strata whose boundaries are aligned to the decision threshold (so $z$ is constant within every stratum), and replace $s$ by its randomized within-stratum rank. Everything is \emph{cross-fitted}: stratum edges and per-stratum CDFs are estimated on the training folds and applied to the held-out fold, mirroring the ridge operator. In the population the output $s_q$ is uniform within every stratum, hence carries no information about the stratum (or any function of it, including $z$) to \emph{any} consumer, linear or not; cross-fitting is what keeps the finite-sample version close to that population guarantee, as measured by the probes here (the in-sample variant leaks: measured recoverability $0.56$--$0.65$).

\textbf{Measured closure and its price.} By the same RF probe, calibrated against a permutation null (the probe run on a shuffled copy of $s_q$, giving the estimator's noise floor $\approx0.50$--$0.51$):

\begin{table}[h]
\centering\small
\caption{Cross-fitted conditional-quantile eraser: RF-probe recoverability of the artifact decision $z$ from the score alone (symmetrized AUC; null = permutation floor), and construct alignment on the declared slice and full population, versus the linear operator. Row-level cross-fitting, as elsewhere in this appendix; grouped headline values differ at the third decimal, except ASNQ, where the row-level fit departs further from the grouped headline (Table~\ref{tab:threepop}). Values are recomputed inside \texttt{quantile\_eraser.py}'s own run, so the linear columns can differ from other row-level tables in the second decimal.}
\label{tab:quantile}
\begin{tabular}{lcccc|cc|cc}
\toprule
 & \multicolumn{4}{c|}{RF recoverability of $z$} & \multicolumn{2}{c|}{$A$ slice} & \multicolumn{2}{c}{$A$ full} \\
Setting & raw & linear & quantile & null & linear & quantile & linear & quantile \\
\midrule
SNLI & 0.581 & 0.827 & \textbf{0.506} & 0.502 & 0.970 & 0.966 & 0.880 & 0.886 \\
SICK & 0.929 & 0.895 & \textbf{0.503} & 0.510 & 0.964 & 0.998 & 0.906 & 0.822 \\
WikiQA & 0.672 & 0.513 & \textbf{0.515} & 0.508 & 0.801 & 0.864 & 0.809 & 0.818 \\
ASNQ & 0.750 & 0.595 & \textbf{0.505} & 0.502 & 0.735 & 0.826 & 0.770 & 0.777 \\
TriviaQA & 0.637 & 0.539 & \textbf{0.504} & 0.512 & 0.626 & 0.687 & 0.636 & 0.649 \\
\bottomrule
\end{tabular}
\end{table}

The eraser reaches the null floor on every positive: the tested RF probe reads AUCs within a few thousandths of its permutation baseline. Under the population conditional CDF the randomized transform is independent of the stratum and hence of $z$; the cross-fitted implementation inherits that property only approximately, and the probe result does not establish information removal against all nonlinear consumers. This is nonetheless the closure direction the linear operator cannot offer, though in the QA settings the linear operator's measured recoverability was already near the floor, so the eraser's practical contribution is confined to the NLI settings, where the tested probe's recoverability on SNLI, the setting in which the linear operator relocated the artifact, falls back to its permutation baseline. The price is not paid where one might expect: same-label slice alignment matches (within $0.004$) or exceeds the linear operator on all five settings (the within-stratum ordering that $s_q$ preserves is the construct-bearing residual variation), and the full-population cost is comparable except on SICK, where destroying the score's scale costs more. And the object changes: $s_q$ is a within-stratum rank with no calibrated scale, so any consumer needing score \emph{magnitudes} is outside its scope: the same slice-scoped, evaluative scope as the linear operator, now with a stronger closure statement (\texttt{quantile\_eraser.py}). The same-label caveat of Appendix~\ref{app:slicebias} applies to the slice columns unchanged.

\section{Slice Definition: Selection-Bias Checks}
\label{app:slicebias}

\textbf{The placebo control and the sub-block refits, in full.} Does conditioning on any predictor's errors favor subtraction by itself? A placebo control measures that directly. We build predictors matched to the surface predictor's AUC but erring independently of $\phi$, apply the declared slice rule to them unchanged, and find the resulting slices gain nothing: because a placebo slice is drawn independently of $\phi$, the adjustment's effect on it is just its effect on a random subset, so each lands on its setting's full-population cost, against declared-slice gains of $+0.055$ to $+0.179$, an excess of $0.141$ to $0.275$ over the placebo (Table~\ref{tab:placebo}). The tested artifact-independent construction therefore contributes no gain at matched AUC, and what remains to be bounded is the narrower bias Property 4 describes, conditioning on the errors of the \emph{same} predictor whose component is subtracted. The derivation of that bias uses the slice being the error set of the model whose component is subtracted, so we refit the slice predictor on a proper sub-block of $\phi$ and leave the subtracted component exactly as the pipeline produces it. Three blocks give three slices in each of the four QA settings (WikiQA, ASNQ, TriviaQA and the held-out SQuAD): answer TF-IDF, the length triple, and question--answer overlap. $A(\hat a)$ and $\Delta_{\mathrm{slice}}$ are recomputed on each new slice, so a pair clears the gates on its own reading; the loading $R^2_{\phi\to s}$ is a property of the operator and is unchanged. The sub-block sits inside $\phi$, so the subtracted component is still fit on those features; the refit removes the identity of the two models, so the bias is only attenuated. Seven of the twelve resulting (setting, slice) pairs clear the slice-eligibility check and the screen. Six gain and one loses, ASNQ on its length-defined slice at $-0.019$ $[-0.037,-0.002]$ (Appendix~\ref{app:crossfamily}). The loss is recorded under falsification condition (i) at the low severity a same-label evaluation carries (Appendix~\ref{app:falsification}). SQuAD carries the sharpest reading: on its overlap-defined slice the gain is $+0.120$ $[+0.102,+0.140]$ against the annotator label and $+0.103$ $[+0.087,+0.121]$ against the disjoint annotator set of Section~\ref{sec:heldout}, so the labels the gain is evaluated against are not the instances that defined the slice. That is the whole of the independence Section~\ref{sec:heldout} establishes: both label sets come from one annotation task. The gains are smaller than on the declared slices (WikiQA reads $+0.067$ on its TF-IDF slice and $+0.140$ on its overlap slice against $+0.181$ declared), so part of the declared magnitude may still be conditioning; the control shows the gains persist when the slice definition and the subtraction no longer share their full feature channel, and it does not quantify the bias that remains.

\begin{table}[!ht]
\centering\small
\caption{The placebo-slice control. A placebo predictor is matched to the surface predictor's AUC but errs independently of $\phi$; the declared slice rule is then applied to it unchanged. The excess column is the declared gain minus the setting's full-population cost. All four columns come from one run of \texttt{placebo\_slice\_test.py} (fixed seed, reproduced across two independent runs), so the declared-slice values differ in the third decimal from the main pipeline's $+0.176$, $+0.181$ and $+0.056$ and the comparison is meaningful only within a run. The placebo band is the $2.5$--$97.5$ percentile over $B{=}200$ placebo draws, where the paper's other intervals are $1{,}000$-resample trace-level bootstraps, and the run records no interval on the declared column (Appendix~\ref{app:slicebias}).}
\label{tab:placebo}
\begin{tabular}{lcccc}
\toprule
Setting & Declared slice & Placebo slice & Full population & Excess \\
\midrule
SQuAD  & $+0.177$ & $-0.098$ {\scriptsize$[-0.114,-0.085]$} & $-0.098$ & $0.275$ \\
WikiQA & $+0.179$ & $-0.052$ {\scriptsize$[-0.067,-0.039]$} & $-0.052$ & $0.232$ \\
DuoRC  & $+0.055$ & $-0.086$ {\scriptsize$[-0.101,-0.070]$} & $-0.085$ & $0.141$ \\
\bottomrule
\end{tabular}
\end{table}

\textbf{Dual-measurement check: breaking the double use of $y$.} One label both defines the slice and evaluates the effect. A second, provenance-separated construct evaluation $y'$ measures how much that double use inflates the gain. \emph{Synthetic} (two independent noisy readings; slice and screen from $y_1$ only): against the slice-defining label the gain is inflated and nearly flat in $\rho$ ($+0.47$ to $+0.54$); against $y_2$ it is smaller and ordered as theory requires ($+0.24/{+}0.20/{+}0.15/{+}0.06$ as $\rho$ rises $0.2\to0.8$). Same-label gains are therefore biased upward in this model. \emph{SNLI} (five annotator labels; pipeline from annotators $\{1,2,3\}$, evaluation from disjoint $\{4,5\}$; agreement $0.993$, $n{=}2{,}273$): the gain survives, $+0.026$ against the disjoint labels vs.\ $+0.034$ same-label (\texttt{dual\_measurement\_check.py}). \emph{SQuAD} (slice and pipeline from the first annotator span, evaluation from the disjoint remaining spans): $+0.155$, CI $[+0.135,+0.176]$, against $+0.176$ same-label (\texttt{squad\_dual\_eval.py}); on the one setting with a second label set at headline scale, the effect is not attributable solely to reuse of the slice-defining labels, though the conditioning structure itself remains (Section~\ref{sec:heldout}).

The headline positives are measured on slices defined by a surface-only predictor, and the adjustment residualizes on features of the same channel, inviting the question of whether the slice conditioning makes the evaluation favorable. The checks below assess several sources of the observed gains; they do not quantify the remaining selection bias.

\textbf{(1) Slice membership is always out-of-fold.} In every main result, the artifact predictor's decisions come from cross-fitted (out-of-fold) predictions; no example's slice membership is decided by a model that saw it.

\textbf{(2) Independently defined slices (nested definition, full-data operator).} We split each dataset in half, train the slice-defining artifact predictor on one half only, and evaluate the paper's operator on the other half's slice (both directions). The gain is positive in all six half-evaluations: SNLI $+0.012/{+}0.036$, SICK $+0.204/{+}0.122$, WikiQA $+0.156/{+}0.206$. A complementary variant that keeps the full sample but breaks fold-sharing (slice definition and adjustment on \emph{independent} fold assignments) agrees: SNLI $+0.031\pm0.003$, SICK $+0.208\pm0.015$. Selection bias from shared training data between slice definition and evaluation is not driving the positives. The same runs also yield a worst-group check: adjusted SICK improves its worst group ($0.753\to0.868$) while adjusted SNLI trades its worst group down ($0.928\to0.844$), consistent with the adjustment being a subgroup-targeted measurement correction, which is how Section~\ref{sec:discussion} scopes deployment.

\textbf{(3) The fully-nested protocol, and the caveat that demotes SNLI.} Making the \emph{adjustment fit} nested as well (operator cross-fitted within the evaluation half, $n/2$) leaves SICK ($+0.172/{+}0.077$) and WikiQA ($+0.164/{+}0.203$) intact but eliminates the SNLI margin ($-0.019/{-}0.003$). Check (2) rules out slice selection as the cause. A dimensionality-reduction control then rules out the simplest remaining explanation: projecting $\phi$ to $k\in\{64,128,256\}$ SVD components (fit within the evaluation half, no leakage) keeps the pooled SNLI gain positive ($+0.027$ to $+0.037$) but does \emph{not} restore the nested margin (gains $-0.019$ to $+0.022$ across $k$ and halves), so the failure is not narrowly a $p\!\gg\!n$ ridge artifact. The SNLI gain does not persist under the fully nested half-sample protocol: it does not persist at any dimensionality we tested once the data available to the operator is halved. This is why Section~\ref{sec:positives} treats SNLI as illustrative and rests the same-label empirical claim primarily on the QA settings (WikiQA, ASNQ, TriviaQA, and held-out SQuAD/DuoRC), whose intervals are narrow. SICK passes every protocol here, but its slice is small ($n{=}455$) and its intervals correspondingly wide (slice gain CI $[+0.018,+0.450]$; $\Delta_{\mathrm{slice}}$ CI $[0.002,0.429]$, Table~\ref{tab:screenci}), so it corroborates the pattern without carrying it.

\textbf{(4) Slices and scorers not defined by the artifact predictor.} HANS's per-heuristic slices are \emph{dataset-defined} (fixed by the benchmark's construction, independent of any predictor we train), and the adjustment still splits them as the condition predicts, including the $+0.104$ positive on lexical overlap. And re-scoring WikiQA with two architecturally and distributionally different scorers replicates the positive on the same slice recipe: QNLI DistilRoBERTa $+0.028$, CI $[+0.008,+0.047]$ ($R^2_{\phi\to s}=0.082$, $\Delta_{\mathrm{slice}}=0.153$) and STS-B RoBERTa $+0.238$, CI $[+0.204,+0.271]$ ($R^2_{\phi\to s}=0.463$, $\Delta_{\mathrm{slice}}=0.324$). Across the three scorers the gain rises monotonically with how heavily each rides the channel (as originally recorded, row-level: $R^2_{\phi\to s}$ $0.082/0.435/0.463\to$ gain $+0.028/{+}0.177/{+}0.238$; the grouped values quoted in the body, $+0.045/{+}0.181/{+}0.239$, preserve the ordering), the dose--response of Figure~\ref{fig:repairability}, measured with dataset, features, and slice recipe held fixed.

\textbf{(5) Placebo slices: the tested generic construction contributes no gain.} If conditioning on a predictor's errors favored subtraction \emph{per se}, slices defined by any label-informed predictor of the same strength should show it. We built placebo predictors $u=\sqrt{2}\,\Phi^{-1}(A(\hat a))\,y+\varepsilon$ with $\varepsilon\sim\mathcal{N}(0,1)$, matched to the surface predictor's AUC but with errors unrelated to $\phi$, and applied the rate-calibrated slice rule to $u$ verbatim ($B{=}200$ draws, fixed seed, deterministic across two independent runs; \texttt{placebo\_slice\_test.py}). On all three QA settings the placebo-slice gains concentrate at the full-population cost, nowhere near the declared-slice gain: SQuAD declared $+0.177$ vs.\ placebo $-0.098$ $[-0.114,-0.085]$ (full-population $-0.098$); WikiQA $+0.179$ vs.\ $-0.052$ $[-0.067,-0.039]$ (full $-0.052$); DuoRC $+0.055$ vs.\ $-0.086$ $[-0.101,-0.070]$ (full $-0.085$). The tested error-conditioning construction contributes no gain at matched AUC; this does not identify the contribution of every conceivable selection mechanism. What Property 4's favoring requires is conditioning on the errors of the \emph{same} predictor whose component is subtracted; that specific bias is what the dual-measurement and nested checks above probe, without quantifying what remains.

\section{Slices Drawn With Other Features}
\label{app:crossfamily}

\textbf{The control, and what it does not do.} Property 4's derivation takes the slice to be the error set of the model whose component is subtracted. Refitting the slice predictor on a proper sub-block of $\phi$ changes the feature block the slice is drawn from while leaving the operator untouched: the subtracted component is still the pipeline's full-$\phi$ cross-fitted fit, and only the slice moves. The refit does not separate the features. The sub-block is inside $\phi$, so the subtracted component is fit on the very features the slice was drawn with, and any route from feature overlap to a favored subtraction survives. The control therefore weakens the shared-channel dependence rather than removing it. The surviving gains show the effect persists under these refits; they do not quantify the remaining selection bias. In the QA settings the available blocks are the answer TF-IDF matrix, the length triple, and question--answer overlap (Table~\ref{tab:phispec}). Everything else follows the recipe of Section~\ref{sec:setup}, and $A(\hat a)$ is read off the refitted predictor. Table~\ref{tab:crossfamily} reports all twelve pairs. Seven clear both gates; six of those gain with an interval above zero and one loses (ASNQ on the length-defined slice, $-0.019$ $[-0.037,-0.002]$). The five that fail do so at the slice-eligibility check, where the sub-block predicts the label too weakly for the slice to be informative, which is the condition the slice-eligibility check is designed to refuse on.

\textbf{The edge of the bound.} Cutting the adjustment block down as well, so that the subtracted component is fit only on features the slice predictor never saw, removes the gain: over the same twelve pairs none of the seven that clear both gates gains, four lose and three are null (\texttt{feature\_split\_slice.py}). Appendix~\ref{app:identifiability} predicts this. A component fit on features uncorrelated with the channel the slice was drawn from recovers little of that channel, and as recovery falls the operator degenerates to subtracting noise, which costs through the variance it removes. Together the two runs locate the dependence: in these feature-split runs the gains disappear when the residualizer is restricted to the complementary block, and they persist under slices drawn with other features that carry the channel. (Scripts: \texttt{cross\_family\_slice.py}, \texttt{feature\_split\_slice.py}.)

\begin{table}[h]
\centering
\small
\caption{Slice gains when the slice predictor is refit on a sub-block of $\phi$ and the subtracted component is left at the pipeline's full-$\phi$ fit. $A(\hat a)$ (the router statistic) is the refitted predictor's AUC against the label, gate $\approx0.65$; $\Delta_{\mathrm{slice}}$ is measured on the new slice, gate $\approx0.03$; $R^2_{\phi\to s}$ is unchanged from the main pipeline and passes throughout. Intervals are query-clustered over $1{,}000$ resamples. The declared-slice rows are re-derived inside this run for comparability, so they differ from Table~\ref{tab:scope}'s headline values in the third decimal (ASNQ reads $+0.198$ against $+0.200$); SQuAD's declared row, which Table~\ref{tab:scope} does not carry, comes from the held-out run. $^{\ddagger}$SQuAD's overlap slice against the disjoint annotator set: $+0.103$ $[+0.087,+0.121]$.}
\label{tab:crossfamily}
\begin{tabular}{llccll}
\toprule
Setting & Slice from & $A(\hat a)$ & $\Delta_{\mathrm{slice}}$ & Gain (CI) & Gates \\
\midrule
WikiQA & declared (dense) & $0.747$ & $0.241$ & $+0.181$ $[+0.148,+0.209]$ & pass \\
WikiQA & answer TF-IDF & $0.654$ & $0.092$ & $+0.067$ $[+0.047,+0.089]$ & pass \\
WikiQA & lengths & $0.642$ & $0.057$ & $+0.008$ $[-0.013,+0.029]$ & router fails \\
WikiQA & overlap & $0.659$ & $0.169$ & $+0.140$ $[+0.111,+0.167]$ & pass \\
\midrule
ASNQ & declared (dense) & $0.751$ & $0.360$ & $+0.198$ $[+0.179,+0.219]$ & pass \\
ASNQ & answer TF-IDF & $0.813$ & $0.048$ & $+0.019$ $[+0.0001,+0.035]$ & pass \\
ASNQ & lengths & $0.719$ & $0.178$ & $-0.019$ $[-0.037,-0.002]$ & pass \\
ASNQ & overlap & $0.720$ & $0.341$ & $+0.183$ $[+0.164,+0.203]$ & pass \\
\midrule
TriviaQA & declared (dense) & $0.676$ & $0.232$ & $+0.146$ $[+0.131,+0.162]$ & pass \\
TriviaQA & answer TF-IDF & $0.634$ & $0.116$ & $+0.069$ $[+0.057,+0.082]$ & router fails \\
TriviaQA & lengths & $0.657$ & $0.101$ & $+0.044$ $[+0.033,+0.057]$ & pass \\
TriviaQA & overlap & $0.610$ & $0.335$ & $+0.175$ $[+0.161,+0.191]$ & router fails \\
\midrule
SQuAD & declared (dense) & $0.813$ & $0.227$ & $+0.177$ $[+0.159,+0.200]$ & pass \\
SQuAD & answer TF-IDF & $0.621$ & $0.032$ & $-0.010$ $[-0.024,+0.003]$ & router fails \\
SQuAD & lengths & $0.634$ & $0.042$ & $-0.050$ $[-0.065,-0.035]$ & router fails \\
SQuAD & overlap & $0.766$ & $0.170$ & $+0.120$ $[+0.102,+0.140]^{\ddagger}$ & pass \\
\bottomrule
\end{tabular}
\end{table}

\section{Synthetic Stress Tests: Nonlinear Artifacts, Multiple Artifacts, and Construct Noise}
\label{app:synthstress}

One notational warning applies throughout this appendix: the multi-artifact and nonlinear synthetics report slice gains at $\rho{=}0$, and those are large. There the slice is defined from the \emph{known} artifact $a$, which is what makes it informative at $\rho{=}0$, and what real settings, where only $\hat a$ exists, cannot do (Section~\ref{sec:synthetic}).

\subsection{Nonlinear synthetic artifacts}
\label{app:nonlinear}

The synthetic sweep of Section~\ref{sec:synthetic} is linear by construction. Here the artifact enters the score through nonlinear functions of the artifact features ($N{=}4000$, $\rho{=}0$, score $= c + 1.2a + \varepsilon$ as before): \emph{quadratic} $a=(w^\top\phi)^2$, \emph{XOR} $a=\mathrm{sign}(\phi_1)\mathrm{sign}(\phi_2)$, and \emph{piecewise} $a=\mathbf{1}[\phi_1>0]\cdot(v^\top\phi)$, with a linear control. Table~\ref{tab:nonlinear} shows the screening logic operating as derived in Appendix~\ref{app:identifiability}. One notational care: because the artifact $a$ is \emph{known} in a synthetic, we here measure channel recoverability directly ($r_{\mathrm{chan}}$, the AUC of a linear model recovering the artifact decision from $\phi$), in place of the label-mediated slice-eligibility check statistic $A(\hat a)$ of Section~\ref{sec:condition}, which is necessarily near chance at $\rho{=}0$ (the artifact is independent of the construct label there). Where the channel is not linearly recoverable ($r_{\mathrm{chan}}$ fails: quadratic 0.49, XOR 0.47), linear adjustment is inert ($\pm0.001$ on the declared slice) and the artifact remains readable by a nonlinear probe; the operator has nothing it is warranted to remove, and the recoverability screen predicts this \emph{before} running it. The piecewise case is the instructive intermediate: half its variance is linear, $r_{\mathrm{chan}}$ passes (0.95), and linear adjustment recovers precisely the linear half of the damage ($0.447\to0.676$). A random-forest residualizer adjusts all four cases ($0.76$--$0.91$ on the slice) but forfeits the linear criterion's closed form; it is the natural extension when the slice-eligibility check fails and a nonlinear artifact channel is documented.

\begin{table}[h]
\centering\small
\caption{Nonlinear synthetic artifacts: linear channel recoverability $r_{\mathrm{chan}}$ (measured against the known artifact; see text for why this differs from the label-mediated $A(\hat a)$), declared-slice alignment raw $\to$ linear adjustment $\to$ RF adjustment, and RF-probe artifact floor after linear adjustment.}
\label{tab:nonlinear}
\begin{tabular}{lccccc}
\toprule
Artifact & $r_{\mathrm{chan}}$ & $A_{\mathrm{slice}}$ raw & lin.\ adjustment & RF adjustment & RF floor after lin.\ adjustment \\
\midrule
linear (control) & 1.00 & 0.391 & 0.941 & 0.880 & 0.509 (decorr.) \\
quadratic & 0.49 & 0.432 & 0.433 & 0.761 & 0.721 (recoverable) \\
XOR & 0.47 & 0.188 & 0.188 & 0.834 & 0.915 (recoverable) \\
piecewise & 0.95 & 0.447 & 0.676 & 0.912 & 0.567 (partial) \\
\bottomrule
\end{tabular}
\end{table}

\subsection{Multiple artifacts and noisy constructs}
\label{app:multi}
Real scores rarely carry a single artifact, and real construct measurements are rarely noiseless. The same synthetics are extended to probe both departures ($\rho{=}0$, $N{=}4000$).

\textbf{Two artifact channels.} $s = c + 1.0\,a_A + 0.8\,a_B + \varepsilon$ with independent channels $a_A,a_B$. Three variants (Table~\ref{tab:multi}): when both channels are linear and both feature blocks are in $\phi$, one ridge removes both jointly (declared slice $0.370\to0.944$; both floors close). When $a_B$'s features are \emph{missing from} $\phi$, the adjustment correctly removes only the measured channel: the gain is partial ($\to0.695$) and $a_B$'s floor stays open ($0.781$); the residualized score is better but still unclean, and the open floor is what an auditor who later measures $\phi_B$ would find. When $a_B$ is nonlinear (XOR), the outcome is the same in structure: the linear channel is removed, the nonlinear one survives ($ \to0.644$; floor $0.861$). The operator composes across artifacts channel-by-channel, and its guarantee is scoped to the channels $\phi$ actually measures: an unmeasured artifact shows up as an open floor once its channel is measured; the audit's own report covers only the channels $\phi$ measures.

\begin{table}[h]
\centering\small
\caption{Two-artifact synthetic: declared-slice alignment and per-channel linear floors after adjustment.}
\label{tab:multi}
\begin{tabular}{lccc}
\toprule
Variant & $A_{\mathrm{slice}}$ raw $\to$ rep & floor $a_A$ (rep) & floor $a_B$ (rep) \\
\midrule
both linear, both observed & $0.370\to0.944$ & 0.481 (decorr.) & 0.497 (decorr.) \\
$a_B$ unobserved by $\phi$ & $0.369\to0.695$ & 0.492 (decorr.) & 0.781 (recoverable) \\
$a_B$ nonlinear (XOR) & $0.376\to0.644$ & 0.491 (decorr.) & 0.861 (recoverable) \\
\bottomrule
\end{tabular}
\end{table}

\textbf{Noisy construct measurement.} The slice-gain test compares alignments against $y$; if $y$ itself is noisy, does the outcome survive? The design mirrors an auditor's position: the construct label is flipped independently at $10/20/30\%$ \emph{before} the audit, so the slice is recomputed from the noisy labels and both alignments are evaluated against them (\texttt{multi\_artifact\_noise\_synthetic.py}). The declared-slice \emph{gain}, the quantity the outcome rests on, degrades gently: $+0.539$ clean, then $+0.516$, $+0.482$, $+0.450$. For a slice held fixed, symmetric evaluation noise alone would scale any AUC gap by about $1-2\eta$ (a $0.4\times$ factor at $30\%$); the measured decline is gentler because the slice is recomputed with the noise: flip-created members carry wrong labels for the raw and adjusted scores alike, depressing both slice alignments together (raw $0.410\to0.244$) more than their gap. The outcome (the adjustment helps, strongly) is unchanged even when nearly a third of construct labels are wrong; the cost is statistical power and the calibration of magnitudes, \emph{provided the noise is independent of the artifact}.

\subsection{Correlated construct noise: sensitivity to artifact-dependent label errors}
\label{app:corrnoise}
 Artifact-dependent label errors can reverse the measured effect of adjustment, and a synthetic makes the failure mode concrete (the independence requirement on $y$, Section~\ref{sec:method}). Take $\rho{=}0.8$, collinear enough that the adjustment truly hurts (true full-set gain $-0.191$ in this appendix's own construction, generated independently of the main sweep; an audit should not endorse this adjustment), and corrupt $y$ at a matched overall rate ($\approx20\%$) under four noise models. Noise not directed at the artifact channel leaves the measured sign intact: symmetric ($-0.115$), class-dependent ($-0.136$), and score-correlated judge-style noise ($-0.227$; correlated with the artifact only through the score) all still show the adjustment hurting. But \emph{artifact-correlated} noise (flips concentrated on examples where the artifact agrees with the label, the signature of an annotation pipeline that itself leaned on the artifact) reverses the measured full-set change to $+0.087$. This demonstrates a corrupted construct measurement target: any full-population comparison against this $y$ reads the harmful adjustment as helpful. It does not by itself establish a false approval by the complete procedure, whose outcome also depends on the slice, the screening statistics, and the gain interval; what it establishes is that no measurement resting on this $y$ can be trusted about the original construct. Running the committed pipeline against each noisy $y$ makes the split concrete (\texttt{t3\_verdict.py}): under the three noise models not directed at the artifact, every gate passes and the slice-gain test returns the large positive gain the model prescribes at $\rho{=}0.8$---the regime where the claim is slice-scoped and the dual report carries the warning---while under the artifact-directed noise the router itself refuses ($A(\hat a)=0.491$), because this particular corruption also destroys the slice basis; the false reading materializes in the full-set comparison, not in the procedure's outcome. The failure is bidirectional: in the mirrored construction (a genuinely adjustable $\rho{=}0$ setting, noise flips concentrated where the artifact \emph{disagrees} with the label, overall rate held at $45\%$), independent noise leaves the measured full-set change positive ($+0.014$) while an artifact-directed share of only $25\%$ turns it to $-0.005$ (worsening to $-0.065$ as the share grows). Artifact-dependent noise can therefore reverse the measured full-population effect in either direction. Both reversals are properties of the comparison against the corrupted $y$; neither is an outcome the procedure returned, and in the construction above the router refused before any such comparison entered an outcome. This is the case the independence requirement on $y$ excludes ($y$'s error correlated with the artifact), and it is why a construct measurement sharing the score's annotation pipeline (RewardBench, Section~\ref{sec:boundary-refusals}) cannot be used for the slice-gain test at all: with dependent noise the comparison itself is biased in the direction of the noise, so a passed check cannot be read as evidence about the original construct.

\paragraph{Operator variants.}
Ridge vs.\ OLS residualization and fold sensitivity are quantified in Appendix~\ref{app:robustness}; nonlinear (random-forest) residualizers for nonlinear channels, with their weaker decorrelation diagnostics, in Appendix~\ref{app:nonlinear}. LEACE is not among the variants run here, and the reason is that it degenerates on this object. LEACE guarantees that no linear classifier recovers a concept from a \emph{vector} representation, by an affine transformation that whitens and projects. Applied to a \emph{scalar} score $s$ with a one-dimensional surface index, the guarantee reduces to zero covariance between the transformed score and that index, and within LEACE's own family (affine in the representation, here $as+b$) the only map achieving it is the constant one; achieving decorrelation while keeping any variance requires conditioning on $\phi$, which is exactly the $s-\hat g(\phi)$ operator already deployed in Eq.~(\ref{eq:op}). LEACE-style closed-form projections \citep{belrose2023leace} have nothing to add until the adjusted object is vector-valued (representation-level scores), where per-coordinate ridge residualization no longer guarantees joint linear closure. On a scalar the genuinely different primitive is distributional, which is what the conditional-quantile eraser of Appendix~\ref{app:quantile} instantiates.

\end{document}

%% file: math_commands.tex
\usepackage{amsmath,amsfonts,bm}

\def\eqref#1{equation~\ref{#1}}
\def\1{\bm{1}}

\DeclareMathAlphabet{\mathsfit}{\encodingdefault}{\sfdefault}{m}{sl}
\SetMathAlphabet{\mathsfit}{bold}{\encodingdefault}{\sfdefault}{bx}{n}

%% file: arxiv_main.bbl
\begin{thebibliography}{70}
\providecommand{\natexlab}[1]{#1}
\providecommand{\url}[1]{\texttt{#1}}
\expandafter\ifx\csname urlstyle\endcsname\relax
  \providecommand{\doi}[1]{doi: #1}\else
  \providecommand{\doi}{doi: \begingroup \urlstyle{rm}\Url}\fi

\bibitem[{Anthropic}(2025)]{anthropic2025haiku}
{Anthropic}.
\newblock Claude {H}aiku 4.5.
\newblock \url{https://www.anthropic.com/news/claude-haiku-4-5}, 2025.

\bibitem[Austin et~al.(2021)Austin, Odena, Nye, Bosma, Michalewski, Dohan,
  Jiang, Cai, Terry, Le, and Sutton]{austin2021mbpp}
Jacob Austin, Augustus Odena, Maxwell Nye, Maarten Bosma, Henryk Michalewski,
  David Dohan, Ellen Jiang, Carrie Cai, Michael Terry, Quoc Le, and Charles
  Sutton.
\newblock Program synthesis with large language models.
\newblock \emph{arXiv preprint arXiv:2108.07732}, 2021.

\bibitem[Bai et~al.(2022)Bai, Jones, Ndousse, Askell, Chen, DasSarma, Drain,
  Fort, Ganguli, Henighan, Joseph, Kadavath, Kernion, Conerly, El-Showk,
  Elhage, Hatfield-Dodds, Hernandez, Hume, Johnston, Kravec, Lovitt, Nanda,
  Olsson, Amodei, Brown, Clark, McCandlish, Olah, Mann, and
  Kaplan]{bai2022hhrlhf}
Yuntao Bai, Andy Jones, Kamal Ndousse, Amanda Askell, Anna Chen, Nova DasSarma,
  Dawn Drain, Stanislav Fort, Deep Ganguli, Tom Henighan, Nicholas Joseph,
  Saurav Kadavath, Jackson Kernion, Tom Conerly, Sheer El-Showk, Nelson Elhage,
  Zac Hatfield-Dodds, Danny Hernandez, Tristan Hume, Scott Johnston, Shauna
  Kravec, Liane Lovitt, Neel Nanda, Catherine Olsson, Dario Amodei, Tom Brown,
  Jack Clark, Sam McCandlish, Chris Olah, Ben Mann, and Jared Kaplan.
\newblock Training a helpful and harmless assistant with reinforcement learning
  from human feedback.
\newblock \emph{arXiv preprint arXiv:2204.05862}, 2022.

\bibitem[Bean et~al.(2025)Bean, Kearns, Romanou, Hafner, Mayne, Batzner,
  Foroutan, Schmitz, Korgul, Batra, Deb, Beharry, Emde, Foster, Gausen,
  Grandury, Han, Hofmann, Ibrahim, Kim, Kirk, Lin, Liu, Luettgau, Magomere,
  Rystr{\o}m, Sotnikova, Yang, Zhao, Bibi, Bosselut, Clark, Cohan, Foerster,
  Gal, Hale, Raji, Summerfield, Torr, Ududec, Rocher, and
  Mahdi]{measuring2025matters}
Andrew~M. Bean, Ryan~Othniel Kearns, Angelika Romanou, Franziska~Sofia Hafner,
  Harry Mayne, Jan Batzner, Negar Foroutan, Chris Schmitz, Karolina Korgul,
  Hunar Batra, Oishi Deb, Emma Beharry, Cornelius Emde, Thomas Foster, Anna
  Gausen, Mar{\'i}a Grandury, Simeng Han, Valentin Hofmann, Lujain Ibrahim,
  Hazel Kim, Hannah~Rose Kirk, Fangru Lin, Gabrielle Kaili-May Liu, Lennart
  Luettgau, Jabez Magomere, Jonathan Rystr{\o}m, Anna Sotnikova, Yushi Yang,
  Yilun Zhao, Adel Bibi, Antoine Bosselut, Ronald Clark, Arman Cohan, Jakob
  Foerster, Yarin Gal, Scott~A. Hale, Inioluwa~Deborah Raji, Christopher
  Summerfield, Philip~H.S. Torr, Cozmin Ududec, Luc Rocher, and Adam Mahdi.
\newblock Measuring what matters: Construct validity in large language model
  benchmarks.
\newblock In \emph{Advances in Neural Information Processing Systems}, 2025.
\newblock arXiv:2511.04703.

\bibitem[Belrose et~al.(2023)Belrose, Schneider-Joseph, Ravfogel, Cotterell,
  Raff, and Biderman]{belrose2023leace}
Nora Belrose, David Schneider-Joseph, Shauli Ravfogel, Ryan Cotterell, Edward
  Raff, and Stella Biderman.
\newblock {LEACE}: Perfect linear concept erasure in closed form.
\newblock \emph{Advances in Neural Information Processing Systems}, 2023.

\bibitem[Borkan et~al.(2019)Borkan, Dixon, Sorensen, Thain, and
  Vasserman]{borkan2019civil}
Daniel Borkan, Lucas Dixon, Jeffrey Sorensen, Nithum Thain, and Lucy Vasserman.
\newblock Nuanced metrics for measuring unintended bias with real data for text
  classification.
\newblock In \emph{Companion Proceedings of WWW}, 2019.

\bibitem[Bowman et~al.(2015)Bowman, Angeli, Potts, and Manning]{bowman2015snli}
Samuel~R. Bowman, Gabor Angeli, Christopher Potts, and Christopher~D. Manning.
\newblock A large annotated corpus for learning natural language inference.
\newblock In \emph{Proceedings of EMNLP}, 2015.

\bibitem[Cer et~al.(2017)Cer, Diab, Agirre, Lopez-Gazpio, and
  Specia]{cer2017stsb}
Daniel Cer, Mona Diab, Eneko Agirre, I{\~n}igo Lopez-Gazpio, and Lucia Specia.
\newblock {S}em{E}val-2017 task 1: Semantic textual similarity multilingual and
  crosslingual focused evaluation.
\newblock In \emph{Proceedings of the 11th International Workshop on Semantic
  Evaluation (SemEval-2017)}, pp.\  1--14, 2017.

\bibitem[Chen et~al.(2026)Chen, Chen, Lin, and Vong]{chen2026judgevalidity}
Jianlin Chen, Wenhui Chen, Ziyao Lin, and Chi~Man Vong.
\newblock A judge should know what changed: Construct validity for
  {LLM}-as-a-judge evaluation.
\newblock \emph{arXiv preprint arXiv:2608.24419}, 2026.

\bibitem[Chen et~al.(2021)Chen, Tworek, Jun, Yuan, Pinto, Kaplan, Edwards,
  Burda, Joseph, Brockman, Ray, Puri, Krueger, Petrov, Khlaaf, Sastry, Mishkin,
  Chan, Gray, Ryder, Pavlov, Power, Kaiser, Bavarian, Winter, Tillet, Such,
  Cummings, Plappert, Chantzis, Barnes, Herbert-Voss, Guss, Nichol, Paino,
  Tezak, Tang, Babuschkin, Balaji, Jain, Saunders, Hesse, Carr, Leike, Achiam,
  Misra, Morikawa, Radford, Knight, Brundage, Murati, Mayer, Welinder, McGrew,
  Amodei, McCandlish, Sutskever, and Zaremba]{chen2021humaneval}
Mark Chen, Jerry Tworek, Heewoo Jun, Qiming Yuan, Henrique Ponde de~Oliveira
  Pinto, Jared Kaplan, Harri Edwards, Yuri Burda, Nicholas Joseph, Greg
  Brockman, Alex Ray, Raul Puri, Gretchen Krueger, Michael Petrov, Heidy
  Khlaaf, Girish Sastry, Pamela Mishkin, Brooke Chan, Scott Gray, Nick Ryder,
  Mikhail Pavlov, Alethea Power, Lukasz Kaiser, Mohammad Bavarian, Clemens
  Winter, Philippe Tillet, Felipe~Petroski Such, Dave Cummings, Matthias
  Plappert, Fotios Chantzis, Elizabeth Barnes, Ariel Herbert-Voss,
  William~Hebgen Guss, Alex Nichol, Alex Paino, Nikolas Tezak, Jie Tang, Igor
  Babuschkin, Suchir Balaji, Shantanu Jain, William Saunders, Christopher
  Hesse, Andrew~N. Carr, Jan Leike, Josh Achiam, Vedant Misra, Evan Morikawa,
  Alec Radford, Matthew Knight, Miles Brundage, Mira Murati, Katie Mayer, Peter
  Welinder, Bob McGrew, Dario Amodei, Sam McCandlish, Ilya Sutskever, and
  Wojciech Zaremba.
\newblock Evaluating large language models trained on code.
\newblock \emph{arXiv preprint arXiv:2107.03374}, 2021.

\bibitem[Chernozhukov et~al.(2018)Chernozhukov, Chetverikov, Demirer, Duflo,
  Hansen, Newey, and Robins]{chernozhukov2018double}
Victor Chernozhukov, Denis Chetverikov, Mert Demirer, Esther Duflo, Christian
  Hansen, Whitney Newey, and James Robins.
\newblock Double/debiased machine learning for treatment and structural
  parameters.
\newblock \emph{The Econometrics Journal}, 21\penalty0 (1):\penalty0 C1--C68,
  2018.

\bibitem[Clark et~al.(2019{\natexlab{a}})Clark, Lee, Chang, Kwiatkowski,
  Collins, and Toutanova]{clark2019boolq}
Christopher Clark, Kenton Lee, Ming-Wei Chang, Tom Kwiatkowski, Michael
  Collins, and Kristina Toutanova.
\newblock {B}ool{Q}: Exploring the surprising difficulty of natural yes/no
  questions.
\newblock In \emph{Proceedings of the 2019 Conference of the North American
  Chapter of the Association for Computational Linguistics: Human Language
  Technologies}, pp.\  2924--2936, 2019{\natexlab{a}}.

\bibitem[Clark et~al.(2019{\natexlab{b}})Clark, Yatskar, and
  Zettlemoyer]{clark2019dont}
Christopher Clark, Mark Yatskar, and Luke Zettlemoyer.
\newblock Don't take the easy way out: Ensemble based methods for avoiding
  known dataset biases.
\newblock In \emph{Proceedings of EMNLP}, 2019{\natexlab{b}}.

\bibitem[Clark et~al.(2020)Clark, Luong, Le, and Manning]{clark2020electra}
Kevin Clark, Minh-Thang Luong, Quoc~V. Le, and Christopher~D. Manning.
\newblock {ELECTRA}: Pre-training text encoders as discriminators rather than
  generators.
\newblock In \emph{International Conference on Learning Representations
  (ICLR)}, 2020.

\bibitem[Cronbach \& Meehl(1955)Cronbach and Meehl]{cronbach1955construct}
Lee~J. Cronbach and Paul~E. Meehl.
\newblock Construct validity in psychological tests.
\newblock \emph{Psychological Bulletin}, 52\penalty0 (4):\penalty0 281--302,
  1955.

\bibitem[Devlin et~al.(2019)Devlin, Chang, Lee, and Toutanova]{devlin2019bert}
Jacob Devlin, Ming-Wei Chang, Kenton Lee, and Kristina Toutanova.
\newblock {BERT}: Pre-training of deep bidirectional transformers for language
  understanding.
\newblock In \emph{Proceedings of the 2019 Conference of the North {A}merican
  Chapter of the Association for Computational Linguistics: Human Language
  Technologies, Volume 1 (Long and Short Papers)}, pp.\  4171--4186, 2019.
\newblock \doi{10.18653/v1/N19-1423}.

\bibitem[Dubois et~al.(2024)Dubois, Galambosi, Liang, and
  Hashimoto]{dubois2024length}
Yann Dubois, Bal{\'a}zs Galambosi, Percy Liang, and Tatsunori~B. Hashimoto.
\newblock Length-controlled {AlpacaEval}: A simple way to debias automatic
  evaluators.
\newblock In \emph{Conference on Language Modeling (COLM)}, 2024.
\newblock arXiv:2404.04475.

\bibitem[Elazar \& Goldberg(2018)Elazar and Goldberg]{elazar2018adversarial}
Yanai Elazar and Yoav Goldberg.
\newblock Adversarial removal of demographic attributes from text data.
\newblock In \emph{Proceedings of the 2018 Conference on Empirical Methods in
  Natural Language Processing}, pp.\  11--21, 2018.

\bibitem[Ethayarajh et~al.(2022)Ethayarajh, Choi, and
  Swayamdipta]{ethayarajh2022shp}
Kawin Ethayarajh, Yejin Choi, and Swabha Swayamdipta.
\newblock Understanding dataset difficulty with $\mathcal{V}$-usable
  information.
\newblock In \emph{Proceedings of the 39th International Conference on Machine
  Learning}, volume 162 of \emph{PMLR}, pp.\  5988--6008, 2022.

\bibitem[Eyuboglu et~al.(2022)Eyuboglu, Varma, Saab, Delbrouck, Lee-Messer,
  Dunnmon, Zou, and R{\'e}]{eyuboglu2022domino}
Sabri Eyuboglu, Maya Varma, Khaled Saab, Jean-Benoit Delbrouck, Christopher
  Lee-Messer, Jared Dunnmon, James Zou, and Christopher R{\'e}.
\newblock Domino: Discovering systematic errors with cross-modal embeddings.
\newblock In \emph{International Conference on Learning Representations}, 2022.

\bibitem[Feng et~al.(2019)Feng, Wallace, and Boyd-Graber]{feng2019misleading}
Shi Feng, Eric Wallace, and Jordan Boyd-Graber.
\newblock Misleading failures of partial-input baselines.
\newblock In \emph{Proceedings of the 57th Annual Meeting of the Association
  for Computational Linguistics}, pp.\  5533--5538, 2019.

\bibitem[Frisch \& Waugh(1933)Frisch and Waugh]{frisch1933partial}
Ragnar Frisch and Frederick~V. Waugh.
\newblock Partial time regressions as compared with individual trends.
\newblock \emph{Econometrica}, 1\penalty0 (4):\penalty0 387--401, 1933.

\bibitem[Garg et~al.(2020)Garg, Vu, and Moschitti]{garg2020tanda}
Siddhant Garg, Thuy Vu, and Alessandro Moschitti.
\newblock {TANDA}: Transfer and adapt pre-trained transformer models for answer
  sentence selection.
\newblock In \emph{Proceedings of AAAI}, 2020.

\bibitem[Geirhos et~al.(2020)Geirhos, Jacobsen, Michaelis, Zemel, Brendel,
  Bethge, and Wichmann]{geirhos2020shortcut}
Robert Geirhos, J{\"o}rn-Henrik Jacobsen, Claudio Michaelis, Richard Zemel,
  Wieland Brendel, Matthias Bethge, and Felix~A. Wichmann.
\newblock Shortcut learning in deep neural networks.
\newblock \emph{Nature Machine Intelligence}, 2:\penalty0 665--673, 2020.

\bibitem[Gonen \& Goldberg(2019)Gonen and Goldberg]{gonen2019lipstick}
Hila Gonen and Yoav Goldberg.
\newblock Lipstick on a pig: Debiasing methods cover up systematic gender
  biases in word embeddings but do not remove them.
\newblock In \emph{Proceedings of the 2019 Conference of the North American
  Chapter of the Association for Computational Linguistics: Human Language
  Technologies, Volume 1 (Long and Short Papers)}, pp.\  609--614, 2019.

\bibitem[Gururangan et~al.(2018)Gururangan, Swayamdipta, Levy, Schwartz,
  Bowman, and Smith]{gururangan2018annotation}
Suchin Gururangan, Swabha Swayamdipta, Omer Levy, Roy Schwartz, Samuel~R.
  Bowman, and Noah~A. Smith.
\newblock Annotation artifacts in natural language inference data.
\newblock In \emph{Proceedings of NAACL-HLT}, 2018.

\bibitem[He et~al.(2019)He, Zha, and Wang]{he2019drift}
He~He, Sheng Zha, and Haohan Wang.
\newblock Unlearn dataset bias in natural language inference by fitting the
  residual.
\newblock In \emph{Proceedings of the 2nd Workshop on Deep Learning Approaches
  for Low-Resource NLP}, 2019.

\bibitem[He et~al.(2021)He, Liu, Gao, and Chen]{he2021deberta}
Pengcheng He, Xiaodong Liu, Jianfeng Gao, and Weizhu Chen.
\newblock {D}e{BERT}a: Decoding-enhanced {BERT} with disentangled attention.
\newblock In \emph{International Conference on Learning Representations
  (ICLR)}, 2021.

\bibitem[Holland \& Wainer(1993)Holland and Wainer]{holland1993dif}
Paul~W. Holland and Howard Wainer (eds.).
\newblock \emph{Differential Item Functioning}.
\newblock Lawrence Erlbaum Associates, 1993.

\bibitem[Huang et~al.(2025)Huang, Qiu, Wang, Ponti, and
  Titov]{huang2024posthoc}
Zeyu Huang, Zihan Qiu, Zili Wang, Edoardo~M. Ponti, and Ivan Titov.
\newblock Post-hoc reward calibration: A case study on length bias.
\newblock In \emph{International Conference on Learning Representations
  (ICLR)}, 2025.
\newblock arXiv:2409.17407.

\bibitem[Jacobs \& Wallach(2021)Jacobs and Wallach]{jacobs2021measurement}
Abigail~Z. Jacobs and Hanna Wallach.
\newblock Measurement and fairness.
\newblock In \emph{Proceedings of FAccT}, 2021.

\bibitem[Joshi et~al.(2017)Joshi, Choi, Weld, and
  Zettlemoyer]{joshi2017triviaqa}
Mandar Joshi, Eunsol Choi, Daniel Weld, and Luke Zettlemoyer.
\newblock {T}rivia{QA}: A large scale distantly supervised challenge dataset
  for reading comprehension.
\newblock In \emph{Proceedings of the 55th Annual Meeting of the Association
  for Computational Linguistics (Volume 1: Long Papers)}, pp.\  1601--1611,
  2017.

\bibitem[Karimi~Mahabadi et~al.(2020)Karimi~Mahabadi, Belinkov, and
  Henderson]{mahabadi2020endtoend}
Rabeeh Karimi~Mahabadi, Yonatan Belinkov, and James Henderson.
\newblock End-to-end bias mitigation by modelling biases in corpora.
\newblock In \emph{Proceedings of ACL}, 2020.

\bibitem[K{\"o}pf et~al.(2023)K{\"o}pf, Kilcher, von R{\"u}tte, Anagnostidis,
  Tam, Stevens, Barhoum, Nguyen, Stanley, Nagyfi, ES, Suri, Glushkov,
  Dantuluri, Maguire, Schuhmann, Nguyen, and Mattick]{kopf2023openassistant}
Andreas K{\"o}pf, Yannic Kilcher, Dimitri von R{\"u}tte, Sotiris Anagnostidis,
  Zhi~Rui Tam, Keith Stevens, Abdullah Barhoum, Duc Nguyen, Oliver Stanley,
  Rich{\'a}rd Nagyfi, Shahul ES, Sameer Suri, David Glushkov, Arnav Dantuluri,
  Andrew Maguire, Christoph Schuhmann, Huu Nguyen, and Alexander Mattick.
\newblock {OpenAssistant} conversations -- democratizing large language model
  alignment.
\newblock In \emph{Advances in Neural Information Processing Systems}, 2023.

\bibitem[Kwiatkowski et~al.(2019)Kwiatkowski, Palomaki, Redfield, Collins,
  Parikh, Alberti, Epstein, Polosukhin, Devlin, Lee, Toutanova, Jones, Kelcey,
  Chang, Dai, Uszkoreit, Le, and Petrov]{kwiatkowski2019nq}
Tom Kwiatkowski, Jennimaria Palomaki, Olivia Redfield, Michael Collins, Ankur
  Parikh, Chris Alberti, Danielle Epstein, Illia Polosukhin, Jacob Devlin,
  Kenton Lee, Kristina Toutanova, Llion Jones, Matthew Kelcey, Ming-Wei Chang,
  Andrew~M. Dai, Jakob Uszkoreit, Quoc Le, and Slav Petrov.
\newblock Natural {Questions}: A benchmark for question answering research.
\newblock \emph{Transactions of the Association for Computational Linguistics},
  7:\penalty0 452--466, 2019.

\bibitem[Lambert et~al.(2025)Lambert, Pyatkin, Morrison, Miranda, Lin, Chandu,
  Dziri, Kumar, Zick, Choi, Smith, and Hajishirzi]{lambert2024rewardbench}
Nathan Lambert, Valentina Pyatkin, Jacob Morrison, LJ~Miranda, Bill~Yuchen Lin,
  Khyathi Chandu, Nouha Dziri, Sachin Kumar, Tom Zick, Yejin Choi, Noah~A.
  Smith, and Hannaneh Hajishirzi.
\newblock {RewardBench}: Evaluating reward models for language modeling.
\newblock In \emph{Findings of the Association for Computational Linguistics:
  NAACL 2025}, pp.\  1755--1797, 2025.
\newblock arXiv:2403.13787.

\bibitem[Lamparth et~al.(2026)Lamparth, Fein, Haupt, Hussing, and
  Kochenderfer]{lamparth2026substitution}
Max Lamparth, Daniel Fein, Andreas Haupt, Marcel Hussing, and Mykel~J.
  Kochenderfer.
\newblock Reward bias substitution: Single-axis bias mitigations redirect
  optimization pressure.
\newblock \emph{arXiv preprint arXiv:2605.27996}, 2026.

\bibitem[Li et~al.(2024)Li, Angelopoulos, and Chiang]{li2024style}
Tianle Li, Anastasios Angelopoulos, and Wei-Lin Chiang.
\newblock Does style matter? {D}isentangling style and substance in {C}hatbot
  {A}rena.
\newblock LMSYS Org Blog, 2024.
\newblock URL \url{https://lmsys.org/blog/2024-08-28-style-control/}.

\bibitem[Li et~al.(2025)Li, Chiang, Frick, Dunlap, Wu, Zhu, Gonzalez, and
  Stoica]{li2025arenahard}
Tianle Li, Wei-Lin Chiang, Evan Frick, Lisa Dunlap, Tianhao Wu, Banghua Zhu,
  Joseph~E. Gonzalez, and Ion Stoica.
\newblock From crowdsourced data to high-quality benchmarks: {A}rena-{H}ard and
  {B}ench{B}uilder pipeline.
\newblock In \emph{Proceedings of the 42nd International Conference on Machine
  Learning}, pp.\  34209--34231, 2025.

\bibitem[Liu et~al.(2019)Liu, Ott, Goyal, Du, Joshi, Chen, Levy, Lewis,
  Zettlemoyer, and Stoyanov]{liu2019roberta}
Yinhan Liu, Myle Ott, Naman Goyal, Jingfei Du, Mandar Joshi, Danqi Chen, Omer
  Levy, Mike Lewis, Luke Zettlemoyer, and Veselin Stoyanov.
\newblock {R}o{BERT}a: A robustly optimized {BERT} pretraining approach.
\newblock \emph{arXiv preprint arXiv:1907.11692}, 2019.

\bibitem[Lovell(1963)]{lovell1963seasonal}
Michael~C. Lovell.
\newblock Seasonal adjustment of economic time series and multiple regression
  analysis.
\newblock \emph{Journal of the American Statistical Association}, 58\penalty0
  (304):\penalty0 993--1010, 1963.

\bibitem[Marelli et~al.(2014)Marelli, Menini, Baroni, Bentivogli, Bernardi, and
  Zamparelli]{marelli2014sick}
Marco Marelli, Stefano Menini, Marco Baroni, Luisa Bentivogli, Raffaella
  Bernardi, and Roberto Zamparelli.
\newblock A {SICK} cure for the evaluation of compositional distributional
  semantic models.
\newblock In \emph{Proceedings of LREC}, 2014.

\bibitem[McCoy et~al.(2019)McCoy, Pavlick, and Linzen]{mccoy2019hans}
R.~Thomas McCoy, Ellie Pavlick, and Tal Linzen.
\newblock Right for the wrong reasons: Diagnosing syntactic heuristics in
  natural language inference.
\newblock In \emph{Proceedings of ACL}, 2019.

\bibitem[Meredith(1993)]{meredith1993invariance}
William Meredith.
\newblock Measurement invariance, factor analysis and factorial invariance.
\newblock \emph{Psychometrika}, 58\penalty0 (4):\penalty0 525--543, 1993.

\bibitem[Millsap(2011)]{millsap2011invariance}
Roger~E. Millsap.
\newblock \emph{Statistical Approaches to Measurement Invariance}.
\newblock Routledge, 2011.

\bibitem[Nie et~al.(2020)Nie, Williams, Dinan, Bansal, Weston, and
  Kiela]{nie2020anli}
Yixin Nie, Adina Williams, Emily Dinan, Mohit Bansal, Jason Weston, and Douwe
  Kiela.
\newblock Adversarial {NLI}: A new benchmark for natural language
  understanding.
\newblock In \emph{Proceedings of ACL}, 2020.

\bibitem[Poliak et~al.(2018)Poliak, Naradowsky, Haldar, Rudinger, and
  Van~Durme]{poliak2018hypothesis}
Adam Poliak, Jason Naradowsky, Aparajita Haldar, Rachel Rudinger, and Benjamin
  Van~Durme.
\newblock Hypothesis only baselines in natural language inference.
\newblock In \emph{Proceedings of *SEM}, 2018.

\bibitem[Rajpurkar et~al.(2016)Rajpurkar, Zhang, Lopyrev, and
  Liang]{rajpurkar2016squad}
Pranav Rajpurkar, Jian Zhang, Konstantin Lopyrev, and Percy Liang.
\newblock {SQ}u{AD}: 100,000+ questions for machine comprehension of text.
\newblock In \emph{Proceedings of the 2016 Conference on Empirical Methods in
  Natural Language Processing}, pp.\  2383--2392, 2016.

\bibitem[Ravfogel et~al.(2020)Ravfogel, Elazar, Gonen, Twiton, and
  Goldberg]{ravfogel2020inlp}
Shauli Ravfogel, Yanai Elazar, Hila Gonen, Michael Twiton, and Yoav Goldberg.
\newblock Null it out: Guarding protected attributes by iterative nullspace
  projection.
\newblock In \emph{Proceedings of ACL}, 2020.

\bibitem[Ravfogel et~al.(2022)Ravfogel, Twiton, Goldberg, and
  Cotterell]{ravfogel2022rlace}
Shauli Ravfogel, Michael Twiton, Yoav Goldberg, and Ryan Cotterell.
\newblock Linear adversarial concept erasure.
\newblock In \emph{Proceedings of the 39th International Conference on Machine
  Learning}, volume 162 of \emph{Proceedings of Machine Learning Research},
  pp.\  18400--18421, 2022.
\newblock arXiv:2201.12091.

\bibitem[Reimers \& Gurevych(2019)Reimers and
  Gurevych]{reimers2019sentencebert}
Nils Reimers and Iryna Gurevych.
\newblock Sentence-{BERT}: Sentence embeddings using siamese {BERT}-networks.
\newblock In \emph{Proceedings of EMNLP}, 2019.

\bibitem[R\"{o}ttger et~al.(2021)R\"{o}ttger, Vidgen, Nguyen, Waseem, Margetts,
  and Pierrehumbert]{rottger2021hatecheck}
Paul R\"{o}ttger, Bertie Vidgen, Dong Nguyen, Zeerak Waseem, Helen Margetts,
  and Janet Pierrehumbert.
\newblock {H}ate{C}heck: Functional tests for hate speech detection models.
\newblock In \emph{Proceedings of the 59th Annual Meeting of the Association
  for Computational Linguistics and the 11th International Joint Conference on
  Natural Language Processing (Volume 1: Long Papers)}, pp.\  41--58, 2021.

\bibitem[Saha et~al.(2018)Saha, Aralikatte, Khapra, and
  Sankaranarayanan]{saha2018duorc}
Amrita Saha, Rahul Aralikatte, Mitesh~M. Khapra, and Karthik Sankaranarayanan.
\newblock {DuoRC}: Towards complex language understanding with paraphrased
  reading comprehension.
\newblock In \emph{Proceedings of ACL}, 2018.

\bibitem[Sahoo et~al.(2026)Sahoo, Karnuthala, Budhwani, Agarwal, Vaidyanathan,
  Siu, Dernoncourt, Healey, Lipka, Rossi, Bhattacharya, and
  Kveton]{sahoo2026quantjudges}
Aishwarya Sahoo, Jeevana~Kruthi Karnuthala, Tushar~Parmanand Budhwani, Pranchal
  Agarwal, Sankaran Vaidyanathan, Alexa Siu, Franck Dernoncourt, Jennifer
  Healey, Nedim Lipka, Ryan~A. Rossi, Uttaran Bhattacharya, and Branislav
  Kveton.
\newblock Quantitative {LLM} judges using post-hoc score calibration.
\newblock \emph{Transactions on Machine Learning Research}, 2026.
\newblock arXiv:2506.02945.

\bibitem[Sanh et~al.(2019)Sanh, Debut, Chaumond, and Wolf]{sanh2019distilbert}
Victor Sanh, Lysandre Debut, Julien Chaumond, and Thomas Wolf.
\newblock {D}istil{BERT}, a distilled version of {BERT}: smaller, faster,
  cheaper and lighter.
\newblock \emph{arXiv preprint arXiv:1910.01108}, 2019.

\bibitem[Singhal et~al.(2024)Singhal, Goyal, Xu, and Durrett]{singhal2023long}
Prasann Singhal, Tanya Goyal, Jiacheng Xu, and Greg Durrett.
\newblock A long way to go: Investigating length correlations in {RLHF}.
\newblock In \emph{Conference on Language Modeling (COLM)}, 2024.
\newblock arXiv:2310.03716.

\bibitem[Soumik(2026)]{soumik2026judging}
Sadman~Kabir Soumik.
\newblock Judging the judges: A systematic evaluation of bias mitigation
  strategies in {LLM}-as-a-judge pipelines.
\newblock \emph{Transactions on Machine Learning Research}, 2026.
\newblock arXiv:2604.23178.

\bibitem[Utama et~al.(2020)Utama, Moosavi, and Gurevych]{utama2020tradeoff}
Prasetya~Ajie Utama, Nafise~Sadat Moosavi, and Iryna Gurevych.
\newblock Mind the trade-off: Debiasing {NLU} models without degrading the
  in-distribution performance.
\newblock In \emph{Proceedings of ACL}, 2020.

\bibitem[Wang et~al.(2019)Wang, Singh, Michael, Hill, Levy, and
  Bowman]{wang2019glue}
Alex Wang, Amanpreet Singh, Julian Michael, Felix Hill, Omer Levy, and
  Samuel~R. Bowman.
\newblock {GLUE}: A multi-task benchmark and analysis platform for natural
  language understanding.
\newblock In \emph{International Conference on Learning Representations
  (ICLR)}, 2019.
\newblock arXiv:1804.07461.

\bibitem[Wang et~al.(2020)Wang, Wei, Dong, Bao, Yang, and Zhou]{wang2020minilm}
Wenhui Wang, Furu Wei, Li~Dong, Hangbo Bao, Nan Yang, and Ming Zhou.
\newblock {M}ini{LM}: Deep self-attention distillation for task-agnostic
  compression of pre-trained transformers.
\newblock In \emph{Advances in Neural Information Processing Systems},
  volume~33, pp.\  5776--5788, 2020.

\bibitem[Welbl et~al.(2017)Welbl, Liu, and Gardner]{welbl2017sciq}
Johannes Welbl, Nelson~F. Liu, and Matt Gardner.
\newblock Crowdsourcing multiple choice science questions.
\newblock In \emph{Proceedings of the 3rd Workshop on Noisy User-generated Text
  (W-NUT)}, pp.\  94--106, 2017.

\bibitem[Williams et~al.(2018)Williams, Nangia, and Bowman]{williams2018mnli}
Adina Williams, Nikita Nangia, and Samuel~R. Bowman.
\newblock A broad-coverage challenge corpus for sentence understanding through
  inference.
\newblock In \emph{Proceedings of NAACL-HLT}, 2018.

\bibitem[Wu et~al.(2026)Wu, Han, and Cai]{wu2026lightning}
Yecheng Wu, Song Han, and Han Cai.
\newblock Lightning {OPD} 2.0: Mitigating style bias in cross-teacher on-policy
  distillation for large reasoning models.
\newblock \emph{arXiv preprint arXiv:2607.28449}, 2026.

\bibitem[Xu et~al.(2026)Xu, Zeng, Paisley, and Zhao]{xu2026debias}
Jian Xu, Delu Zeng, John Paisley, and Qibin Zhao.
\newblock When can you debias an {LLM} judge? {I}dentifiability limits, a test,
  and designs for top-k ranking.
\newblock \emph{arXiv preprint arXiv:2607.02104}, 2026.

\bibitem[Yang et~al.(2015)Yang, Yih, and Meek]{yang2015wikiqa}
Yi~Yang, Wen-tau Yih, and Christopher Meek.
\newblock {WikiQA}: A challenge dataset for open-domain question answering.
\newblock In \emph{Proceedings of EMNLP}, 2015.

\bibitem[Zellers et~al.(2018)Zellers, Bisk, Schwartz, and
  Choi]{zellers2018swag}
Rowan Zellers, Yonatan Bisk, Roy Schwartz, and Yejin Choi.
\newblock {SWAG}: A large-scale adversarial dataset for grounded commonsense
  inference.
\newblock In \emph{Proceedings of EMNLP}, 2018.

\bibitem[Zellers et~al.(2019)Zellers, Holtzman, Bisk, Farhadi, and
  Choi]{zellers2019hellaswag}
Rowan Zellers, Ari Holtzman, Yonatan Bisk, Ali Farhadi, and Yejin Choi.
\newblock {HellaSwag}: Can a machine really finish your sentence?
\newblock In \emph{Proceedings of ACL}, 2019.

\bibitem[Zhao et~al.(2026{\natexlab{a}})Zhao, Shin, Huang, Namburi, and
  Sala]{zhao2026care}
Jitian Zhao, Changho Shin, Tzu-Heng Huang, Satya Sai~Srinath Namburi, and
  Frederic Sala.
\newblock {CARE}: Confounder-aware aggregation for reliable {LLM} evaluation.
\newblock In \emph{International Conference on Machine Learning (ICML)},
  2026{\natexlab{a}}.
\newblock arXiv:2603.00039.

\bibitem[Zhao et~al.(2026{\natexlab{b}})Zhao, Cai, Zhu, Sun, Xue, Zhou, Li, and
  Li]{zhao2025biasfitting}
Kangwen Zhao, Jianfeng Cai, Jinhua Zhu, Ruopei Sun, Dongyun Xue, Wengang Zhou,
  Li~Li, and Houqiang Li.
\newblock Bias fitting to mitigate length bias of reward model in {RLHF}.
\newblock In \emph{Proceedings of the 64th Annual Meeting of the Association
  for Computational Linguistics (Volume 1: Long Papers)}, pp.\  2912--2927,
  2026{\natexlab{b}}.
\newblock arXiv:2505.12843.

\bibitem[Zheng et~al.(2023)Zheng, Chiang, Sheng, Zhuang, Wu, Zhuang, Lin, Li,
  Li, Xing, Zhang, Gonzalez, and Stoica]{zheng2023judging}
Lianmin Zheng, Wei-Lin Chiang, Ying Sheng, Siyuan Zhuang, Zhanghao Wu, Yonghao
  Zhuang, Zi~Lin, Zhuohan Li, Dacheng Li, Eric~P. Xing, Hao Zhang, Joseph~E.
  Gonzalez, and Ion Stoica.
\newblock Judging {LLM}-as-a-judge with {MT}-{B}ench and {C}hatbot {A}rena.
\newblock In \emph{Advances in Neural Information Processing Systems}, 2023.
\newblock arXiv:2306.05685.

\end{thebibliography}
